%% file: preprint.tex
\documentclass{article}

\usepackage{iclr2027_conference,times}
\input{paper_preamble}
\preprintversiontrue
\iclrfinalcopy

\begin{document}
\maketitle
\lhead{Preprint}
\input{paper_body}

\end{document}

%% file: paper_preamble.tex
\usepackage[utf8]{inputenc}
\usepackage[T1]{fontenc}
\usepackage{amsmath,amssymb,amsfonts,mathtools}
\usepackage{array,booktabs,multirow}
\usepackage{graphicx}
\usepackage{wrapfig}
\usepackage{needspace}
\usepackage{microtype}
\usepackage{xspace}
\usepackage{enumitem}
\usepackage{tikz}
\usetikzlibrary{arrows.meta,calc,decorations.pathreplacing,positioning,shapes.geometric}
\usepackage{pgfplots}
\pgfplotsset{compat=1.17}
\usepgfplotslibrary{groupplots}
\usepackage{algorithm}
\usepackage{algpseudocode}
\usepackage{placeins}
\usepackage{hyperref}
\usepackage{url}
\usepackage[nameinlink,capitalise]{cleveref}
\crefname{appendixsection}{Appendix}{Appendices}
\Crefname{appendixsection}{Appendix}{Appendices}
\crefname{appendix}{Appendix}{Appendices}
\Crefname{appendix}{Appendix}{Appendices}
\crefname{subappendix}{Appendix}{Appendices}
\Crefname{subappendix}{Appendix}{Appendices}
\crefname{subsubappendix}{Appendix}{Appendices}
\Crefname{subsubappendix}{Appendix}{Appendices}
\hypersetup{hidelinks}

\newif\ifworkshopversion
\newif\ifpreprintversion

\newcommand{\iid}{i.i.d.\xspace}

\setlist[itemize]{leftmargin=*,topsep=2pt,itemsep=1pt,parsep=0pt}
\setlist[itemize,2]{topsep=1pt,itemsep=0pt}

\title{Task Vector Descent: \\Learning from Non-IID Batches}
\author{%
Anton Baumann\textsuperscript{\normalfont 1,2}\quad
Jonas Hübotter\textsuperscript{\normalfont 1}\quad
Zeynep Akata\textsuperscript{\normalfont 2}\quad
Andreas Krause\textsuperscript{\normalfont 1}\\
\normalfont\textsuperscript{1}ETH Zurich\quad
\normalfont\textsuperscript{2}Technical University of Munich
}

%% file: paper_body.tex
\begin{abstract}

\input{sections/00_abstract}
\end{abstract}

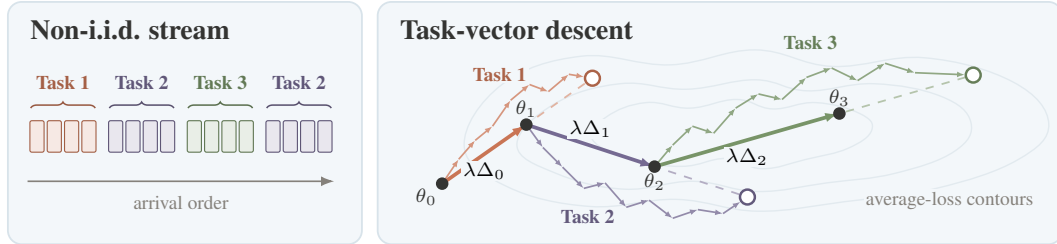
\begin{figure}[h]
    \noindent\makebox[\linewidth][c]{%
        \resizebox{\linewidth}{!}{%
            \input{figures/overview_task_vector}%
        }%
    }
    \caption{Task-vector updates for non-\iid minibatch streams.
    Minibatches arrive in task-clustered sequences (left). For each task,
    noisy minibatch updates produce a local endpoint; the continuing model
    applies a scaled task vector and begins the next task from the resulting
    checkpoint (right).}
    \label{fig:overview}
\end{figure}

\input{sections/01_introduction}
\input{sections/02_related_work}
\input{sections/03_method}
\input{sections/04_experiments}
\input{sections/05_results}
\input{sections/06_discussion}

\ifpreprintversion
\section*{Acknowledgments}
This project was supported in part by the Swiss National Science Foundation under NCCR Automation, grant agreement 51NF40 180545, and by the Swiss AI compute grant a156.
\fi

\section*{Reproducibility Statement}
Complete descriptions of the method, training objectives, model and optimizer configurations, stream construction, evaluation procedures, and random seeds are provided in~\Cref{app:method,app:setup}.
\ifpreprintversion
Code is available at \url{https://github.com/antonbaumann/tvd}.
\else
We intend to release the code and experiment configurations required to reproduce the results.
\fi

\section*{Statement on the use of AI}
In this paper, LLMs were used for minor grammatical edits, word polishing, or rephrasing, as well as for assistance with the implementation of the proposed method and experimental code.
All suggestions from LLMs were manually verified and edited by the authors prior to final inclusion.

Further, LLMs were used as components of the experimental methodology. In the persona-adaptation experiments, we used Qwen3-8B as a user simulator during training and as a pairwise judge during evaluation. In the reasoning SFT experiments, Claude Sonnet 4.6 was used to generate fixed reasoning traces and answers for the training examples. 

\bibliography{refs}
\bibliographystyle{iclr2027_conference}

\clearpage
\appendix
\section*{Appendix}
The appendix is organized as follows. \Cref{app:extended-related-work}
provides an extended discussion of related work, and
\cref{app:sec:analysis} analyzes task-vector updates.
\Cref{app:method,app:setup} give the complete method and experimental
specifications, and \cref{app:extended-results} provides additional results
and interpolation diagnostics.

\input{sections/appendix}

%% file: sections/00_abstract.tex
A central challenge in continual learning is to acquire new knowledge without forgetting what the model has already learned.
This challenge appears in language model training when training data comes from various domain-, user-, or task-specific distributions that are encountered unevenly over time.
In such settings, successive minibatches are temporally clustered by distribution instead of being sampled \iid from the overall data mixture.
Training on temporally clustered data induces a stability-plasticity tradeoff. 
Adapting the model to the active distribution can improve the model on the active distribution but may lead to a performance degradation on data it previously trained on.
We find that this tradeoff intensifies with longer exposure to the same distribution.
We therefore ask if the parameter displacement produced by such a sequence (the task vector) should be fully retained or applied partially.
We compare applying the full displacement ($\lambda=1$) with partial integration, which scales the task vector by $\lambda$ before applying it to the continuing model and scales the optimizer state by the same coefficient.
Across continual pretraining, pretraining from random initialization, supervised post-training, and reinforcement post-training, we find that intermediate values of $\lambda$ often improve average continuing-model performance relative to full integration, particularly after longer same-distribution sequences.
In continual-pretraining experiments with both controlled streams and naturally defined adaptation sequences, task-vector scaling outperforms full integration at the matched learning rate, showing that its benefits are not reproduced by learning-rate scaling alone.

%% file: figures/overview_task_vector.tex
\definecolor{overviewink}{HTML}{343432}%
\definecolor{overviewmuted}{HTML}{85817B}%
\definecolor{overviewborder}{HTML}{D8E3EA}%
\definecolor{overviewpanel}{HTML}{F3F7FA}%
\definecolor{overviewtaskone}{HTML}{C97555}%
\definecolor{overviewtasktwo}{HTML}{756A92}%
\definecolor{overviewtaskthree}{HTML}{789468}%
\begin{tikzpicture}[
    font=\small,
    card/.style={draw=overviewborder,fill=overviewpanel,rounded corners=5pt,
        line width=0.65pt},
    batch/.style={draw=#1!80!black,fill=#1!16,rounded corners=0.7pt,
        minimum width=0.18cm,minimum height=0.40cm,inner sep=0pt},
    localstep/.style={-{Latex[length=1.0mm,width=0.75mm]},line width=0.62pt},
    taskvector/.style={-{Latex[length=1.8mm,width=1.2mm]},line width=1.35pt},
    checkpoint/.style={circle,fill=overviewink,inner sep=1.6pt},
    endpoint/.style={circle,draw=#1!85!black,fill=white,line width=0.9pt,
        inner sep=1.9pt},
    contour/.style={draw=overviewborder!75,line width=0.5pt},
]
    \draw[card] (0.05,0.10) rectangle (4.55,3.18);
    \draw[card] (4.75,0.10) rectangle (13.45,3.18);

    \node[anchor=west,inner sep=0pt,font=\bfseries,text=overviewink] at (0.33,2.83)
        {Non-i.i.d. stream};

    \foreach \xleft/\xright/\taskname/\taskcolor in {
        0.30/1.20/Task 1/overviewtaskone,
        1.30/2.20/Task 2/overviewtasktwo,
        2.30/3.20/Task 3/overviewtaskthree,
        3.30/4.20/Task 2/overviewtasktwo}
    {
        \draw[draw=\taskcolor!75!black,line width=0.55pt,decorate,
            decoration={brace,amplitude=2.2pt}]
            ($(\xleft,1.83)+(0.04,0)$) --
            ($(\xright,1.83)+(-0.04,0)$);
        \node[font=\bfseries\scriptsize,text=\taskcolor!82!black]
            at ($(\xleft,1.91)!0.5!(\xright,1.91)+(0,0.25)$) {\taskname};
    }

    \foreach \x in {0.42,0.64,0.86,1.08}
        \node[batch=overviewtaskone] at (\x,1.47) {};
    \foreach \x in {1.42,1.64,1.86,2.08}
        \node[batch=overviewtasktwo] at (\x,1.47) {};
    \foreach \x in {2.42,2.64,2.86,3.08}
        \node[batch=overviewtaskthree] at (\x,1.47) {};
    \foreach \x in {3.42,3.64,3.86,4.08}
        \node[batch=overviewtasktwo] at (\x,1.47) {};
    \draw[-{Latex[length=1.7mm]},overviewmuted,line width=0.6pt]
        (0.30,0.90) -- (4.20,0.90)
        node[midway,below=1.8pt,font=\scriptsize,text=overviewmuted]
        {arrival order};

    \node[anchor=west,inner sep=0pt,font=\bfseries,text=overviewink] at (5.03,2.83)
        {Task-vector descent};

    \begin{scope}[shift={(0.30,0.18)}]

    \draw[contour] plot[smooth cycle,tension=0.68] coordinates {
        (5.55,1.17) (6.10,0.65) (7.15,0.40) (8.25,0.49)
        (9.28,0.34) (10.38,0.46) (11.55,0.63) (12.58,0.99)
        (12.98,1.45) (12.53,1.98) (11.40,2.29) (10.23,2.18)
        (9.25,2.39) (8.17,2.20) (7.04,2.31) (6.03,1.95)};
    \draw[contour] plot[smooth cycle,tension=0.70] coordinates {
        (6.22,1.18) (6.72,0.82) (7.58,0.62) (8.46,0.72)
        (9.28,0.59) (10.16,0.70) (11.02,0.82) (11.78,1.10)
        (11.99,1.47) (11.61,1.80) (10.76,1.96) (9.91,1.87)
        (9.25,2.01) (8.38,1.89) (7.53,1.96) (6.69,1.70)};
    \draw[contour] plot[smooth cycle,tension=0.72] coordinates {
        (7.18,1.25) (7.68,1.00) (8.32,0.91) (8.88,1.01)
        (9.34,0.88) (9.90,1.00) (10.43,1.14) (10.73,1.37)
        (10.58,1.58) (10.02,1.72) (9.47,1.64) (9.01,1.76)
        (8.42,1.65) (7.86,1.72) (7.36,1.54)};
    \node[font=\scriptsize,text=overviewmuted,fill=overviewpanel,
        inner sep=0.8pt] at (11.72,0.48) {average-loss contours};

    \coordinate (thetaZero) at (5.27,0.69);
    \coordinate (localOne) at (7.18,2.03);
    \coordinate (thetaOne) at ($(thetaZero)!0.56!(localOne)$);
    \coordinate (localTwo) at (9.15,0.52);
    \coordinate (thetaTwo) at ($(thetaOne)!0.58!(localTwo)$);
    \coordinate (localThree) at (12.02,2.07);
    \coordinate (thetaThree) at ($(thetaTwo)!0.58!(localThree)$);

    \draw[overviewtaskone!55,dashed,line width=0.65pt]
        (thetaZero) -- (localOne);
    \draw[overviewtasktwo!55,dashed,line width=0.65pt]
        (thetaOne) -- (localTwo);
    \draw[overviewtaskthree!55,dashed,line width=0.65pt]
        (thetaTwo) -- (localThree);

    \coordinate (aOne) at (5.48,1.00);
    \coordinate (aTwo) at (5.69,1.24);
    \coordinate (aThree) at (5.91,1.42);
    \coordinate (aFour) at (6.14,1.75);
    \coordinate (aFive) at (6.38,1.95);
    \coordinate (aSix) at (6.62,1.86);
    \coordinate (aSeven) at (6.88,2.08);
    \foreach \from/\to in {
        thetaZero/aOne,aOne/aTwo,aTwo/aThree,aThree/aFour,
        aFour/aFive,aFive/aSix,aSix/aSeven}
        \draw[localstep,draw=overviewtaskone!76] (\from) -- (\to);
    \draw[localstep,draw=overviewtaskone!76,shorten >=3pt]
        (aSeven) -- (localOne);

    \coordinate (bOne) at (6.55,1.15);
    \coordinate (bTwo) at (6.80,0.90);
    \coordinate (bThree) at (7.07,0.63);
    \coordinate (bFour) at (7.34,0.70);
    \coordinate (bFive) at (7.61,0.39);
    \coordinate (bSix) at (7.90,0.48);
    \coordinate (bSeven) at (8.20,0.25);
    \coordinate (bEight) at (8.52,0.36);
    \coordinate (bNine) at (8.84,0.31);
    \foreach \from/\to in {
        thetaOne/bOne,bOne/bTwo,bTwo/bThree,bThree/bFour,
        bFour/bFive,bFive/bSix,bSix/bSeven,bSeven/bEight,bEight/bNine}
        \draw[localstep,draw=overviewtasktwo!78] (\from) -- (\to);
    \draw[localstep,draw=overviewtasktwo!78,shorten >=3pt]
        (bNine) -- (localTwo);

    \coordinate (cOne) at (8.20,1.18);
    \coordinate (cTwo) at (8.45,1.42);
    \coordinate (cThree) at (8.72,1.33);
    \coordinate (cFour) at (9.00,1.65);
    \coordinate (cFive) at (9.29,1.83);
    \coordinate (cSix) at (9.59,1.76);
    \coordinate (cSeven) at (9.90,2.04);
    \coordinate (cEight) at (10.22,2.16);
    \coordinate (cNine) at (10.55,2.00);
    \coordinate (cTen) at (10.90,2.22);
    \coordinate (cEleven) at (11.28,2.10);
    \foreach \from/\to in {
        thetaTwo/cOne,cOne/cTwo,cTwo/cThree,cThree/cFour,cFour/cFive,
        cFive/cSix,cSix/cSeven,cSeven/cEight,cEight/cNine,cNine/cTen,
        cTen/cEleven}
        \draw[localstep,draw=overviewtaskthree!82] (\from) -- (\to);
    \draw[localstep,draw=overviewtaskthree!82,shorten >=3pt]
        (cEleven) -- (localThree);

    \draw[taskvector,draw=overviewtaskone] (thetaZero) -- (thetaOne)
        node[midway,below=1.8pt,fill=overviewpanel,inner sep=0.6pt,
        font=\scriptsize] {$\lambda\Delta_0$};
    \draw[taskvector,draw=overviewtasktwo] (thetaOne) -- (thetaTwo)
        node[midway,above=1.8pt,fill=overviewpanel,inner sep=0.6pt,
         font=\scriptsize] {$\lambda\Delta_1$};
    \draw[taskvector,draw=overviewtaskthree] (thetaTwo) -- (thetaThree)
        node[midway,below=1.8pt,fill=overviewpanel,inner sep=0.6pt,
        font=\scriptsize] {$\lambda\Delta_2$};

    \node[checkpoint,label={[font=\scriptsize,text=overviewink,inner sep=0pt]below left:$\theta_0$}]
        at (thetaZero) {};
    \node[checkpoint,label={[font=\scriptsize,text=overviewink,inner sep=0pt]above:$\theta_1$}]
        at (thetaOne) {};
    \node[checkpoint,label={[font=\scriptsize,text=overviewink,inner sep=0pt]below:$\theta_2$}]
        at (thetaTwo) {};
    \node[checkpoint,label={[font=\scriptsize,text=overviewink,inner sep=0pt]above:$\theta_3$}]
        at (thetaThree) {};
    \node[endpoint=overviewtaskone] at (localOne) {};
    \node[endpoint=overviewtasktwo] at (localTwo) {};
    \node[endpoint=overviewtaskthree] at (localThree) {};

    \node[font=\bfseries\scriptsize,text=overviewtaskone!85!black,
        fill=overviewpanel,inner sep=0.8pt] at (6.00,2.08) {Task 1};
    \node[font=\bfseries\scriptsize,text=overviewtasktwo!85!black,
        fill=overviewpanel,inner sep=0.8pt] at (7.12,0.27) {Task 2};
    \node[font=\bfseries\scriptsize,text=overviewtaskthree!85!black,
        fill=overviewpanel,inner sep=0.8pt] at (9.95,2.48) {Task 3};
    \end{scope}
\end{tikzpicture}%

%% file: sections/01_introduction.tex
\section{Introduction}
\label{sec:introduction}

\looseness -1 A long-term goal for language-model agents is to become more capable through experience, i.e., to learn from interactions with users, adapt to new tasks, and carry what they learn into future interactions.
Achieving this requires accumulating knowledge and capabilities over time without losing those acquired earlier.
The data available for such adaptation comes from multiple domains, sources, users, tasks, or reinforcement-learning environments~\citep{udandarao2024practitioners}.
These distributions are encountered unevenly over time, with samples from the same distribution often becoming available close together~\citep{cai2021online,jones2023learning}, while staged post-training exposes models to successive tasks and objectives, such as supervised fine-tuning followed by preference optimization or reinforcement learning.
In conventional offline training, data collected across distributions can be pooled and shuffled to approximate \iid sampling from the overall mixture~\citep{mayo2023multitask}.
We instead consider an online continual learner that updates when new data become available rather than waiting for a globally mixed dataset~\citep{hu2023firehose,cai2021online}.
Such a learner, therefore, encounters a temporally clustered, non-\iid stream and must maintain performance across all distributions represented in that stream.
A further motivation comes from test-time training, where adaptation can improve performance on a target task~\citep{huebotter2025learning}.
Extending such adaptation to successive tasks raises the question of how much of each update to retain.

When consecutive minibatches come from the same distribution, standard gradient-based training progressively adapts the model to that targeted distribution.
Thus, longer sequences can produce stronger adaptation and a larger displacement from the model at the start of the sequence.
We refer to this displacement as a \emph{task vector}~\citep{ilharco2023editing}.
Standard continual training uses the model obtained at the end of each sequence as the starting point for the next, retaining the entire parameter displacement.
For long sequences, full integration risks overadapting the model to the targeted distribution and degrading performance on non-targeted distributions~\citep{mayo2023multitask}.
As this interference accumulates across the stream, it can lead to catastrophic forgetting~\citep{mccloskey1989catastrophic,goodfellow2014empirical}.

\textit{How can a language model benefit from sustained adaptation to one distribution while retaining capabilities needed on others?}
We study whether full integration remains the most effective update as the minibatch stream moves further from \iid sampling, or whether the task vector should be partially integrated instead.
Our contribution is a controlled empirical study of how temporal clustering affects partial integration across pretraining and post-training settings. Holding the selected training inputs fixed, we vary the length of same-distribution sequences and the integration coefficient~$\lambda$, and examine their effects on targeted adaptation and continuing-model performance.
Specifically, we study a linear family of update rules that moves the continuing model a fraction~$\lambda$ along the task vector and applies the same coefficient to the optimizer state. 
Full integration ($\lambda=1$) recovers standard continual training, whereas partial integration ($0<\lambda<1$) retains only a fraction of the task vector.

This formulation motivates the following \textbf{research questions}.
\begin{itemize}
\item \textbf{RQ1:} Across adaptation objectives and lengths of same-distribution sequences, when does task-vector scaling improve average continuing-model performance relative to full integration, and does it outperform full integration at the matched learning rate $\lambda\eta$?
\item \textbf{RQ2:} How does $\lambda$ trade targeted-distribution adaptation against non-targeted retention?
\item \textbf{RQ3:} Does task-vector scaling improve continuing-model performance during pretraining from random initialization, when no shared pretrained representation is available?
\item \textbf{RQ4:} Does task-vector scaling remain effective when adaptation sequences follow natural data boundaries and vary in length?
\end{itemize}

\looseness -1 Across continual pretraining, pretraining from random initialization, persona adaptation with on-policy self-distillation, supervised fine-tuning~(SFT), and Group Relative Policy Optimization~(GRPO), intermediate values of $\lambda$ improve the continuing-model performance compared with full integration, particularly after long same-distribution sequences. In continual pretraining, task-vector scaling also outperforms matched learning-rate scaling. The coefficient $\lambda$ trades adaptation to the targeted distribution \emph{(plasticity)} against retention on non-targeted distributions \emph{(stability)}. Improvements during pretraining from random initialization show that this effect does not require a shared pretrained representation. 
These benefits also extend to book-level continual pretraining with naturally defined, variable-length adaptation sequences.
Finally, in our experiments, interpolation remains well-behaved even after extended supervised fine-tuning, whereas pretraining-scale adaptation produces severe performance barriers.

%% file: sections/02_related_work.tex
\section{Related work}
\label{sec:related-work}

\noindent\textbf{Continual learning on non-\iid data}\hspace{0.5em}
Continual learning aims to learn from sequentially arriving data without forgetting earlier capabilities. 
Sequential learning is also studied through regret minimization in online convex optimization~\citep{zinkevich2003online,hazan2016introduction} and lifelong bandit optimization~\citep{schur2023lifelong}.
These works analyze cumulative performance on encountered tasks, while continual learning additionally emphasizes retaining performance on earlier tasks.
Long sequences of gradient updates on one distribution can overadapt the model and cause catastrophic forgetting \citep{mccloskey1989catastrophic,goodfellow2014empirical,delange2022continual}. Prior work addresses forgetting through replay, regularization, and task-specific parameter isolation \citep{delange2022continual} and studies forgetting in pretrained representations, language-model tuning, and continual pretraining \citep{luo2023investigating,li2024revisiting,udandarao2024practitioners}. Complementing this line of work, we study the effect of temporal clustering on continual training by holding the samples fixed across conditions and varying only their order, thereby changing how many consecutive minibatches come from the same distribution.

\noindent\textbf{Model merging, task arithmetic, and mode connectivity}\hspace{0.5em}
Model merging combines checkpoints adapted separately to different tasks or domains: Fisher-weighted averaging accounts for parameter importance, task arithmetic represents fine-tuning as a displacement from the shared initialization, and TIES-merging addresses interference between task-specific updates \citep{matena2022merging,ilharco2023editing,yadav2023ties}.
Even when two checkpoints perform well individually, models along the straight line between their parameters may perform poorly. Mode-connectivity work shows that such checkpoints can often be connected by curved paths along which the loss remains low \citep{garipov2018loss,draxler2018essentially}.
Linear connectivity is more likely when checkpoints share optimization history or parameter symmetries are aligned \citep{frankle2020linear,vlaar2022linear,entezari2022role,ainsworth2023git}. 
Because each locally adapted endpoint descends from the corresponding continuing-model parameters, these results motivate task-vector integration while also raising the question of when it ceases to be effective.

\noindent\textbf{Continual model merging}\hspace{0.5em}
\looseness-1Lookahead performs several inner-optimizer steps on fast weights before moving slow weights partway toward them, yielding the same parameter-interpolation form studied here. It also evaluates retaining, interpolating, or resetting SGD momentum \citep{zhang2019lookahead}.
Sequential Fine-tuning with Averaging (SFA) periodically averages the adapted model with the checkpoint preceding the current task and continues from the merged model \citep{kleiman2025soup}. AIMMerging determines when to merge using parameter displacement and replay loss, and combines task vectors obtained from task-specific and replay data \citep{feng2025aimmerging}.
Perturb-and-Merge uses a dynamic merging coefficient derived from empirical Fisher information and a curvature-regularized adaptation objective \citep{qiu2025train}.
We study how the benefits of partial integration depend on temporal clustering in language-model training, varying the length of same-distribution sequences while holding the selected training inputs fixed.
An extended discussion of related work is provided in~\cref{app:extended-related-work}.

%% file: sections/03_method.tex
\section{Task-vector updates for non-\iid minibatch streams}
\label{sec:method}

\subsection{Task-vector updates}
We separate adaptation to the current distribution from the amount of parameter change carried into subsequent learning.
This allows the model to follow the local optimization trajectory at the original learning rate while controlling how far the continuing model moves toward the locally adapted endpoint.
At sequence boundary $t$, the continuing checkpoint is the pair
$(\theta_t,\omega_t)$, comprising the model parameters and optimizer state.
We call the model parameterized by $\theta_t$ the \emph{continuing model}.

\paragraph{Learning objective} Let $P_1,\ldots,P_K$ be fixed data distributions, each corresponding to a domain, user, task, or environment, and let $\mathcal{L}(\theta;z)$ denote the loss on a sample $z$.
Our goal is to minimize the loss averaged uniformly across these distributions
\begin{align}
    \min_\theta
    \mathbb{E}_{k}
    \mathbb{E}_{z\sim P_k}
    \left[
        \mathcal{L}(\theta;z)
    \right].
    \label{eq:average-objective}
\end{align}
For a given distribution $P_k$, training estimates its population loss from a minibatch $B$ containing \iid samples from $P_k$, using the empirical loss
$
    \mathcal{L}(\theta;B)
    =
    \frac{1}{|B|}
    \sum_{z\in B}
    \mathcal{L}(\theta;z).
    \label{eq:minibatch-loss}
$
We consider $L$ consecutive minibatches $B_1,\ldots,B_L$ drawn from the same distribution.

\paragraph{Minibatch updates}
For clarity, we illustrate local adaptation using SGD without momentum.
Starting from $\theta_t$, we process the next $L$ minibatches with learning rate $\eta$:
\begin{align}
    \theta_t^{(0)}
    =
    \theta_t,
    \qquad
    \theta_t^{(i)}
    &=
    \theta_t^{(i-1)}
    -
    \eta
    \nabla_\theta
    \mathcal{L}\left(\theta_t^{(i-1)};B_i\right),
    \qquad i=1,\ldots,L.
    \label{eq:local-adaptation}
\end{align}
We refer to these $L$ minibatch updates as \emph{local adaptation} and to~$\theta_t^{(L)}$ as the \emph{locally adapted endpoint}.
Our experiments use AdamW, which updates both the model parameters and optimizer state.
The task-vector definition below applies to the resulting parameter displacement regardless of the optimizer used for local adaptation.

\paragraph{Task-vector update}
The displacement produced by $L$ minibatch updates defines the task
vector~$\Delta_t$. We construct the next continuing-model parameters by taking a
step of size $\lambda$ along this vector:
\begin{align}
    \Delta_t
    =
    \theta_t^{(L)}-\theta^{(0)}_t,\qquad
    \theta_{t+1}
    =
    \theta_t+\lambda\Delta_t,
    \qquad \lambda\in[0,1].
    \label{eq:linear-integration}
\end{align}
We call the operation that forms $\theta_{t+1}$ from $\theta_t$ and $\Delta_t$ \emph{integration}. Full integration uses $\lambda=1$ and, together with full optimizer-state integration, recovers standard training with the chosen optimizer, while partial integration uses $0<\lambda<1$. Setting $\lambda=0$ discards the task vector. We refer to this family of scaled updates as \emph{Task Vector Descent}.

\paragraph{Optimizer-state integration}
Our experiments use AdamW \citep{loshchilov2019decoupled}. We interpolate the optimizer state alongside the parameters. For the first moment, this reduces the current sequence's contribution while preserving more of the inherited momentum. At $\lambda=0$, we restore the pre-sequence parameters and optimizer state; at $\lambda=1$, we recover standard training.
Let $\omega_t=(m_t,v_t,c_t)$ denote the optimizer state associated with $\theta_t$, comprising the first moment, second moment, and step count, and let $\omega_t^{(L)}=(m_t^{(L)},v_t^{(L)},c_t^{(L)})$ denote the state after the $L$ minibatch updates. We integrate these states using the same coefficient $\lambda$:
\begin{align}
    m_{t+1}
    =
    (1-\lambda)m_t+\lambda m_t^{(L)},
    \quad
    v_{t+1}
    =
    \left(
        (1-\lambda)\sqrt{v_t}
        +
        \lambda\sqrt{v_t^{(L)}}
    \right)^2.
    \label{eq:optimizer-state-integration}
\end{align}
\looseness=-1 The first moment and step count are interpolated linearly, while the second moment is interpolated in square-root space because AdamW’s adaptive normalization depends on the root-mean-square gradient scale rather than the squared scale.
Together, $(\theta_{t+1},\omega_{t+1})$ form the continuing checkpoint passed to the next sequence.
The complete training procedure, AdamW updates, and optimizer-state integration details are provided in~\cref{app:method}.

\paragraph{Understanding partial integration}
Our analysis clarifies why task-vector scaling differs from reducing the learning rate and when partial integration can improve on full integration. A second-order SGD analysis shows that both methods agree to first order, but task-vector scaling retains a larger contribution from second-order interactions within each adaptation sequence~(\cref{app:sec:lr-scaling-analysis,app:sec:cross_task_analysis}). A local quadratic model favors partial integration when the full update overshoots the average-loss minimum along the task vector~(\cref{app:sec:optimal-scaling-analysis}).

\subsection{Controlled stream construction}

For each distribution~$P_k$, we use disjoint training and evaluation data.
The controlled streams contain only training examples, while the evaluation
data are used solely to compute the scores defined below.

We construct controlled non-\iid streams by fixing the minibatches drawn from
each distribution and varying only their order. For a chosen $L$, we partition
the minibatches associated with each $P_k$ into sequences of $L$ consecutive
minibatches and shuffle the order of these sequences. Let $k_t$ index the distribution
supplying sequence $t$. We call $P_{k_t}$ the targeted distribution and all $P_k$ with
$k\ne k_t$ the non-targeted distributions. Minibatches are consumed without replacement and the
stream is balanced across distributions. After each sequence, we apply the task-vector
update defined in~\Cref{eq:linear-integration}.
For fixed distributions and sampling frequencies, $L$ controls temporal clustering and integration frequency: larger $L$ produces longer same-distribution sequences, whereas smaller $L$ brings the stream closer to \iid sampling. 
Even when $L=1$, however, each minibatch contains samples from only one distribution and is not fully \iid We thus construct a separate \iid reference that mixes distributions within each minibatch. 
We hold the samples and training budget fixed across conditions and vary only their order.
\cref{fig:controlled-streams} in the appendix illustrates this construction.

\subsection{Evaluation}

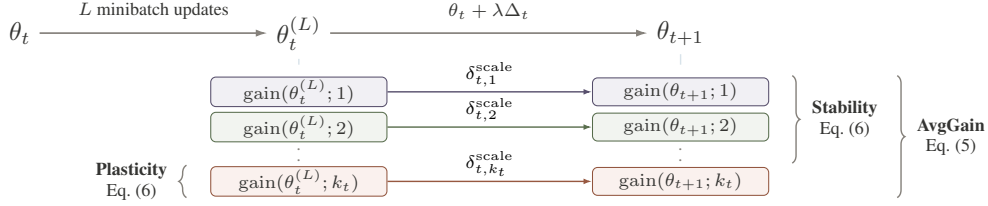
\begin{figure}[h]
    \vspace{-0.6em}
    \centering
    \noindent\makebox[\linewidth][c]{%
        \resizebox{0.94\linewidth}{!}{%
            \input{figures/evaluation_protocol}%
        }%
    }
    \vspace{-1.6em}
    \caption{Sequence-boundary evaluation. Plasticity uses the locally adapted endpoint; stability and AvgGain use the continuing model.}
    \label{fig:evaluation-protocol}
    \vspace{-1em}
\end{figure}

At each evaluated sequence boundary, we score both the locally adapted endpoint $\theta_t^{(L)}$ and the continuing model $\theta_{t+1}$ to distinguish local adaptation from performance retained for subsequent learning.
The endpoint $\theta_t^{(L)}$ measures performance on the targeted distribution after local adaptation, while $\theta_{t+1}$ measures performance carried forward after integration.
Comparing their scores isolates the immediate effect of scaling at that boundary, with both checkpoints sharing the same preceding training history.
These evaluations define continuing-model gain, stability and plasticity,
and the immediate scaling effect, as summarized in~\cref{fig:evaluation-protocol}.

Let $\operatorname{score}(\theta;k)$ denote the higher-is-better held-out score of the model with parameters $\theta$ on distribution $k$. For language modeling, we use negative bits per token (BPT) as the higher-is-better score. We define its gain relative
to the initial model $\theta_0$ as
$
    \operatorname{gain}(\theta;k)
    =
    \operatorname{score}(\theta;k)
    -
    \operatorname{score}(\theta_0;k).
$

\noindent\textbf{Average gain (AvgGain)}\hspace{0.5em}
We use average gain to measure improvement across all distributions after integration. 
Intuitively, it asks whether the model becomes better overall as it learns from successive distributions.
At sequence boundary $t$, the average gain of the continuing model after
integration is
\begin{align}
    \operatorname{AvgGain}_t
    =
    \frac{1}{K}
    \sum_{k=1}^{K}
    \operatorname{gain}(\theta_{t+1};k).
    \label{eq:general-gain}
\end{align}
\noindent\textbf{Stability and plasticity}\hspace{0.5em}
However, average gain can conceal a trade-off between adapting to the current distribution and maintaining performance elsewhere. At sequence boundary~$t$, we therefore measure plasticity as the targeted-distribution gain at the locally adapted endpoint and stability as the mean non-targeted gain after integration:
\begin{align}
    \operatorname{Plasticity}_t
    =\operatorname{gain}\left(\theta_t^{(L)};k_t\right),
    \qquad
    \operatorname{Stability}_t
    =
    \frac{1}{K-1}
    \sum_{k\ne k_t}
    \operatorname{gain}\left(\theta_{t+1};k\right).
    \label{eq:stability-plasticity}
\end{align}
\looseness-1We report these quantities averaged over evaluated sequence boundaries and seeds.
Although plasticity is measured before scaling, it still depends on $\lambda$ through the model produced by previous steps.

%% file: figures/evaluation_protocol.tex
\definecolor{evalink}{HTML}{343432}%
\definecolor{evalmuted}{HTML}{85817B}%
\definecolor{evalborder}{HTML}{D8E3EA}%
\definecolor{evalpanel}{HTML}{F3F7FA}%
\definecolor{evaltaskone}{HTML}{C97555}%
\definecolor{evaltasktwo}{HTML}{756A92}%
\definecolor{evaltaskthree}{HTML}{789468}%
\begin{tikzpicture}[
    font=\small,
    flow/.style={-{Latex[length=1.6mm,width=1.0mm]},draw=evalmuted,
        line width=0.65pt},
    checkpoint/.style={font=\normalsize,text=evalink},
    score/.style args={#1}{draw=#1!72!black,fill=#1!10,
        rounded corners=2pt,minimum width=2.40cm,minimum height=0.38cm,
        inner sep=1.0pt,font=\scriptsize,text=evalink},
    scorechange/.style args={#1}{-{Latex[length=1.15mm,width=0.8mm]},
        draw=#1!75!black,line width=0.55pt},
    metric/.style={font=\scriptsize,align=center,text=evalink},
    brace/.style={draw=evalmuted,line width=0.55pt,decorate,
        decoration={brace,amplitude=3pt}},
]
    \node[checkpoint] (start) at (0.05,3.30) {$\theta_t$};
    \node[checkpoint] (endpoint) at (3.85,3.30) {$\theta_t^{(L)}$};
    \node[checkpoint] (continuing) at (9.05,3.30) {$\theta_{t+1}$};

    \draw[flow] (start.east) -- (endpoint.west)
        node[midway,above=2pt,font=\scriptsize,text=evalink]
        {$L$ minibatch updates};
    \draw[flow] (endpoint.east) -- (continuing.west)
        node[midway,above=2pt,font=\scriptsize,text=evalink]
        {$\theta_t+\lambda\Delta_t$};

    \draw[draw=evalborder,line width=0.55pt]
        (endpoint.south) -- (3.85,2.87);
    \draw[draw=evalborder,line width=0.55pt]
        (continuing.south) -- (9.05,2.87);
    \node[score=evaltasktwo] (leftone) at (3.85,2.50)
        {$\operatorname{gain}(\theta_t^{(L)};1)$};
    \node[score=evaltaskthree] (lefttwo) at (3.85,2.02)
        {$\operatorname{gain}(\theta_t^{(L)};2)$};
    \node[font=\scriptsize,text=evalmuted] at (3.85,1.68) {$\vdots$};
    \node[score=evaltaskone] (leftactive) at (3.85,1.28)
        {$\operatorname{gain}(\theta_t^{(L)};k_t)$};

    \node[score=evaltasktwo] (rightone) at (9.05,2.50)
        {$\operatorname{gain}(\theta_{t+1};1)$};
    \node[score=evaltaskthree] (righttwo) at (9.05,2.02)
        {$\operatorname{gain}(\theta_{t+1};2)$};
    \node[font=\scriptsize,text=evalmuted] at (9.05,1.68) {$\vdots$};
    \node[score=evaltaskone] (rightactive) at (9.05,1.28)
        {$\operatorname{gain}(\theta_{t+1};k_t)$};

    \draw[brace] (2.32,1.06) -- (2.32,1.50);
    \node[metric,anchor=east] at (2.18,1.28)
        {\textbf{Plasticity}\\Eq.~\eqref{eq:stability-plasticity}};

    \draw[brace] (10.57,2.72) -- (10.57,1.53);
    \node[metric,anchor=west] at (10.72,2.13)
        {\textbf{Stability}\\Eq.~\eqref{eq:stability-plasticity}};

    \draw[brace] (12.00,2.72) -- (12.00,1.06);
    \node[metric,anchor=west] at (12.15,1.89)
        {\textbf{AvgGain}\\Eq.~\eqref{eq:general-gain}};

    \draw[scorechange=evaltasktwo] (leftone.east) -- (rightone.west)
        node[midway,above=1pt,font=\scriptsize,fill=white,inner sep=0.7pt]
        {$\delta^{\mathrm{scale}}_{t,1}$};
    \draw[scorechange=evaltaskthree] (lefttwo.east) -- (righttwo.west)
        node[midway,above=1pt,font=\scriptsize,fill=white,inner sep=0.7pt]
        {$\delta^{\mathrm{scale}}_{t,2}$};
    \draw[scorechange=evaltaskone] (leftactive.east) -- (rightactive.west)
        node[midway,above=1pt,font=\scriptsize,fill=white,inner sep=0.7pt]
        {$\delta^{\mathrm{scale}}_{t,k_t}$};

\end{tikzpicture}%

%% file: sections/04_experiments.tex
\section{Experimental setup and applications}
\label{sec:experiments}

\looseness-1 We choose settings covering the main stages of LLM training to test whether the benefits of partial integration extend across different initializations and learning objectives.
Complete training configurations, evaluation protocols, and interpolation setups are provided in~\cref{app:setup}.

\noindent\textbf{Common setup}\hspace{0.5em}
\looseness-1 Except for book-level pretraining, selected data are fixed across seeds; seeds control distribution order, training randomness, and stochastic generation. Each run uses a fixed integration coefficient $\lambda$. We average scores over examples, tasks, and evaluated integration steps, then report the mean and standard error of the mean (SEM) across seeds. A training unit is an optimizer minibatch for pretraining and SFT, a prompt--persona interaction for persona adaptation, or a four-prompt rollout batch for GRPO. \Cref{app:tab:experiments} in the appendix summarizes the experimental settings and metrics.

\noindent\textbf{Continual pretraining}\hspace{0.5em}
A continually improving language model must incorporate knowledge from newly available sources without losing its usefulness on others. 
We study this balance through continual pretraining on temporally clustered domains.
We continue training the released 124M-parameter GPT-2 model~\citep{radford2019language} on ten Pile Uncopyrighted domains~\citep{gao2022pile}, using document-disjoint training and evaluation splits.
We select the learning rate $\eta=4\times10^{-5}$ through a grid search on the \iid reference.
As a matched learning-rate baseline, we use full integration with learning rate $\lambda\eta$.
The primary score is BPT reduction relative to the initial checkpoint.

\noindent\textbf{Persona adaptation}\hspace{0.5em}
Persona adaptation tests whether a model can learn a user's preferences while retaining performance for other users.
We adapt a Qwen3-4B student to six style preferences using Qwen3-8B as simulator and judge \citep{yang2025qwen3,buening2026aligning}.
Each preference uses the same 32 HelpSteer prompts \citep{wang2023helpsteer}.
Rejected interactions provide simulator feedback and trigger Self-Distillation Policy Optimization (SDPO) updates \citep{hubotter2026reinforcement}, while accepted interactions are skipped.
SDPO uses the student's feedback-conditioned predictions to teach the same student without feedback.
The primary score is macro-average pairwise win rate against the initial student on 128 held-out prompts.

\noindent\textbf{Reasoning SFT and GRPO}\hspace{0.5em}
\looseness-1 Learning to solve new reasoning tasks should preserve the ability to solve previously encountered ones. We use SFT and GRPO to test whether partial integration improves this balance under supervised and reward-based learning objectives.
We adapt Qwen2.5-3B-Instruct \citep{yang2024qwen25} to Countdown, Mini-Sudoku, N-Queens, and String Manipulation from Reasoning Gym~\citep{stojanovski2025reasoninggym}.
SFT uses completion-only cross-entropy on 256 verified expert traces per task; GRPO uses verifiable rewards on 4,096 prompts per task~\citep{shao2024deepseekmath,deepseekai2025deepseekr1}.
We evaluate 100 prompts per task with four sampled responses, using Best@4 as the primary score.

\noindent\textbf{Pretraining from random initialization}\hspace{0.5em}
To test whether partial integration is effective without a pretrained representation, we train a randomly initialized 124M-parameter GPT-2-size decoder-only Transformer in BF16 on approximately 10.77B tokens from ten Pile Uncopyrighted domains \citep{gao2022pile}. Each effective minibatch contains 512 packed 1,028-token blocks. The \iid reference uses globally shuffled mixed-domain minibatches.

\noindent\textbf{Book-level continual pretraining}\hspace{0.5em}
\looseness-1 Books provide natural sequence boundaries for testing adaptation beyond our controlled streams.
We continually adapt SmolLM-360M~\citep{allal2024smollm} on 200 PG-19 books~\citep{rae2020compressive}, treating each book as one sequence and integrating after each book.
We compare partial integration with full integration at matched learning rates $\lambda\eta$, using identical training manifests within each of three paired seeds.
The primary score is BPT reduction relative to initialization on 50 fixed validation books, averaged equally over books and the 200 integration boundaries.

\noindent\textbf{Controlled long-SFT interpolation}\hspace{0.5em}
To test interpolation after extended adaptation, we interpolate SmolLM2-1.7B-Instruct with checkpoints along a MetaMathQA SFT trajectory and evaluate GSM8K accuracy.
We also examine pretraining-scale interpolation between Qwen2.5 base models and their Qwen2.5-Coder descendants.

%% file: sections/05_results.tex
\section{Results}
\label{sec:results}

\subsection{Intermediate task-vector scaling improves the continuing model}
\label{sec:results-general}

To address \textbf{RQ1}, we test whether partial integration improves performance across distributions and whether its benefits can be explained by smaller optimization steps.
We measure the cumulative effect of the update rule through
$\operatorname{AvgGain}_t$~(\cref{eq:general-gain}), averaged over the
evaluated sequence boundaries.
This captures both the immediate effect of each integration and its influence on the continuing checkpoints inherited by later sequences.

\paragraph{Continual pretraining}
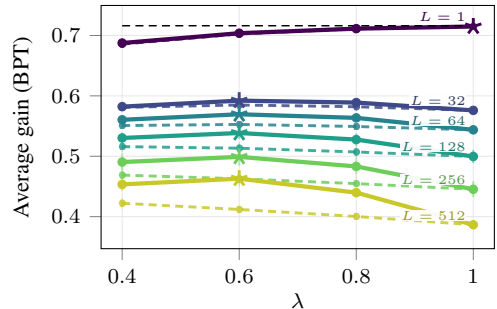
\begin{wrapfigure}{r}{0.49\textwidth}
    \vspace{-1.4\baselineskip}
    \centering
    \makebox[\linewidth][r]{%
        \input{figures/results/observation_general_gain_continual}}
    \vspace{-1.7\baselineskip}
    \caption{Continual-pretraining gain. Solid curves show task-vector scaling; dashed curves show full integration at learning rate $\lambda\eta$. 
    Stars mark the best task-vector coefficient, error bars
    show SEM, and the black dashed line denotes the \iid reference.}
    \label{fig:general-gain-continual}
    \vspace{-1\baselineskip}
\end{wrapfigure}

\looseness -1 Task-vector scaling improves average gain, with the benefits becoming more
pronounced for longer same-distribution sequences
(\cref{fig:general-gain-continual}). For every evaluated $L\geq32$, an
intermediate coefficient performs best, and task-vector scaling consistently
outperforms matched learning-rate scaling for every $\lambda<1$. At $L=512$,
for example, $\lambda=0.6$ increases the BPT gain from $0.386$ under full
integration to $0.463$, whereas full integration at the matched learning rate
$0.6\eta$ reaches only $0.412$.
This is consistent with the distinction between task-vector scaling and learning-rate scaling discussed in~\cref{sec:method}.
For $L=1$, by contrast, the two scaling methods
perform nearly identically, and full integration performs best and nearly matches the \iid reference.

\begin{figure*}[h]
    \raggedleft
    \input{figures/results/observation_general_gain_posttraining}\hspace*{-0.9mm}
    \caption{Continuing-model gain for SDPO, SFT, and GRPO (left to right) as a function of $\lambda$. Curves denote $L$; error bars show SEM across seeds, and stars mark the best $\lambda$ for each $L$. Dashed lines are \iid or joint-data references evaluated on the smallest-$L$ schedule in each setting.}
    \label{fig:general-gain-posttraining}
\end{figure*}
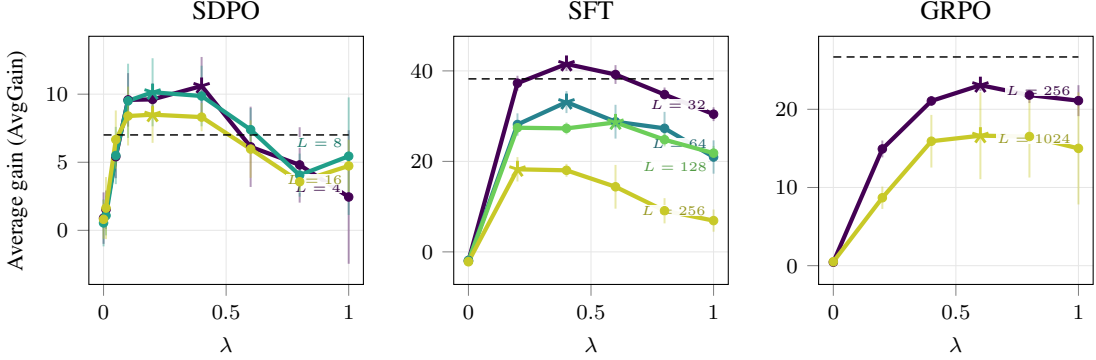

\paragraph{Post-training}
We next examine whether the benefit extends beyond next-token pretraining to supervised and feedback-based adaptation.
Across SDPO, SFT, and GRPO, intermediate coefficients outperform $\lambda=1$ for every evaluated sequence length~(\cref{fig:general-gain-posttraining}).
For reasoning
SFT with $L=32$, $\lambda=0.4$ increases the Best@4 gain from $30.43$ to
$41.53$ percentage points. Both evaluated GRPO sequence lengths attain their
largest average gain at $\lambda=0.6$, whereas the best coefficient for SDPO
is $\lambda=0.4$ at $L=4$ and $\lambda=0.2$ at $L\in\{8,16\}$.

Overall, intermediate task-vector scaling is most consistently beneficial
after longer sequences from the same distribution, but no single value of
$\lambda$ is optimal across adaptation regimes and sequence lengths.
Comparisons of $\lambda$ within a fixed $L$ are direct; absolute results
across different values of $L$ are not directly comparable because they are
evaluated at different sequence boundaries and frequencies.

\subsection{Task-vector scaling controls the stability--plasticity trade-off}
\label{sec:results-tradeoff}

\looseness-1 Improved average performance can conceal different effects on adaptation and retention.
To address \textbf{RQ2}, we examine how task-vector scaling affects learning from the targeted distribution and retaining performance on non-targeted distributions.
Separating the locally adapted
endpoint from the continuing model is particularly useful in settings such
as test-time training: task-specific inference can use the locally adapted endpoint,
while only a scaled task vector is retained by the continuing model. We
therefore use the stability and plasticity measures defined in
\cref{eq:stability-plasticity} and illustrated in
\cref{fig:evaluation-protocol}.

\paragraph{Continual pretraining}
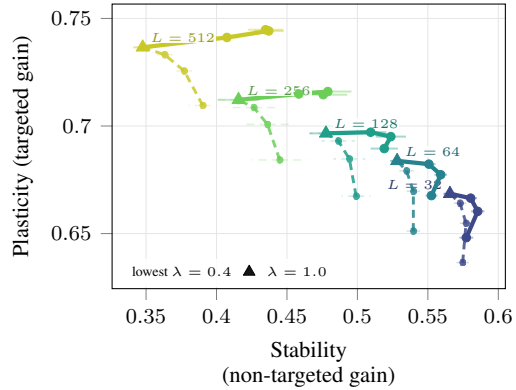
\begin{wrapfigure}{r}{0.49\textwidth}
    \vspace{-0.8\baselineskip}
    \centering
    \makebox[\linewidth][r]{%
        \input{figures/results/observation_nontargeted_targeted_continual}}
    \caption{Stability--plasticity trade-off in continual pretraining. Solid
    curves show task-vector scaling and dashed curves show full integration at
    learning rate $\lambda\eta$; error bars show SEM.}
    \label{fig:nontargeted-targeted-continual}
    \vspace{-1.5em}
\end{wrapfigure}

Within task-vector scaling, increasing $\lambda$ generally raises plasticity
at the cost of stability for $L\in\{32,64,128\}$
(\cref{fig:nontargeted-targeted-continual}). At $L=256$, the frontier begins to
flatten, and at $L=512$, smaller coefficients improve both coordinates.
Task-vector scaling also achieves higher stability and plasticity than matched learning-rate scaling for every sequence length shown, with the gap increasing for longer sequences. 
At $L=512$, reducing the learning rate recovers some stability by sacrificing plasticity, whereas partial
integration improves both relative to full integration at $\eta$. Because
plasticity is measured before the current task vector is
scaled, its improvement within task-vector scaling reflects the cumulative
effect of $\lambda$ on checkpoints inherited from earlier sequences.

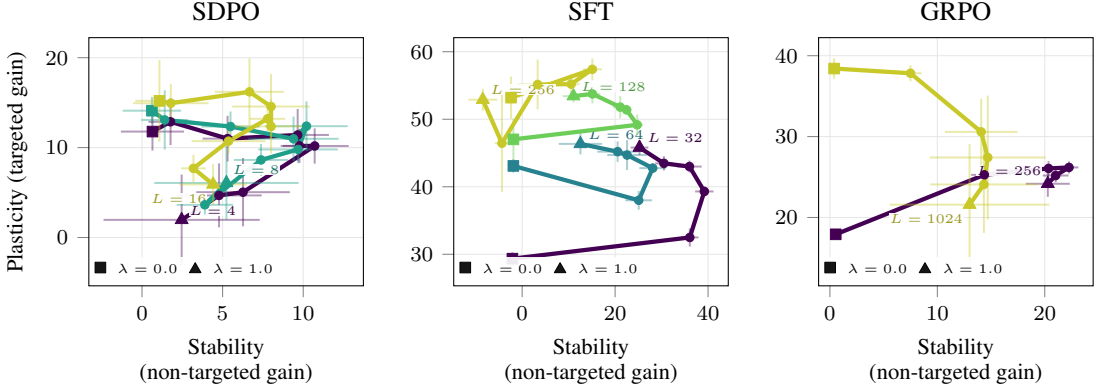
\begin{figure*}[h]
    \vspace{-0.5em}
    \raggedleft
    \input{figures/results/observation_nontargeted_targeted_posttraining}\hspace*{-0.9mm}
    \caption{Non-targeted retention versus targeted adaptation for SDPO, SFT,
    and GRPO from left to right across task-vector scaling coefficients. The
    stability axis is the mean continuing-model gain over non-targeted
    distributions or contexts; the plasticity axis is targeted gain at the
    endpoint $\theta_t^{(L)}$.}
    \label{fig:nontargeted-targeted-posttraining}
\end{figure*}

\paragraph{Post-training}
The post-training settings show the same regime-dependent transition
(\cref{fig:nontargeted-targeted-posttraining}). For shorter SFT and GRPO
sequences, larger coefficients generally improve plasticity, while
intermediate coefficients preserve more stability. With longer sequences,
the locally adapted endpoint already reflects substantial within-sequence adaptation; stronger
integration then reduces stability and can also impair plasticity at later
sequences. This reversal is clearest for SFT with $L=256$ and GRPO with
$L=1024$. SDPO is less monotonic: because its data are collected on-policy,
some intermediate coefficients improve targeted and non-targeted style scores
simultaneously.

Overall, $\lambda$ has both immediate and cumulative effects. 
It directly trades targeted-distribution adaptation against non-targeted retention at the current sequence boundary.
By changing the checkpoint used to initialize subsequent sequences, it also affects their later adaptation and retention, particularly when same-distribution sequences are long.
To isolate the immediate effect of task-vector scaling, we compare performance before and after integration.
We find that scaling recovers substantial non-targeted performance in SFT but little in GRPO, consistent with reduced forgetting under on-policy methods reported by \emph{RL's Razor}~\citep{shenfeld2026rl}.
Further analysis is provided in~\cref{app:secondary-results}.

\subsection{Task-vector scaling improves pretraining from random initialization}
\label{sec:results-scratch}

The preceding experiments leave open whether the benefits of partial integration depend on preserving a useful pretrained representation.
To address \textbf{RQ3}, we test task-vector scaling during pretraining from random initialization.

\begin{figure*}[h]
    \centering
    \input{figures/results/pretraining_scratch_outcomes}
    \caption{\looseness-1 Pre-training from random initialization. The left panel reports continuing-model macro BPT gain as a function of the task-vector scaling coefficient.
    Stars mark the best coefficient on each curve, and the dashed line is the shuffled \iid reference. 
    The right panel compares plasticity and stability.}
    \label{fig:scratch-outcomes}
\end{figure*}
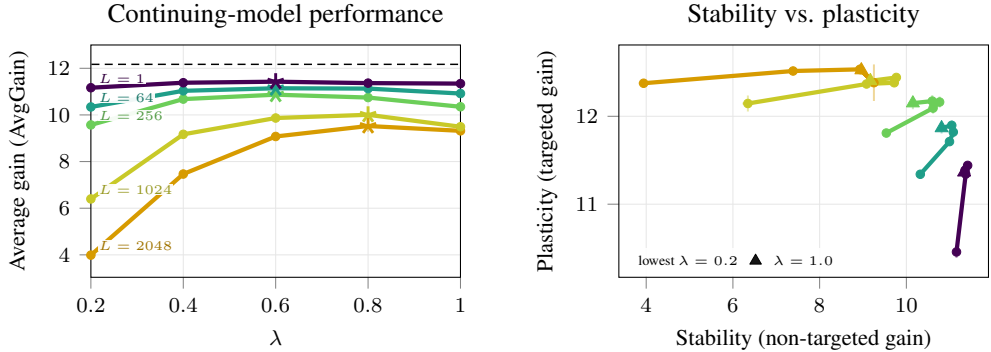

\Cref{fig:scratch-outcomes} shows that an intermediate
coefficient improves average continuing-model gain relative to full integration
for every evaluated $L$. The best coefficient is $\lambda=0.6$ for
$L\in\{1,64,256\}$ and $\lambda=0.8$ for $L\in\{1024,2048\}$.
Thus, the benefit of task-vector scaling does not depend on starting from a
shared pretrained representation.

The right panel shows that the improvement at the preferred coefficients is
driven primarily by higher non-targeted gain, while targeted-distribution gain
changes comparatively little. The shuffled \iid condition remains a reference and
performs better than all non-\iid configurations.
Additional analysis of the immediate effect of scaling is provided in~\cref{app:secondary-results}.

\subsection{Task-vector scaling improves adaptation with natural sequence boundaries}
\label{sec:results-pg19}

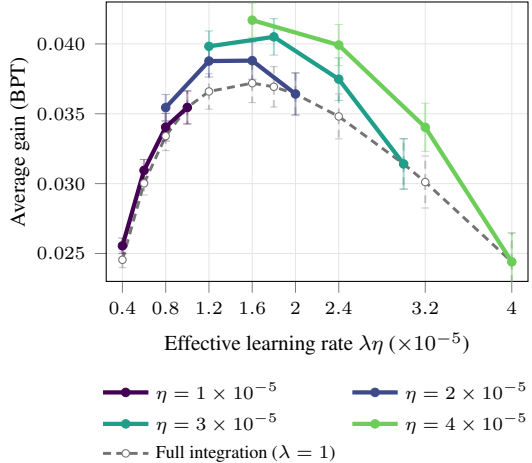
\begin{wrapfigure}{r}{0.49\textwidth}
    \vspace{-0.8\baselineskip}
    \centering
    \makebox[\linewidth][r]{\input{figures/results/pg19_general_gain}}
    \caption{PG-19 BPT gain relative to initialization.
    Colors denote base learning rates $\eta$; points show
    $\lambda\in\{0.4,0.6,0.8,1\}$. Gray dashed: full integration
    at matched effective rates. Error bars: $\pm1$ SEM
    over three paired seeds.}
    \label{fig:pg19-general-gain}
    \vspace{-1\baselineskip}
\end{wrapfigure}
The previous experiments use equal-length adaptation sequences drawn from a fixed set of recurring tasks. To address \textbf{RQ4}, we continually pretrain on a sequence of books, integrating the task vector after training on each book. Book boundaries define the adaptation sequences, whose lengths vary with the number of training minibatches per book.
The distribution of token counts per book is reported in~\cref{app:pg19-books}.

\looseness-1\Cref{fig:pg19-general-gain} shows that partial integration outperforms full
integration at the matched learning rate $\lambda\eta$ in all 12 evaluated
comparisons with $\lambda<1$.
Using $\eta=4\times10^{-5}$ and $\lambda=0.4$ yields 12.1\% greater average BPT gain than full integration at its best evaluated learning rate, $\eta=1.6\times10^{-5}$.
The preferred scaling coefficient depends on the base learning rate: at $\eta=10^{-5}$, below the best full-integration rate, partial integration reduces gain, whereas it improves gain at the three higher base learning rates.
This pattern is consistent with our local analysis, which favors partial integration when the full update overshoots the minimum along the task-vector direction~(\cref{app:sec:optimal-scaling-analysis}).

\subsection{Linear interpolation remains effective through long SFT}
\label{sec:results-connectivity}

To probe the limits of task-vector scaling, we test whether linear interpolation remains effective after extended adaptation.
We find no pronounced performance barriers along a controlled SFT trajectory, but severe barriers after pretraining-scale adaptation.
Experimental details and interpolation curves are provided in~\cref{app:long-sft-results,app:coder}.

%% file: figures/results/observation_general_gain_continual.tex
\definecolor{runL1}{HTML}{440154}
\definecolor{runL4}{HTML}{482878}
\definecolor{runL8}{HTML}{3E4989}
\definecolor{runL16}{HTML}{31688E}
\definecolor{runL32}{HTML}{26828E}
\definecolor{runL64}{HTML}{1F9E89}
\definecolor{runL128}{HTML}{35B779}
\definecolor{runL256}{HTML}{6CCE59}
\definecolor{runL512}{HTML}{A0DA39}
\definecolor{runL1024}{HTML}{C8C928}
\definecolor{runL2048}{HTML}{D79B00}
\definecolor{runscratchdeltaL64}{HTML}{1F9E89}
\definecolor{runscratchdeltaL256}{HTML}{6CCE59}
\definecolor{runscratchdeltaL1024}{HTML}{C8C928}
\definecolor{runscratchdeltaL2048}{HTML}{D79B00}
\definecolor{runpileL1}{HTML}{440154}
\definecolor{runpileL32}{HTML}{3E4989}
\definecolor{runpileL64}{HTML}{26828E}
\definecolor{runpileL128}{HTML}{1F9E89}
\definecolor{runpileL256}{HTML}{6CCE59}
\definecolor{runpileL512}{HTML}{C8C928}
\definecolor{runpersonaL4}{HTML}{440154}
\definecolor{runpersonaL8}{HTML}{1F9E89}
\definecolor{runpersonaL16}{HTML}{C8C928}
\definecolor{runsftL32}{HTML}{440154}
\definecolor{runsftL64}{HTML}{26828E}
\definecolor{runsftL128}{HTML}{6CCE59}
\definecolor{runsftL256}{HTML}{C8C928}
\definecolor{runrlvrL256}{HTML}{440154}
\definecolor{runrlvrL1024}{HTML}{C8C928}
\begin{tikzpicture}[trim axis group left,trim axis group right]
\begin{groupplot}[
  group style={group size=1 by 1, horizontal sep=0pt, vertical sep=1.0em},
  width=0.76\linewidth,
  height=0.47\linewidth,
  scale only axis,
  title style={font=\fontsize{10}{11}\selectfont,align=center,yshift=-0.7ex},
  label style={font=\fontsize{8.5}{9.5}\selectfont},
  tick label style={font=\fontsize{8}{9}\selectfont,/pgf/number format/fixed,/pgf/number format/precision=2},
  tick align=outside,
  tick pos=left,
  axis line style={line width=0.4pt},
  major grid style={gray!20,line width=0.3pt},
  grid=major,
  enlarge x limits=0.06,
  enlarge y limits=0.10,
  scaled ticks=false,
  every axis plot/.append style={line join=round},
]
\nextgroupplot[xlabel={$\lambda$},xlabel style={yshift=3pt},ylabel={Average gain (BPT)}]
\addplot[color=runpileL1,densely dashed,line width=1.1pt,opacity=0.85,mark=*,mark size=1.25pt,mark options={line width=0pt},error bars/error bar style={draw=runpileL1,opacity=0.22,draw opacity=0.22,line width=0.55pt},error bars/error mark=none,error bars/.cd,y dir=both,y explicit] coordinates {(0.4,0.6856558) +- (0,0.0005083) (0.6,0.7025570) +- (0,0.0004416) (0.8,0.7111416) +- (0,0.0004153) (1,0.7148264) +- (0,0.0006205)};
\addplot[color=runpileL1,line width=1.6pt,mark=none,error bars/error bar style={draw=runpileL1,opacity=0.4,draw opacity=0.4,fill opacity=0.4,line width=0.75pt},error bars/error mark=none,error bars/.cd,y dir=both,y explicit] coordinates {(0.4,0.6874734) +- (0,0.0006447868) (0.6,0.7037503) +- (0,0.0003509229) (0.8,0.7113108) +- (0,0.0003293923) (1,0.7148564) +- (0,0.0006140359)};
\addplot[only marks,color=runpileL1,mark=*,mark size=1.6pt,mark options={line width=0pt}] coordinates {(0.4,0.6874734) (0.6,0.7037503) (0.8,0.7113108)};
\addplot[only marks,color=runpileL1,mark=star,mark size=3.2pt,line width=1.1pt] coordinates {(1,0.7148564)};
\node[font=\fontsize{5}{5.5}\selectfont,text=runpileL1!75!black,anchor=south east,xshift=-2.5pt,yshift=1.5pt,fill=white,fill opacity=0.94,text opacity=1,inner sep=0.25pt] at (axis cs:1,0.7148564) {$L=1$};
\addplot[color=runpileL32,densely dashed,line width=1.1pt,opacity=0.85,mark=*,mark size=1.25pt,mark options={line width=0pt},error bars/error bar style={draw=runpileL32,opacity=0.22,draw opacity=0.22,line width=0.55pt},error bars/error mark=none,error bars/.cd,y dir=both,y explicit] coordinates {(0.4,0.5808882) +- (0,0.0039011) (0.6,0.5848696) +- (0,0.0039158) (0.8,0.5819840) +- (0,0.0038870) (1,0.5759489) +- (0,0.0040563)};
\addplot[color=runpileL32,line width=1.6pt,mark=none,error bars/error bar style={draw=runpileL32,opacity=0.4,draw opacity=0.4,fill opacity=0.4,line width=0.75pt},error bars/error mark=none,error bars/.cd,y dir=both,y explicit] coordinates {(0.4,0.5821633) +- (0,0.004311285) (0.6,0.5920803) +- (0,0.00414656) (0.8,0.5889063) +- (0,0.004002608) (1,0.5759497) +- (0,0.004055347)};
\addplot[only marks,color=runpileL32,mark=*,mark size=1.6pt,mark options={line width=0pt}] coordinates {(0.4,0.5821633) (0.8,0.5889063) (1,0.5759497)};
\addplot[only marks,color=runpileL32,mark=star,mark size=3.2pt,line width=1.1pt] coordinates {(0.6,0.5920803)};
\node[font=\fontsize{5}{5.5}\selectfont,text=runpileL32!75!black,anchor=south east,xshift=-2.5pt,yshift=1.5pt,fill=white,fill opacity=0.94,text opacity=1,inner sep=0.25pt] at (axis cs:1,0.5759497) {$L=32$};
\addplot[color=runpileL64,densely dashed,line width=1.1pt,opacity=0.85,mark=*,mark size=1.25pt,mark options={line width=0pt},error bars/error bar style={draw=runpileL64,opacity=0.22,draw opacity=0.22,line width=0.55pt},error bars/error mark=none,error bars/.cd,y dir=both,y explicit] coordinates {(0.4,0.5508272) +- (0,0.0043465) (0.6,0.5527080) +- (0,0.0042829) (0.8,0.5493060) +- (0,0.0045883) (1,0.5437450) +- (0,0.0043934)};
\addplot[color=runpileL64,line width=1.6pt,mark=none,error bars/error bar style={draw=runpileL64,opacity=0.4,draw opacity=0.4,fill opacity=0.4,line width=0.75pt},error bars/error mark=none,error bars/.cd,y dir=both,y explicit] coordinates {(0.4,0.5603833) +- (0,0.003804927) (0.6,0.5695018) +- (0,0.00409041) (0.8,0.5635652) +- (0,0.004247908) (1,0.5437449) +- (0,0.004393417)};
\addplot[only marks,color=runpileL64,mark=*,mark size=1.6pt,mark options={line width=0pt}] coordinates {(0.4,0.5603833) (0.8,0.5635652) (1,0.5437449)};
\addplot[only marks,color=runpileL64,mark=star,mark size=3.2pt,line width=1.1pt] coordinates {(0.6,0.5695018)};
\node[font=\fontsize{5}{5.5}\selectfont,text=runpileL64!75!black,anchor=south east,xshift=-2.5pt,yshift=1.5pt,fill=white,fill opacity=0.94,text opacity=1,inner sep=0.25pt] at (axis cs:1,0.5437449) {$L=64$};
\addplot[color=runpileL128,densely dashed,line width=1.1pt,opacity=0.85,mark=*,mark size=1.25pt,mark options={line width=0pt},error bars/error bar style={draw=runpileL128,opacity=0.22,draw opacity=0.22,line width=0.55pt},error bars/error mark=none,error bars/.cd,y dir=both,y explicit] coordinates {(0.4,0.5159388) +- (0,0.0091830) (0.6,0.5133287) +- (0,0.0101020) (0.8,0.5070995) +- (0,0.0107357) (1,0.4995646) +- (0,0.0105589)};
\addplot[color=runpileL128,line width=1.6pt,mark=none,error bars/error bar style={draw=runpileL128,opacity=0.4,draw opacity=0.4,fill opacity=0.4,line width=0.75pt},error bars/error mark=none,error bars/.cd,y dir=both,y explicit] coordinates {(0.4,0.5302711) +- (0,0.008645337) (0.6,0.5385054) +- (0,0.009690308) (0.8,0.5276133) +- (0,0.01008741) (1,0.499567) +- (0,0.01055847)};
\addplot[only marks,color=runpileL128,mark=*,mark size=1.6pt,mark options={line width=0pt}] coordinates {(0.4,0.5302711) (0.8,0.5276133) (1,0.499567)};
\addplot[only marks,color=runpileL128,mark=star,mark size=3.2pt,line width=1.1pt] coordinates {(0.6,0.5385054)};
\node[font=\fontsize{5}{5.5}\selectfont,text=runpileL128!75!black,anchor=south east,xshift=-2.5pt,yshift=1.5pt,fill=white,fill opacity=0.94,text opacity=1,inner sep=0.25pt] at (axis cs:1,0.499567) {$L=128$};
\addplot[color=runpileL256,densely dashed,line width=1.1pt,opacity=0.85,mark=*,mark size=1.25pt,mark options={line width=0pt},error bars/error bar style={draw=runpileL256,opacity=0.22,draw opacity=0.22,line width=0.55pt},error bars/error mark=none,error bars/.cd,y dir=both,y explicit] coordinates {(0.4,0.4688083) +- (0,0.0173453) (0.6,0.4626656) +- (0,0.0155481) (0.8,0.4547126) +- (0,0.0138070) (1,0.4452819) +- (0,0.0130819)};
\addplot[color=runpileL256,line width=1.6pt,mark=none,error bars/error bar style={draw=runpileL256,opacity=0.4,draw opacity=0.4,fill opacity=0.4,line width=0.75pt},error bars/error mark=none,error bars/.cd,y dir=both,y explicit] coordinates {(0.4,0.490417) +- (0,0.01454194) (0.6,0.4989846) +- (0,0.01473665) (0.8,0.4831744) +- (0,0.01342645) (1,0.4452805) +- (0,0.01308267)};
\addplot[only marks,color=runpileL256,mark=*,mark size=1.6pt,mark options={line width=0pt}] coordinates {(0.4,0.490417) (0.8,0.4831744) (1,0.4452805)};
\addplot[only marks,color=runpileL256,mark=star,mark size=3.2pt,line width=1.1pt] coordinates {(0.6,0.4989846)};
\node[font=\fontsize{5}{5.5}\selectfont,text=runpileL256!75!black,anchor=south east,xshift=-2.5pt,yshift=1.5pt,fill=white,fill opacity=0.94,text opacity=1,inner sep=0.25pt] at (axis cs:1,0.4452805) {$L=256$};
\addplot[color=runpileL512,densely dashed,line width=1.1pt,opacity=0.85,mark=*,mark size=1.25pt,mark options={line width=0pt},error bars/error bar style={draw=runpileL512,opacity=0.22,draw opacity=0.22,line width=0.55pt},error bars/error mark=none,error bars/.cd,y dir=both,y explicit] coordinates {(0.4,0.4221754) +- (0,0.0091198) (0.6,0.4119439) +- (0,0.0076701) (0.8,0.4002814) +- (0,0.0066351) (1,0.3865223) +- (0,0.0059575)};
\addplot[color=runpileL512,line width=1.6pt,mark=none,error bars/error bar style={draw=runpileL512,opacity=0.4,draw opacity=0.4,fill opacity=0.4,line width=0.75pt},error bars/error mark=none,error bars/.cd,y dir=both,y explicit] coordinates {(0.4,0.4533983) +- (0,0.01073989) (0.6,0.4629031) +- (0,0.009406336) (0.8,0.4397623) +- (0,0.007553816) (1,0.3864951) +- (0,0.005991666)};
\addplot[only marks,color=runpileL512,mark=*,mark size=1.6pt,mark options={line width=0pt}] coordinates {(0.4,0.4533983) (0.8,0.4397623) (1,0.3864951)};
\addplot[only marks,color=runpileL512,mark=star,mark size=3.2pt,line width=1.1pt] coordinates {(0.6,0.4629031)};
\node[font=\fontsize{5}{5.5}\selectfont,text=runpileL512!75!black,anchor=south east,xshift=-2.5pt,yshift=1.5pt,fill=white,fill opacity=0.94,text opacity=1,inner sep=0.25pt] at (axis cs:1,0.3864951) {$L=512$};
\addplot[black,densely dashed,line width=0.55pt] coordinates {(0.4,0.7161106) (1,0.7161106)};
\end{groupplot}
\end{tikzpicture}

%% file: figures/results/observation_general_gain_posttraining.tex
\definecolor{runL1}{HTML}{440154}
\definecolor{runL4}{HTML}{482878}
\definecolor{runL8}{HTML}{3E4989}
\definecolor{runL16}{HTML}{31688E}
\definecolor{runL32}{HTML}{26828E}
\definecolor{runL64}{HTML}{1F9E89}
\definecolor{runL128}{HTML}{35B779}
\definecolor{runL256}{HTML}{6CCE59}
\definecolor{runL512}{HTML}{A0DA39}
\definecolor{runL1024}{HTML}{C8C928}
\definecolor{runL2048}{HTML}{D79B00}
\definecolor{runscratchdeltaL64}{HTML}{1F9E89}
\definecolor{runscratchdeltaL256}{HTML}{6CCE59}
\definecolor{runscratchdeltaL1024}{HTML}{C8C928}
\definecolor{runscratchdeltaL2048}{HTML}{D79B00}
\definecolor{runpileL1}{HTML}{440154}
\definecolor{runpileL32}{HTML}{3E4989}
\definecolor{runpileL64}{HTML}{26828E}
\definecolor{runpileL128}{HTML}{1F9E89}
\definecolor{runpileL256}{HTML}{6CCE59}
\definecolor{runpileL512}{HTML}{C8C928}
\definecolor{runpersonaL4}{HTML}{440154}
\definecolor{runpersonaL8}{HTML}{1F9E89}
\definecolor{runpersonaL16}{HTML}{C8C928}
\definecolor{runsftL32}{HTML}{440154}
\definecolor{runsftL64}{HTML}{26828E}
\definecolor{runsftL128}{HTML}{6CCE59}
\definecolor{runsftL256}{HTML}{C8C928}
\definecolor{runrlvrL256}{HTML}{440154}
\definecolor{runrlvrL1024}{HTML}{C8C928}
\begin{tikzpicture}[trim axis group left,trim axis group right]
\begin{groupplot}[
  group style={group size=3 by 1, horizontal sep=3.4em, vertical sep=1.0em},
  width=0.260\textwidth,
  height=0.235\textwidth,
  scale only axis,
  title style={font=\fontsize{10}{11}\selectfont,align=center,yshift=-0.7ex},
  label style={font=\fontsize{8.5}{9.5}\selectfont},
  tick label style={font=\fontsize{8}{9}\selectfont,/pgf/number format/fixed,/pgf/number format/precision=2},
  tick align=outside,
  tick pos=left,
  axis line style={line width=0.4pt},
  major grid style={gray!20,line width=0.3pt},
  grid=major,
  enlarge x limits=0.06,
  enlarge y limits=0.10,
  scaled ticks=false,
  every axis plot/.append style={line join=round},
]
\nextgroupplot[title={SDPO},xlabel={$\lambda$},ylabel={Average gain (AvgGain)}]
\addplot[color=runpersonaL4,line width=1.6pt,mark=none,error bars/error bar style={draw=runpersonaL4,opacity=0.4,draw opacity=0.4,fill opacity=0.4,line width=0.75pt},error bars/error mark=none,error bars/.cd,y dir=both,y explicit] coordinates {(0,0.8977459) +- (0,1.890684) (0.01,1.515881) +- (0,1.406935) (0.05,5.408873) +- (0,1.579929) (0.1,9.581419) +- (0,1.964535) (0.2,9.615978) +- (0,1.186504) (0.4,10.60121) +- (0,2.131751) (0.6,6.134144) +- (0,2.949077) (0.8,4.803472) +- (0,2.769158) (1,2.448839) +- (0,4.904647)};
\addplot[only marks,color=runpersonaL4,mark=*,mark size=1.6pt,mark options={line width=0pt}] coordinates {(0,0.8977459) (0.01,1.515881) (0.05,5.408873) (0.1,9.581419) (0.2,9.615978) (0.6,6.134144) (0.8,4.803472) (1,2.448839)};
\addplot[only marks,color=runpersonaL4,mark=star,mark size=3.2pt,line width=1.1pt] coordinates {(0.4,10.60121)};
\node[font=\fontsize{5}{5.5}\selectfont,text=runpersonaL4!75!black,anchor=south east,xshift=-2.5pt,yshift=1.5pt,fill=white,fill opacity=0.94,text opacity=1,inner sep=0.25pt] at (axis cs:1,2.448839) {$L=4$};
\addplot[color=runpersonaL8,line width=1.6pt,mark=none,error bars/error bar style={draw=runpersonaL8,opacity=0.4,draw opacity=0.4,fill opacity=0.4,line width=0.75pt},error bars/error mark=none,error bars/.cd,y dir=both,y explicit] coordinates {(0,0.5454569) +- (0,1.723323) (0.01,1.084343) +- (0,1.456015) (0.05,5.496539) +- (0,2.103959) (0.1,9.512636) +- (0,2.725305) (0.2,10.14066) +- (0,2.498999) (0.4,9.865438) +- (0,2.224669) (0.6,7.393051) +- (0,1.589612) (0.8,4.053576) +- (0,1.582105) (1,5.446037) +- (0,4.317563)};
\addplot[only marks,color=runpersonaL8,mark=*,mark size=1.6pt,mark options={line width=0pt}] coordinates {(0,0.5454569) (0.01,1.084343) (0.05,5.496539) (0.1,9.512636) (0.4,9.865438) (0.6,7.393051) (0.8,4.053576) (1,5.446037)};
\addplot[only marks,color=runpersonaL8,mark=star,mark size=3.2pt,line width=1.1pt] coordinates {(0.2,10.14066)};
\node[font=\fontsize{5}{5.5}\selectfont,text=runpersonaL8!75!black,anchor=south east,xshift=-2.0pt,yshift=3.0pt,fill=white,fill opacity=0.94,text opacity=1,inner sep=0.25pt] at (axis cs:1,5.446037) {$L=8$};
\addplot[color=runpersonaL16,line width=1.6pt,mark=none,error bars/error bar style={draw=runpersonaL16,opacity=0.4,draw opacity=0.4,fill opacity=0.4,line width=0.75pt},error bars/error mark=none,error bars/.cd,y dir=both,y explicit] coordinates {(0,0.8085707) +- (0,1.427598) (0.01,1.636041) +- (0,2.27921) (0.05,6.663936) +- (0,2.140161) (0.1,8.394323) +- (0,2.161581) (0.2,8.494306) +- (0,2.068854) (0.4,8.316928) +- (0,1.039413) (0.6,5.967534) +- (0,2.111544) (0.8,3.569404) +- (0,0.6912513) (1,4.731443) +- (0,1.924141)};
\addplot[only marks,color=runpersonaL16,mark=*,mark size=1.6pt,mark options={line width=0pt}] coordinates {(0,0.8085707) (0.01,1.636041) (0.05,6.663936) (0.1,8.394323) (0.4,8.316928) (0.6,5.967534) (0.8,3.569404) (1,4.731443)};
\addplot[only marks,color=runpersonaL16,mark=star,mark size=3.2pt,line width=1.1pt] coordinates {(0.2,8.494306)};
\node[font=\fontsize{5}{5.5}\selectfont,text=runpersonaL16!75!black,anchor=north east,xshift=-2.0pt,yshift=-3.0pt,fill=white,fill opacity=0.94,text opacity=1,inner sep=0.25pt] at (axis cs:1,4.731443) {$L=16$};
\addplot[black,densely dashed,line width=0.55pt] coordinates {(0,7.00683) (1,7.00683)};
\nextgroupplot[title={SFT},xlabel={$\lambda$}]
\addplot[color=runsftL32,line width=1.6pt,mark=none,error bars/error bar style={draw=runsftL32,opacity=0.4,draw opacity=0.4,fill opacity=0.4,line width=0.75pt},error bars/error mark=none,error bars/.cd,y dir=both,y explicit] coordinates {(0,-1.976562) +- (0,0.08934987) (0.2,37.28437) +- (0,1.642108) (0.4,41.53281) +- (0,1.547105) (0.6,39.2) +- (0,2.076134) (0.8,34.775) +- (0,1.560726) (1,30.43125) +- (0,1.529805)};
\addplot[only marks,color=runsftL32,mark=*,mark size=1.6pt,mark options={line width=0pt}] coordinates {(0,-1.976562) (0.2,37.28437) (0.6,39.2) (0.8,34.775) (1,30.43125)};
\addplot[only marks,color=runsftL32,mark=star,mark size=3.2pt,line width=1.1pt] coordinates {(0.4,41.53281)};
\node[font=\fontsize{5}{5.5}\selectfont,text=runsftL32!75!black,anchor=south east,xshift=-2.5pt,yshift=1.5pt,fill=white,fill opacity=0.94,text opacity=1,inner sep=0.25pt] at (axis cs:1,30.43125) {$L=32$};
\addplot[color=runsftL64,line width=1.6pt,mark=none,error bars/error bar style={draw=runsftL64,opacity=0.4,draw opacity=0.4,fill opacity=0.4,line width=0.75pt},error bars/error mark=none,error bars/.cd,y dir=both,y explicit] coordinates {(0,-1.88125) +- (0,0.217328) (0.2,28.15) +- (0,2.466854) (0.4,33.10625) +- (0,2.430521) (0.6,28.77188) +- (0,3.736142) (0.8,27.30625) +- (0,3.639826) (1,20.96562) +- (0,3.682562)};
\addplot[only marks,color=runsftL64,mark=*,mark size=1.6pt,mark options={line width=0pt}] coordinates {(0,-1.88125) (0.2,28.15) (0.6,28.77188) (0.8,27.30625) (1,20.96562)};
\addplot[only marks,color=runsftL64,mark=star,mark size=3.2pt,line width=1.1pt] coordinates {(0.4,33.10625)};
\node[font=\fontsize{5}{5.5}\selectfont,text=runsftL64!75!black,anchor=south east,xshift=-2.0pt,yshift=3.0pt,fill=white,fill opacity=0.94,text opacity=1,inner sep=0.25pt] at (axis cs:1,20.96562) {$L=64$};
\addplot[color=runsftL128,line width=1.6pt,mark=none,error bars/error bar style={draw=runsftL128,opacity=0.4,draw opacity=0.4,fill opacity=0.4,line width=0.75pt},error bars/error mark=none,error bars/.cd,y dir=both,y explicit] coordinates {(0,-2.06875) +- (0,0.181788) (0.2,27.4375) +- (0,1.265799) (0.4,27.29375) +- (0,1.313489) (0.6,28.6125) +- (0,1.265937) (0.8,24.79375) +- (0,1.481678) (1,21.825) +- (0,1.582053)};
\addplot[only marks,color=runsftL128,mark=*,mark size=1.6pt,mark options={line width=0pt}] coordinates {(0,-2.06875) (0.2,27.4375) (0.4,27.29375) (0.8,24.79375) (1,21.825)};
\addplot[only marks,color=runsftL128,mark=star,mark size=3.2pt,line width=1.1pt] coordinates {(0.6,28.6125)};
\node[font=\fontsize{5}{5.5}\selectfont,text=runsftL128!75!black,anchor=north east,xshift=-2.0pt,yshift=-3.0pt,fill=white,fill opacity=0.94,text opacity=1,inner sep=0.25pt] at (axis cs:1,21.825) {$L=128$};
\addplot[color=runsftL256,line width=1.6pt,mark=none,error bars/error bar style={draw=runsftL256,opacity=0.4,draw opacity=0.4,fill opacity=0.4,line width=0.75pt},error bars/error mark=none,error bars/.cd,y dir=both,y explicit] coordinates {(0,-2.1875) +- (0,0.4406139) (0.2,18.25) +- (0,2.69454) (0.4,18.025) +- (0,1.42585) (0.6,14.3875) +- (0,4.834746) (0.8,9.0875) +- (0,2.786589) (1,6.9125) +- (0,2.44874)};
\addplot[only marks,color=runsftL256,mark=*,mark size=1.6pt,mark options={line width=0pt}] coordinates {(0,-2.1875) (0.4,18.025) (0.6,14.3875) (0.8,9.0875) (1,6.9125)};
\addplot[only marks,color=runsftL256,mark=star,mark size=3.2pt,line width=1.1pt] coordinates {(0.2,18.25)};
\node[font=\fontsize{5}{5.5}\selectfont,text=runsftL256!75!black,anchor=south east,xshift=-2.5pt,yshift=1.5pt,fill=white,fill opacity=0.94,text opacity=1,inner sep=0.25pt] at (axis cs:1,6.9125) {$L=256$};
\addplot[black,densely dashed,line width=0.55pt] coordinates {(0,38.23594) (1,38.23594)};
\nextgroupplot[title={GRPO},xlabel={$\lambda$}]
\addplot[color=runrlvrL256,line width=1.6pt,mark=none,error bars/error bar style={draw=runrlvrL256,opacity=0.4,draw opacity=0.4,fill opacity=0.4,line width=0.75pt},error bars/error mark=none,error bars/.cd,y dir=both,y explicit] coordinates {(0,0.4427083) +- (0,0.198942) (0.2,14.92708) +- (0,1.079759) (0.4,21.06771) +- (0,0.6005866) (0.6,23.08333) +- (0,0.7560699) (0.8,21.81771) +- (0,1.016639) (1,21.10417) +- (0,1.979167)};
\addplot[only marks,color=runrlvrL256,mark=*,mark size=1.6pt,mark options={line width=0pt}] coordinates {(0,0.4427083) (0.2,14.92708) (0.4,21.06771) (0.8,21.81771) (1,21.10417)};
\addplot[only marks,color=runrlvrL256,mark=star,mark size=3.2pt,line width=1.1pt] coordinates {(0.6,23.08333)};
\node[font=\fontsize{5}{5.5}\selectfont,text=runrlvrL256!75!black,anchor=south east,xshift=-2.5pt,yshift=1.5pt,fill=white,fill opacity=0.94,text opacity=1,inner sep=0.25pt] at (axis cs:1,21.10417) {$L=256$};
\addplot[color=runrlvrL1024,line width=1.6pt,mark=none,error bars/error bar style={draw=runrlvrL1024,opacity=0.4,draw opacity=0.4,fill opacity=0.4,line width=0.75pt},error bars/error mark=none,error bars/.cd,y dir=both,y explicit] coordinates {(0,0.5) +- (0,0.09547033) (0.2,8.6875) +- (0,1.431145) (0.4,15.91667) +- (0,3.364632) (0.6,16.625) +- (0,5.562968) (0.8,16.52083) +- (0,5.247188) (1,15) +- (0,7.169089)};
\addplot[only marks,color=runrlvrL1024,mark=*,mark size=1.6pt,mark options={line width=0pt}] coordinates {(0,0.5) (0.2,8.6875) (0.4,15.91667) (0.8,16.52083) (1,15)};
\addplot[only marks,color=runrlvrL1024,mark=star,mark size=3.2pt,line width=1.1pt] coordinates {(0.6,16.625)};
\node[font=\fontsize{5}{5.5}\selectfont,text=runrlvrL1024!75!black,anchor=south east,xshift=-2.5pt,yshift=1.5pt,fill=white,fill opacity=0.94,text opacity=1,inner sep=0.25pt] at (axis cs:1,15) {$L=1024$};
\addplot[black,densely dashed,line width=0.55pt] coordinates {(0,26.69271) (1,26.69271)};
\end{groupplot}
\end{tikzpicture}

%% file: figures/results/observation_nontargeted_targeted_continual.tex
\definecolor{runL1}{HTML}{440154}
\definecolor{runL4}{HTML}{482878}
\definecolor{runL8}{HTML}{3E4989}
\definecolor{runL16}{HTML}{31688E}
\definecolor{runL32}{HTML}{26828E}
\definecolor{runL64}{HTML}{1F9E89}
\definecolor{runL128}{HTML}{35B779}
\definecolor{runL256}{HTML}{6CCE59}
\definecolor{runL512}{HTML}{A0DA39}
\definecolor{runL1024}{HTML}{C8C928}
\definecolor{runL2048}{HTML}{D79B00}
\definecolor{runscratchdeltaL64}{HTML}{1F9E89}
\definecolor{runscratchdeltaL256}{HTML}{6CCE59}
\definecolor{runscratchdeltaL1024}{HTML}{C8C928}
\definecolor{runscratchdeltaL2048}{HTML}{D79B00}
\definecolor{runpileL1}{HTML}{440154}
\definecolor{runpileL32}{HTML}{3E4989}
\definecolor{runpileL64}{HTML}{26828E}
\definecolor{runpileL128}{HTML}{1F9E89}
\definecolor{runpileL256}{HTML}{6CCE59}
\definecolor{runpileL512}{HTML}{C8C928}
\definecolor{runpersonaL4}{HTML}{440154}
\definecolor{runpersonaL8}{HTML}{1F9E89}
\definecolor{runpersonaL16}{HTML}{C8C928}
\definecolor{runsftL32}{HTML}{440154}
\definecolor{runsftL64}{HTML}{26828E}
\definecolor{runsftL128}{HTML}{6CCE59}
\definecolor{runsftL256}{HTML}{C8C928}
\definecolor{runrlvrL256}{HTML}{440154}
\definecolor{runrlvrL1024}{HTML}{C8C928}
\begin{tikzpicture}[trim axis group left,trim axis group right]
\begin{groupplot}[
  group style={group size=1 by 1, horizontal sep=0pt, vertical sep=1.0em},
  width=0.76\linewidth,
  height=0.55\linewidth,
  scale only axis,
  title style={font=\fontsize{10}{11}\selectfont,align=center,yshift=-0.7ex},
  label style={font=\fontsize{8.5}{9.5}\selectfont},
  tick label style={font=\fontsize{8}{9}\selectfont,/pgf/number format/fixed,/pgf/number format/precision=2},
  tick align=outside,
  tick pos=left,
  axis line style={line width=0.4pt},
  major grid style={gray!20,line width=0.3pt},
  grid=major,
  enlarge x limits=0.06,
  enlarge y limits=0.10,
  scaled ticks=false,
  every axis plot/.append style={line join=round},
]
\nextgroupplot[xlabel={Stability\\(non-targeted gain)},ylabel={Plasticity (targeted gain)},xlabel style={align=center},legend style={at={(0.015,0.005)},anchor=south west,draw=none,fill=white,fill opacity=0.92,text opacity=1,font=\fontsize{5.25}{6}\selectfont,inner sep=0.6pt,column sep=0.25em,row sep=-0.4pt},legend columns=2,legend cell align={left},legend image post style={yshift=0.25ex}]
\addlegendimage{empty legend}
\addlegendentry{lowest $\lambda=0.4$}
\addlegendimage{only marks,black,mark=triangle*,mark size=2.5pt,mark options={line width=0pt,yshift=-0.15ex}}
\addlegendentry{$\lambda=1.0$}
\addplot[color=runpileL32,densely dashed,line width=1.1pt,opacity=0.85,mark=*,mark size=1.25pt,mark options={line width=0pt},error bars/error bar style={draw=runpileL32,opacity=0.22,draw opacity=0.22,line width=0.55pt},error bars/error mark=none,error bars/.cd,x dir=both,x explicit,y dir=both,y explicit] coordinates {(0.5747062,0.6365263) +- (0.0044464,0.0011005) (0.5771012,0.6547849) +- (0.0044391,0.0008292) (0.5728599,0.6641015) +- (0.0043903,0.0006705) (0.5656631,0.6685208) +- (0.0045563,0.0005108)};
\addplot[color=runpileL32,line width=1.6pt,mark=none,error bars/error bar style={draw=runpileL32,opacity=0.4,draw opacity=0.4,fill opacity=0.4,line width=0.75pt},error bars/error mark=none,error bars/.cd,x dir=both,x explicit,y dir=both,y explicit] coordinates {(0.5772887,0.6481032) +- (0.00488872,0.00038333) (0.5853802,0.6603296) +- (0.004674373,0.0003981127) (0.5804024,0.6664659) +- (0.004509119,0.0005133169) (0.565664,0.6685209) +- (0.004555219,0.0005107349)};
\addplot[only marks,color=runpileL32,mark=*,mark size=1.6pt,mark options={line width=0pt}] coordinates {(0.5772887,0.6481032) (0.5853802,0.6603296) (0.5804024,0.6664659)};
\addplot[only marks,color=runpileL32,mark=triangle*,mark size=2.8pt,mark options={line width=0pt}] coordinates {(0.565664,0.6685209)};
\node[font=\fontsize{5}{5.5}\selectfont,text=runpileL32!75!black,anchor=south east,xshift=-2.5pt,yshift=1.5pt,fill=white,fill opacity=0.94,text opacity=1,inner sep=0.25pt] at (axis cs:0.565664,0.6685209) {$L=32$};
\addplot[color=runpileL64,densely dashed,line width=1.1pt,opacity=0.85,mark=*,mark size=1.25pt,mark options={line width=0pt},error bars/error bar style={draw=runpileL64,opacity=0.22,draw opacity=0.22,line width=0.55pt},error bars/error mark=none,error bars/.cd,x dir=both,x explicit,y dir=both,y explicit] coordinates {(0.5396795,0.6511562) +- (0.0048343,0.0000745) (0.5397145,0.6696495) +- (0.0047551,0.0001812) (0.5348792,0.6791472) +- (0.0050758,0.0002853) (0.5281828,0.6838052) +- (0.0048626,0.0002523)};
\addplot[color=runpileL64,line width=1.6pt,mark=none,error bars/error bar style={draw=runpileL64,opacity=0.4,draw opacity=0.4,fill opacity=0.4,line width=0.75pt},error bars/error mark=none,error bars/.cd,x dir=both,x explicit,y dir=both,y explicit] coordinates {(0.5525506,0.6676541) +- (0.004269983,0.0001352052) (0.5591328,0.6773554) +- (0.00456719,0.0001301614) (0.55072,0.6822802) +- (0.004726468,0.0001715741) (0.5281826,0.6838053) +- (0.004862619,0.0002523836)};
\addplot[only marks,color=runpileL64,mark=*,mark size=1.6pt,mark options={line width=0pt}] coordinates {(0.5525506,0.6676541) (0.5591328,0.6773554) (0.55072,0.6822802)};
\addplot[only marks,color=runpileL64,mark=triangle*,mark size=2.8pt,mark options={line width=0pt}] coordinates {(0.5281826,0.6838053)};
\node[font=\fontsize{5}{5.5}\selectfont,text=runpileL64!75!black,anchor=south west,xshift=2.5pt,yshift=1.5pt,fill=white,fill opacity=0.94,text opacity=1,inner sep=0.25pt] at (axis cs:0.5281826,0.6838053) {$L=64$};
\addplot[color=runpileL128,densely dashed,line width=1.1pt,opacity=0.85,mark=*,mark size=1.25pt,mark options={line width=0pt},error bars/error bar style={draw=runpileL128,opacity=0.22,draw opacity=0.22,line width=0.55pt},error bars/error mark=none,error bars/.cd,x dir=both,x explicit,y dir=both,y explicit] coordinates {(0.4991116,0.6673830) +- (0.0102638,0.0007991) (0.4942868,0.6847061) +- (0.0112193,0.0005196) (0.4864365,0.6930662) +- (0.0118821,0.0005735) (0.4776729,0.6965902) +- (0.0116486,0.0007977)};
\addplot[color=runpileL128,line width=1.6pt,mark=none,error bars/error bar style={draw=runpileL128,opacity=0.4,draw opacity=0.4,fill opacity=0.4,line width=0.75pt},error bars/error mark=none,error bars/.cd,x dir=both,x explicit,y dir=both,y explicit] coordinates {(0.5190454,0.6895134) +- (0.00969782,0.0004875806) (0.5237188,0.6950772) +- (0.01074432,0.0006567884) (0.5093826,0.697122) +- (0.01114411,0.0007193954) (0.4776755,0.6965897) +- (0.01164803,0.0007976028)};
\addplot[only marks,color=runpileL128,mark=*,mark size=1.6pt,mark options={line width=0pt}] coordinates {(0.5190454,0.6895134) (0.5237188,0.6950772) (0.5093826,0.697122)};
\addplot[only marks,color=runpileL128,mark=triangle*,mark size=2.8pt,mark options={line width=0pt}] coordinates {(0.4776755,0.6965897)};
\node[font=\fontsize{5}{5.5}\selectfont,text=runpileL128!75!black,anchor=south west,xshift=2.5pt,yshift=1.5pt,fill=white,fill opacity=0.94,text opacity=1,inner sep=0.25pt] at (axis cs:0.4776755,0.6965897) {$L=128$};
\addplot[color=runpileL256,densely dashed,line width=1.1pt,opacity=0.85,mark=*,mark size=1.25pt,mark options={line width=0pt},error bars/error bar style={draw=runpileL256,opacity=0.22,draw opacity=0.22,line width=0.55pt},error bars/error mark=none,error bars/.cd,x dir=both,x explicit,y dir=both,y explicit] coordinates {(0.4448729,0.6842266) +- (0.0194312,0.0015079) (0.4362134,0.7007349) +- (0.0174632,0.0017143) (0.4265066,0.7085664) +- (0.0155557,0.0020596) (0.4156263,0.7121823) +- (0.0147693,0.0021899)};
\addplot[color=runpileL256,line width=1.6pt,mark=none,error bars/error bar style={draw=runpileL256,opacity=0.4,draw opacity=0.4,fill opacity=0.4,line width=0.75pt},error bars/error mark=none,error bars/.cd,x dir=both,x explicit,y dir=both,y explicit] coordinates {(0.4757558,0.7145526) +- (0.01705937,0.001216434) (0.4790211,0.7160261) +- (0.01678261,0.001562775) (0.458406,0.7148839) +- (0.01518523,0.001899779) (0.4156244,0.7121849) +- (0.01477027,0.002191064)};
\addplot[only marks,color=runpileL256,mark=*,mark size=1.6pt,mark options={line width=0pt}] coordinates {(0.4757558,0.7145526) (0.4790211,0.7160261) (0.458406,0.7148839)};
\addplot[only marks,color=runpileL256,mark=triangle*,mark size=2.8pt,mark options={line width=0pt}] coordinates {(0.4156244,0.7121849)};
\node[font=\fontsize{5}{5.5}\selectfont,text=runpileL256!75!black,anchor=south west,xshift=2.5pt,yshift=1.5pt,fill=white,fill opacity=0.94,text opacity=1,inner sep=0.25pt] at (axis cs:0.4156244,0.7121849) {$L=256$};
\addplot[color=runpileL512,densely dashed,line width=1.1pt,opacity=0.85,mark=*,mark size=1.25pt,mark options={line width=0pt},error bars/error bar style={draw=runpileL512,opacity=0.22,draw opacity=0.22,line width=0.55pt},error bars/error mark=none,error bars/.cd,x dir=both,x explicit,y dir=both,y explicit] coordinates {(0.3902423,0.7095733) +- (0.0101961,0.0005948) (0.3770980,0.7255566) +- (0.0085845,0.0008090) (0.3632982,0.7331300) +- (0.0074632,0.0010751) (0.3476279,0.7365713) +- (0.0067074,0.0011357)};
\addplot[color=runpileL512,line width=1.6pt,mark=none,error bars/error bar style={draw=runpileL512,opacity=0.4,draw opacity=0.4,fill opacity=0.4,line width=0.75pt},error bars/error mark=none,error bars/.cd,x dir=both,x explicit,y dir=both,y explicit] coordinates {(0.434806,0.7447593) +- (0.01250172,0.0009081441) (0.4370943,0.7441817) +- (0.01068572,0.001106633) (0.4074175,0.7411628) +- (0.008520088,0.001072855) (0.3475973,0.7365759) +- (0.00674542,0.001136092)};
\addplot[only marks,color=runpileL512,mark=*,mark size=1.6pt,mark options={line width=0pt}] coordinates {(0.434806,0.7447593) (0.4370943,0.7441817) (0.4074175,0.7411628)};
\addplot[only marks,color=runpileL512,mark=triangle*,mark size=2.8pt,mark options={line width=0pt}] coordinates {(0.3475973,0.7365759)};
\node[font=\fontsize{5}{5.5}\selectfont,text=runpileL512!75!black,anchor=south west,xshift=2.5pt,yshift=1.5pt,fill=white,fill opacity=0.94,text opacity=1,inner sep=0.25pt] at (axis cs:0.3475973,0.7365759) {$L=512$};
\end{groupplot}
\end{tikzpicture}

%% file: figures/results/observation_nontargeted_targeted_posttraining.tex
\definecolor{runL1}{HTML}{440154}
\definecolor{runL4}{HTML}{482878}
\definecolor{runL8}{HTML}{3E4989}
\definecolor{runL16}{HTML}{31688E}
\definecolor{runL32}{HTML}{26828E}
\definecolor{runL64}{HTML}{1F9E89}
\definecolor{runL128}{HTML}{35B779}
\definecolor{runL256}{HTML}{6CCE59}
\definecolor{runL512}{HTML}{A0DA39}
\definecolor{runL1024}{HTML}{C8C928}
\definecolor{runL2048}{HTML}{D79B00}
\definecolor{runscratchdeltaL64}{HTML}{1F9E89}
\definecolor{runscratchdeltaL256}{HTML}{6CCE59}
\definecolor{runscratchdeltaL1024}{HTML}{C8C928}
\definecolor{runscratchdeltaL2048}{HTML}{D79B00}
\definecolor{runpileL1}{HTML}{440154}
\definecolor{runpileL32}{HTML}{3E4989}
\definecolor{runpileL64}{HTML}{26828E}
\definecolor{runpileL128}{HTML}{1F9E89}
\definecolor{runpileL256}{HTML}{6CCE59}
\definecolor{runpileL512}{HTML}{C8C928}
\definecolor{runpersonaL4}{HTML}{440154}
\definecolor{runpersonaL8}{HTML}{1F9E89}
\definecolor{runpersonaL16}{HTML}{C8C928}
\definecolor{runsftL32}{HTML}{440154}
\definecolor{runsftL64}{HTML}{26828E}
\definecolor{runsftL128}{HTML}{6CCE59}
\definecolor{runsftL256}{HTML}{C8C928}
\definecolor{runrlvrL256}{HTML}{440154}
\definecolor{runrlvrL1024}{HTML}{C8C928}
\begin{tikzpicture}[trim axis group left,trim axis group right]
\begin{groupplot}[
  group style={group size=3 by 1, horizontal sep=3.4em, vertical sep=1.0em},
  width=0.260\textwidth,
  height=0.235\textwidth,
  scale only axis,
  title style={font=\fontsize{10}{11}\selectfont,align=center,yshift=-0.7ex},
  label style={font=\fontsize{8.5}{9.5}\selectfont},
  tick label style={font=\fontsize{8}{9}\selectfont,/pgf/number format/fixed,/pgf/number format/precision=2},
  tick align=outside,
  tick pos=left,
  axis line style={line width=0.4pt},
  major grid style={gray!20,line width=0.3pt},
  grid=major,
  enlarge x limits=0.06,
  enlarge y limits=0.10,
  scaled ticks=false,
  every axis plot/.append style={line join=round},
]
\nextgroupplot[title={SDPO},xlabel={Stability\\(non-targeted gain)},ylabel={Plasticity (targeted gain)},xlabel style={align=center},legend style={at={(0.015,0.005)},anchor=south west,draw=none,fill=white,fill opacity=0.92,text opacity=1,font=\fontsize{5.25}{6}\selectfont,inner sep=0.6pt,column sep=0.25em,row sep=-0.4pt},legend columns=2,legend cell align={left},legend image post style={yshift=0.25ex}]
\addlegendimage{only marks,black,mark=square*,mark size=2.0pt,mark options={line width=0pt}}
\addlegendentry{$\lambda=0.0$}
\addlegendimage{only marks,black,mark=triangle*,mark size=2.5pt,mark options={line width=0pt,yshift=-0.15ex}}
\addlegendentry{$\lambda=1.0$}
\addplot[color=runpersonaL4,line width=1.6pt,mark=none,error bars/error bar style={draw=runpersonaL4,opacity=0.4,draw opacity=0.4,fill opacity=0.4,line width=0.75pt},error bars/error mark=none,error bars/.cd,x dir=both,x explicit,y dir=both,y explicit] coordinates {(0.6490407,11.78275) +- (1.959782,2.110044) (1.770712,12.86986) +- (1.370552,2.5936) (5.307295,10.99778) +- (1.552188,2.544722) (9.654839,11.40741) +- (1.938169,2.889741) (9.764523,10.19287) +- (1.032964,1.902629) (10.71048,10.18191) +- (2.109772,1.961126) (6.266789,5.043195) +- (2.904639,3.801998) (4.778934,4.683067) +- (2.647713,3.555017) (2.458953,1.964363) +- (4.844703,5.036925)};
\addplot[only marks,color=runpersonaL4,mark=*,mark size=1.6pt,mark options={line width=0pt}] coordinates {(1.770712,12.86986) (5.307295,10.99778) (9.654839,11.40741) (9.764523,10.19287) (10.71048,10.18191) (6.266789,5.043195) (4.778934,4.683067)};
\addplot[only marks,color=runpersonaL4,mark=square*,mark size=2.0pt,mark options={line width=0pt}] coordinates {(0.6490407,11.78275)};
\addplot[only marks,color=runpersonaL4,mark=triangle*,mark size=2.8pt,mark options={line width=0pt}] coordinates {(2.458953,1.964363)};
\node[font=\fontsize{5}{5.5}\selectfont,text=runpersonaL4!75!black,anchor=south west,xshift=2.5pt,yshift=1.5pt,fill=white,fill opacity=0.94,text opacity=1,inner sep=0.25pt] at (axis cs:2.458953,1.964363) {$L=4$};
\addplot[color=runpersonaL8,line width=1.6pt,mark=none,error bars/error bar style={draw=runpersonaL8,opacity=0.4,draw opacity=0.4,fill opacity=0.4,line width=0.75pt},error bars/error mark=none,error bars/.cd,x dir=both,x explicit,y dir=both,y explicit] coordinates {(0.6153617,14.10615) +- (1.811392,1.518129) (1.410339,13.07855) +- (1.465369,3.294457) (5.489408,12.3437) +- (1.802707,1.675838) (9.39901,10.97867) +- (2.794764,2.479135) (10.20239,12.38383) +- (2.559744,2.758155) (9.681941,9.797109) +- (2.468615,1.531047) (7.382655,8.629073) +- (1.679039,1.782731) (3.898136,3.63639) +- (1.75129,1.077282) (5.23846,6.068621) +- (4.478252,4.087722)};
\addplot[only marks,color=runpersonaL8,mark=*,mark size=1.6pt,mark options={line width=0pt}] coordinates {(1.410339,13.07855) (5.489408,12.3437) (9.39901,10.97867) (10.20239,12.38383) (9.681941,9.797109) (7.382655,8.629073) (3.898136,3.63639)};
\addplot[only marks,color=runpersonaL8,mark=square*,mark size=2.0pt,mark options={line width=0pt}] coordinates {(0.6153617,14.10615)};
\addplot[only marks,color=runpersonaL8,mark=triangle*,mark size=2.8pt,mark options={line width=0pt}] coordinates {(5.23846,6.068621)};
\node[font=\fontsize{5}{5.5}\selectfont,text=runpersonaL8!75!black,anchor=south west,xshift=2.0pt,yshift=3.0pt,fill=white,fill opacity=0.94,text opacity=1,inner sep=0.25pt] at (axis cs:5.23846,6.068621) {$L=8$};
\addplot[color=runpersonaL16,line width=1.6pt,mark=none,error bars/error bar style={draw=runpersonaL16,opacity=0.4,draw opacity=0.4,fill opacity=0.4,line width=0.75pt},error bars/error mark=none,error bars/.cd,x dir=both,x explicit,y dir=both,y explicit] coordinates {(1.080174,15.19662) +- (1.494667,4.545094) (1.787632,14.93841) +- (2.342985,2.163617) (6.66442,16.20308) +- (2.146964,3.727158) (8.004651,14.56942) +- (2.407985,3.617391) (8.007241,12.35598) +- (2.318188,3.013794) (7.791023,13.20611) +- (1.100852,1.394961) (5.367352,10.71911) +- (2.288671,1.364797) (3.18877,7.692829) +- (0.7684671,1.520391) (4.408878,5.905305) +- (2.085574,2.241832)};
\addplot[only marks,color=runpersonaL16,mark=*,mark size=1.6pt,mark options={line width=0pt}] coordinates {(1.787632,14.93841) (6.66442,16.20308) (8.004651,14.56942) (8.007241,12.35598) (7.791023,13.20611) (5.367352,10.71911) (3.18877,7.692829)};
\addplot[only marks,color=runpersonaL16,mark=square*,mark size=2.0pt,mark options={line width=0pt}] coordinates {(1.080174,15.19662)};
\addplot[only marks,color=runpersonaL16,mark=triangle*,mark size=2.8pt,mark options={line width=0pt}] coordinates {(4.408878,5.905305)};
\node[font=\fontsize{5}{5.5}\selectfont,text=runpersonaL16!75!black,anchor=north east,xshift=-2.0pt,yshift=-3.0pt,fill=white,fill opacity=0.94,text opacity=1,inner sep=0.25pt] at (axis cs:4.408878,5.905305) {$L=16$};
\nextgroupplot[title={SFT},xlabel={Stability\\(non-targeted gain)},xlabel style={align=center},legend style={at={(0.015,0.005)},anchor=south west,draw=none,fill=white,fill opacity=0.92,text opacity=1,font=\fontsize{5.25}{6}\selectfont,inner sep=0.6pt,column sep=0.25em,row sep=-0.4pt},legend columns=2,legend cell align={left},legend image post style={yshift=0.25ex}]
\addlegendimage{only marks,black,mark=square*,mark size=2.0pt,mark options={line width=0pt}}
\addlegendentry{$\lambda=0.0$}
\addlegendimage{only marks,black,mark=triangle*,mark size=2.5pt,mark options={line width=0pt,yshift=-0.15ex}}
\addlegendentry{$\lambda=1.0$}
\addplot[color=runsftL32,line width=1.6pt,mark=none,error bars/error bar style={draw=runsftL32,opacity=0.4,draw opacity=0.4,fill opacity=0.4,line width=0.75pt},error bars/error mark=none,error bars/.cd,x dir=both,x explicit,y dir=both,y explicit] coordinates {(-2.095833,29.35) +- (0.08779711,0.956444) (36.06875,32.49375) +- (1.869571,1.34311) (39.19792,39.30625) +- (1.985431,0.7346343) (35.99792,42.99375) +- (2.63167,0.8783085) (30.5,43.46875) +- (1.87568,1.05248) (25.25208,45.80625) +- (1.902137,1.105358)};
\addplot[only marks,color=runsftL32,mark=*,mark size=1.6pt,mark options={line width=0pt}] coordinates {(36.06875,32.49375) (39.19792,39.30625) (35.99792,42.99375) (30.5,43.46875)};
\addplot[only marks,color=runsftL32,mark=square*,mark size=2.0pt,mark options={line width=0pt}] coordinates {(-2.095833,29.35)};
\addplot[only marks,color=runsftL32,mark=triangle*,mark size=2.8pt,mark options={line width=0pt}] coordinates {(25.25208,45.80625)};
\node[font=\fontsize{5}{5.5}\selectfont,text=runsftL32!75!black,anchor=south west,xshift=2.5pt,yshift=1.5pt,fill=white,fill opacity=0.94,text opacity=1,inner sep=0.25pt] at (axis cs:25.25208,45.80625) {$L=32$};
\addplot[color=runsftL64,line width=1.6pt,mark=none,error bars/error bar style={draw=runsftL64,opacity=0.4,draw opacity=0.4,fill opacity=0.4,line width=0.75pt},error bars/error mark=none,error bars/.cd,x dir=both,x explicit,y dir=both,y explicit] coordinates {(-1.8875,43.0875) +- (0.2749369,1.005376) (25.07917,38) +- (2.88437,1.393203) (28.09167,42.7625) +- (2.884921,0.6787512) (22.575,44.7) +- (4.266312,2.222435) (20.5375,45.2) +- (4.370948,1.508362) (12.5375,46.35) +- (4.664468,1.554932)};
\addplot[only marks,color=runsftL64,mark=*,mark size=1.6pt,mark options={line width=0pt}] coordinates {(25.07917,38) (28.09167,42.7625) (22.575,44.7) (20.5375,45.2)};
\addplot[only marks,color=runsftL64,mark=square*,mark size=2.0pt,mark options={line width=0pt}] coordinates {(-1.8875,43.0875)};
\addplot[only marks,color=runsftL64,mark=triangle*,mark size=2.8pt,mark options={line width=0pt}] coordinates {(12.5375,46.35)};
\node[font=\fontsize{5}{5.5}\selectfont,text=runsftL64!75!black,anchor=south west,xshift=2.5pt,yshift=1.5pt,fill=white,fill opacity=0.94,text opacity=1,inner sep=0.25pt] at (axis cs:12.5375,46.35) {$L=64$};
\addplot[color=runsftL128,line width=1.6pt,mark=none,error bars/error bar style={draw=runsftL128,opacity=0.4,draw opacity=0.4,fill opacity=0.4,line width=0.75pt},error bars/error mark=none,error bars/.cd,x dir=both,x explicit,y dir=both,y explicit] coordinates {(-1.875,47.025) +- (0.2664713,3.055323) (24.75833,49.175) +- (1.269186,1.537856) (22.45,51.4) +- (1.179792,0.7617824) (21.1,51.8) +- (1.625139,1.637261) (15.125,53.75) +- (1.699826,1.40145) (11.04167,53.425) +- (2.358827,0.9317323)};
\addplot[only marks,color=runsftL128,mark=*,mark size=1.6pt,mark options={line width=0pt}] coordinates {(24.75833,49.175) (22.45,51.4) (21.1,51.8) (15.125,53.75)};
\addplot[only marks,color=runsftL128,mark=square*,mark size=2.0pt,mark options={line width=0pt}] coordinates {(-1.875,47.025)};
\addplot[only marks,color=runsftL128,mark=triangle*,mark size=2.8pt,mark options={line width=0pt}] coordinates {(11.04167,53.425)};
\node[font=\fontsize{5}{5.5}\selectfont,text=runsftL128!75!black,anchor=south west,xshift=2.5pt,yshift=1.5pt,fill=white,fill opacity=0.94,text opacity=1,inner sep=0.25pt] at (axis cs:11.04167,53.425) {$L=128$};
\addplot[color=runsftL256,line width=1.6pt,mark=none,error bars/error bar style={draw=runsftL256,opacity=0.4,draw opacity=0.4,fill opacity=0.4,line width=0.75pt},error bars/error mark=none,error bars/.cd,x dir=both,x explicit,y dir=both,y explicit] coordinates {(-2.383333,53.2) +- (0.338091,3.131892) (15.1,57.4) +- (2.078728,1.623268) (10.55,55.2) +- (1.813836,1.481975) (3.333333,55.15) +- (4.284338,3.691714) (-4.35,46.45) +- (1.550403,7.196961) (-8.466667,52.9) +- (3.089813,1.629033)};
\addplot[only marks,color=runsftL256,mark=*,mark size=1.6pt,mark options={line width=0pt}] coordinates {(15.1,57.4) (10.55,55.2) (3.333333,55.15) (-4.35,46.45)};
\addplot[only marks,color=runsftL256,mark=square*,mark size=2.0pt,mark options={line width=0pt}] coordinates {(-2.383333,53.2)};
\addplot[only marks,color=runsftL256,mark=triangle*,mark size=2.8pt,mark options={line width=0pt}] coordinates {(-8.466667,52.9)};
\node[font=\fontsize{5}{5.5}\selectfont,text=runsftL256!75!black,anchor=south west,xshift=2.5pt,yshift=1.5pt,fill=white,fill opacity=0.94,text opacity=1,inner sep=0.25pt] at (axis cs:-8.466667,52.9) {$L=256$};
\nextgroupplot[title={GRPO},xlabel={Stability\\(non-targeted gain)},xlabel style={align=center},legend style={at={(0.015,0.005)},anchor=south west,draw=none,fill=white,fill opacity=0.92,text opacity=1,font=\fontsize{5.25}{6}\selectfont,inner sep=0.6pt,column sep=0.25em,row sep=-0.4pt},legend columns=2,legend cell align={left},legend image post style={yshift=0.25ex}]
\addlegendimage{only marks,black,mark=square*,mark size=2.0pt,mark options={line width=0pt}}
\addlegendentry{$\lambda=0.0$}
\addlegendimage{only marks,black,mark=triangle*,mark size=2.5pt,mark options={line width=0pt,yshift=-0.15ex}}
\addlegendentry{$\lambda=1.0$}
\addplot[color=runrlvrL256,line width=1.6pt,mark=none,error bars/error bar style={draw=runrlvrL256,opacity=0.4,draw opacity=0.4,fill opacity=0.4,line width=0.75pt},error bars/error mark=none,error bars/.cd,x dir=both,x explicit,y dir=both,y explicit] coordinates {(0.5416667,17.91667) +- (0.1875,0.6676425) (14.375,25.27083) +- (1.229755,0.2659822) (20.35417,26.0625) +- (0.5547303,0.8977692) (22.25694,26.1875) +- (0.8798093,0.75) (21.01389,25.1875) +- (1.203134,0.8408515) (20.28472,24.16667) +- (2.048117,1.609434)};
\addplot[only marks,color=runrlvrL256,mark=*,mark size=1.6pt,mark options={line width=0pt}] coordinates {(14.375,25.27083) (20.35417,26.0625) (22.25694,26.1875) (21.01389,25.1875)};
\addplot[only marks,color=runrlvrL256,mark=square*,mark size=2.0pt,mark options={line width=0pt}] coordinates {(0.5416667,17.91667)};
\addplot[only marks,color=runrlvrL256,mark=triangle*,mark size=2.8pt,mark options={line width=0pt}] coordinates {(20.28472,24.16667)};
\node[font=\fontsize{5}{5.5}\selectfont,text=runrlvrL256!75!black,anchor=south east,xshift=-2.0pt,yshift=3.0pt,fill=white,fill opacity=0.94,text opacity=1,inner sep=0.25pt] at (axis cs:20.28472,24.16667) {$L=256$};
\addplot[color=runrlvrL1024,line width=1.6pt,mark=none,error bars/error bar style={draw=runrlvrL1024,opacity=0.4,draw opacity=0.4,fill opacity=0.4,line width=0.75pt},error bars/error mark=none,error bars/.cd,x dir=both,x explicit,y dir=both,y explicit] coordinates {(0.3888889,38.41667) +- (0.07349309,1.261062) (7.5,37.83333) +- (1.05519,0.9824855) (14.08333,30.58333) +- (3.379198,4.083333) (14.69444,27.41667) +- (5.392324,7.650345) (14.33333,24.08333) +- (4.989804,5.993631) (13,21.58333) +- (7.428517,7.478209)};
\addplot[only marks,color=runrlvrL1024,mark=*,mark size=1.6pt,mark options={line width=0pt}] coordinates {(7.5,37.83333) (14.08333,30.58333) (14.69444,27.41667) (14.33333,24.08333)};
\addplot[only marks,color=runrlvrL1024,mark=square*,mark size=2.0pt,mark options={line width=0pt}] coordinates {(0.3888889,38.41667)};
\addplot[only marks,color=runrlvrL1024,mark=triangle*,mark size=2.8pt,mark options={line width=0pt}] coordinates {(13,21.58333)};
\node[font=\fontsize{5}{5.5}\selectfont,text=runrlvrL1024!75!black,anchor=north east,xshift=-2.0pt,yshift=-3.0pt,fill=white,fill opacity=0.94,text opacity=1,inner sep=0.25pt] at (axis cs:13,21.58333) {$L=1024$};
\end{groupplot}
\end{tikzpicture}

%% file: figures/results/pretraining_scratch_outcomes.tex
\definecolor{runL1}{HTML}{440154}
\definecolor{runL4}{HTML}{482878}
\definecolor{runL8}{HTML}{3E4989}
\definecolor{runL16}{HTML}{31688E}
\definecolor{runL32}{HTML}{26828E}
\definecolor{runL64}{HTML}{1F9E89}
\definecolor{runL128}{HTML}{35B779}
\definecolor{runL256}{HTML}{6CCE59}
\definecolor{runL512}{HTML}{A0DA39}
\definecolor{runL1024}{HTML}{C8C928}
\definecolor{runL2048}{HTML}{D79B00}
\definecolor{runscratchdeltaL64}{HTML}{1F9E89}
\definecolor{runscratchdeltaL256}{HTML}{6CCE59}
\definecolor{runscratchdeltaL1024}{HTML}{C8C928}
\definecolor{runscratchdeltaL2048}{HTML}{D79B00}
\definecolor{runpileL1}{HTML}{440154}
\definecolor{runpileL32}{HTML}{3E4989}
\definecolor{runpileL64}{HTML}{26828E}
\definecolor{runpileL128}{HTML}{1F9E89}
\definecolor{runpileL256}{HTML}{6CCE59}
\definecolor{runpileL512}{HTML}{C8C928}
\definecolor{runpersonaL4}{HTML}{440154}
\definecolor{runpersonaL8}{HTML}{1F9E89}
\definecolor{runpersonaL16}{HTML}{C8C928}
\definecolor{runsftL32}{HTML}{440154}
\definecolor{runsftL64}{HTML}{26828E}
\definecolor{runsftL128}{HTML}{6CCE59}
\definecolor{runsftL256}{HTML}{C8C928}
\definecolor{runrlvrL256}{HTML}{440154}
\definecolor{runrlvrL1024}{HTML}{C8C928}
\begin{tikzpicture}[trim axis group left,trim axis group right]
\begin{groupplot}[
  group style={group size=2 by 1, horizontal sep=0.15\textwidth, vertical sep=1.0em},
  width=0.35\textwidth,
  height=0.220\textwidth,
  scale only axis,
  title style={font=\fontsize{10}{11}\selectfont,align=center,yshift=-0.7ex},
  label style={font=\fontsize{8.5}{9.5}\selectfont},
  tick label style={font=\fontsize{8}{9}\selectfont,/pgf/number format/fixed,/pgf/number format/precision=2},
  tick align=outside,
  tick pos=left,
  axis line style={line width=0.4pt},
  major grid style={gray!20,line width=0.3pt},
  grid=major,
  enlarge x limits=0.06,
  enlarge y limits=0.10,
  scaled ticks=false,
  every axis plot/.append style={line join=round},
]
\nextgroupplot[title={Continuing-model performance},xlabel={$\lambda$},ylabel={Average gain (AvgGain)},xmin=0.2,xmax=1,xtick={0.2,0.4,0.6,0.8,1},enlarge x limits=false,clip marker paths=false]
\addplot[color=runL2048,line width=1.6pt,mark=none,error bars/error bar style={draw=runL2048,opacity=0.4,draw opacity=0.4,fill opacity=0.4,line width=0.75pt},error bars/error mark=none,error bars/.cd,y dir=both,y explicit] coordinates {(0.2,3.98873) +- (0,0.1204854) (0.4,7.465903) +- (0,0.05008733) (0.6,9.076079) +- (0,0.07519295) (0.8,9.524839) +- (0,0.2446566) (1,9.318089) +- (0,0.08005823)};
\addplot[only marks,color=runL2048,mark=*,mark size=1.6pt,mark options={line width=0pt}] coordinates {(0.2,3.98873) (0.4,7.465903) (0.6,9.076079) (1,9.318089)};
\addplot[only marks,color=runL2048,mark=star,mark size=3.2pt,line width=1.1pt] coordinates {(0.8,9.524839)};
\node[font=\fontsize{5}{5.5}\selectfont,text=runL2048!75!black,anchor=south west,xshift=2.5pt,yshift=1.5pt,fill=white,fill opacity=0.94,text opacity=1,inner sep=0.25pt] at (axis cs:0.2,3.98873) {$L=2048$};
\addplot[color=runL1024,line width=1.6pt,mark=none,error bars/error bar style={draw=runL1024,opacity=0.4,draw opacity=0.4,fill opacity=0.4,line width=0.75pt},error bars/error mark=none,error bars/.cd,y dir=both,y explicit] coordinates {(0.2,6.398515) +- (0,0.1277902) (0.4,9.168391) +- (0,0.03125222) (0.6,9.867759) +- (0,0.02639953) (0.8,10.00592) +- (0,0.02599152) (1,9.49335) +- (0,0.05598315)};
\addplot[only marks,color=runL1024,mark=*,mark size=1.6pt,mark options={line width=0pt}] coordinates {(0.2,6.398515) (0.4,9.168391) (0.6,9.867759) (1,9.49335)};
\addplot[only marks,color=runL1024,mark=star,mark size=3.2pt,line width=1.1pt] coordinates {(0.8,10.00592)};
\node[font=\fontsize{5}{5.5}\selectfont,text=runL1024!75!black,anchor=south west,xshift=2.5pt,yshift=1.5pt,fill=white,fill opacity=0.94,text opacity=1,inner sep=0.25pt] at (axis cs:0.2,6.398515) {$L=1024$};
\addplot[color=runL256,line width=1.6pt,mark=none,error bars/error bar style={draw=runL256,opacity=0.4,draw opacity=0.4,fill opacity=0.4,line width=0.75pt},error bars/error mark=none,error bars/.cd,y dir=both,y explicit] coordinates {(0.2,9.571753) +- (0,0.0367845) (0.4,10.67904) +- (0,0.08671609) (0.6,10.86784) +- (0,0.08652971) (0.8,10.74533) +- (0,0.09127955) (1,10.35047) +- (0,0.1040719)};
\addplot[only marks,color=runL256,mark=*,mark size=1.6pt,mark options={line width=0pt}] coordinates {(0.2,9.571753) (0.4,10.67904) (0.8,10.74533) (1,10.35047)};
\addplot[only marks,color=runL256,mark=star,mark size=3.2pt,line width=1.1pt] coordinates {(0.6,10.86784)};
\node[font=\fontsize{5}{5.5}\selectfont,text=runL256!75!black,anchor=south west,xshift=2.5pt,yshift=1.5pt,fill=white,fill opacity=0.94,text opacity=1,inner sep=0.25pt] at (axis cs:0.2,9.571753) {$L=256$};
\addplot[color=runL64,line width=1.6pt,mark=none,error bars/error bar style={draw=runL64,opacity=0.4,draw opacity=0.4,fill opacity=0.4,line width=0.75pt},error bars/error mark=none,error bars/.cd,y dir=both,y explicit] coordinates {(0.2,10.34209) +- (0,0.04660868) (0.4,11.03427) +- (0,0.05850622) (0.6,11.14283) +- (0,0.06750028) (0.8,11.12919) +- (0,0.07478223) (1,10.91779) +- (0,0.090042)};
\addplot[only marks,color=runL64,mark=*,mark size=1.6pt,mark options={line width=0pt}] coordinates {(0.2,10.34209) (0.4,11.03427) (0.8,11.12919) (1,10.91779)};
\addplot[only marks,color=runL64,mark=star,mark size=3.2pt,line width=1.1pt] coordinates {(0.6,11.14283)};
\node[font=\fontsize{5}{5.5}\selectfont,text=runL64!75!black,anchor=south west,xshift=2.5pt,yshift=1.5pt,fill=white,fill opacity=0.94,text opacity=1,inner sep=0.25pt] at (axis cs:0.2,10.34209) {$L=64$};
\addplot[color=runL1,line width=1.6pt,mark=none,error bars/error bar style={draw=runL1,opacity=0.4,draw opacity=0.4,fill opacity=0.4,line width=0.75pt},error bars/error mark=none,error bars/.cd,y dir=both,y explicit] coordinates {(0.2,11.16651) +- (0,0.05891838) (0.4,11.38067) +- (0,0.03728805) (0.6,11.42502) +- (0,0.02886191) (0.8,11.36416) +- (0,0.05531782) (1,11.34313) +- (0,0.05022235)};
\addplot[only marks,color=runL1,mark=*,mark size=1.6pt,mark options={line width=0pt}] coordinates {(0.2,11.16651) (0.4,11.38067) (0.8,11.36416) (1,11.34313)};
\addplot[only marks,color=runL1,mark=star,mark size=3.2pt,line width=1.1pt] coordinates {(0.6,11.42502)};
\node[font=\fontsize{5}{5.5}\selectfont,text=runL1!75!black,anchor=south west,xshift=2.5pt,yshift=1.5pt,fill=white,fill opacity=0.94,text opacity=1,inner sep=0.25pt] at (axis cs:0.2,11.16651) {$L=1$};
\addplot[black,densely dashed,line width=0.55pt] coordinates {(0,12.16924) (1,12.16924)};
\nextgroupplot[title={Stability vs. plasticity},xlabel={Stability (non-targeted gain)},ylabel={Plasticity (targeted gain)},xlabel style={align=center},legend style={at={(0.015,0.005)},anchor=south west,draw=none,fill=white,fill opacity=0.92,text opacity=1,font=\fontsize{5.25}{6}\selectfont,inner sep=0.6pt,column sep=0.25em,row sep=-0.4pt},legend columns=2,legend cell align={left},legend image post style={yshift=0.25ex}]
\addlegendimage{empty legend}
\addlegendentry{lowest $\lambda=0.2$}
\addlegendimage{only marks,black,mark=triangle*,mark size=2.5pt,mark options={line width=0pt,yshift=-0.15ex}}
\addlegendentry{$\lambda=1.0$}
\addplot[color=runL2048,line width=1.6pt,mark=none,error bars/error bar style={draw=runL2048,opacity=0.4,draw opacity=0.4,fill opacity=0.4,line width=0.75pt},error bars/error mark=none,error bars/.cd,x dir=both,x explicit,y dir=both,y explicit] coordinates {(3.940253,12.37589) +- (0.114064,0.03387757) (7.387639,12.51517) +- (0.04612992,0.04604422) (8.926423,12.53381) +- (0.09785042,0.05723352) (9.251918,12.38206) +- (0.2504908,0.2079877) (8.961698,12.52562) +- (0.08811926,0.05600135)};
\addplot[only marks,color=runL2048,mark=*,mark size=1.6pt,mark options={line width=0pt}] coordinates {(3.940253,12.37589) (7.387639,12.51517) (8.926423,12.53381) (9.251918,12.38206)};
\addplot[only marks,color=runL2048,mark=triangle*,mark size=2.8pt,mark options={line width=0pt}] coordinates {(8.961698,12.52562)};
\addplot[color=runL1024,line width=1.6pt,mark=none,error bars/error bar style={draw=runL1024,opacity=0.4,draw opacity=0.4,fill opacity=0.4,line width=0.75pt},error bars/error mark=none,error bars/.cd,x dir=both,x explicit,y dir=both,y explicit] coordinates {(6.345145,12.14556) +- (0.127041,0.09113371) (9.081732,12.36749) +- (0.03002244,0.01918893) (9.720419,12.3814) +- (0.02800153,0.03363718) (9.768786,12.44166) +- (0.03047735,0.0172077) (9.170016,12.40335) +- (0.06212254,0.01402742)};
\addplot[only marks,color=runL1024,mark=*,mark size=1.6pt,mark options={line width=0pt}] coordinates {(6.345145,12.14556) (9.081732,12.36749) (9.720419,12.3814) (9.768786,12.44166)};
\addplot[only marks,color=runL1024,mark=triangle*,mark size=2.8pt,mark options={line width=0pt}] coordinates {(9.170016,12.40335)};
\addplot[color=runL256,line width=1.6pt,mark=none,error bars/error bar style={draw=runL256,opacity=0.4,draw opacity=0.4,fill opacity=0.4,line width=0.75pt},error bars/error mark=none,error bars/.cd,x dir=both,x explicit,y dir=both,y explicit] coordinates {(9.53802,11.80936) +- (0.03728892,0.0380254) (10.61615,12.09306) +- (0.08948482,0.0628888) (10.76732,12.16203) +- (0.08953472,0.06237402) (10.59785,12.17287) +- (0.0940415,0.06801777) (10.15059,12.14939) +- (0.1083929,0.06877072)};
\addplot[only marks,color=runL256,mark=*,mark size=1.6pt,mark options={line width=0pt}] coordinates {(9.53802,11.80936) (10.61615,12.09306) (10.76732,12.16203) (10.59785,12.17287)};
\addplot[only marks,color=runL256,mark=triangle*,mark size=2.8pt,mark options={line width=0pt}] coordinates {(10.15059,12.14939)};
\addplot[color=runL64,line width=1.6pt,mark=none,error bars/error bar style={draw=runL64,opacity=0.4,draw opacity=0.4,fill opacity=0.4,line width=0.75pt},error bars/error mark=none,error bars/.cd,x dir=both,x explicit,y dir=both,y explicit] coordinates {(10.32159,11.34038) +- (0.0464091,0.04318219) (10.99497,11.71206) +- (0.05931978,0.04830174) (11.08255,11.82053) +- (0.06798889,0.06073337) (11.047,11.89681) +- (0.07684048,0.05675947) (10.81238,11.86655) +- (0.09323918,0.06180807)};
\addplot[only marks,color=runL64,mark=*,mark size=1.6pt,mark options={line width=0pt}] coordinates {(10.32159,11.34038) (10.99497,11.71206) (11.08255,11.82053) (11.047,11.89681)};
\addplot[only marks,color=runL64,mark=triangle*,mark size=2.8pt,mark options={line width=0pt}] coordinates {(10.81238,11.86655)};
\addplot[color=runL1,line width=1.6pt,mark=none,error bars/error bar style={draw=runL1,opacity=0.4,draw opacity=0.4,fill opacity=0.4,line width=0.75pt},error bars/error mark=none,error bars/.cd,x dir=both,x explicit,y dir=both,y explicit] coordinates {(11.15618,10.4567) +- (0.05869348,0.06818248) (11.36913,11.34207) +- (0.0331302,0.03664334) (11.41364,11.44111) +- (0.02377961,0.03054224) (11.35329,11.38545) +- (0.05304722,0.05780798) (11.33181,11.36295) +- (0.04564404,0.05211804)};
\addplot[only marks,color=runL1,mark=*,mark size=1.6pt,mark options={line width=0pt}] coordinates {(11.15618,10.4567) (11.36913,11.34207) (11.41364,11.44111) (11.35329,11.38545)};
\addplot[only marks,color=runL1,mark=triangle*,mark size=2.8pt,mark options={line width=0pt}] coordinates {(11.33181,11.36295)};
\end{groupplot}
\end{tikzpicture}

%% file: figures/results/pg19_general_gain.tex
\begingroup
\definecolor{pgNineteenLR1}{HTML}{440154}
\definecolor{pgNineteenLR2}{HTML}{3E4989}
\definecolor{pgNineteenLR3}{HTML}{1F9E89}
\definecolor{pgNineteenLR4}{HTML}{6CCE59}
\begin{tikzpicture}[trim axis right]
\begin{axis}[
    width=\dimexpr\linewidth-36pt\relax,
    height=0.54\linewidth,
    scale only axis,
    xmin=0.25, xmax=4.15,
    ymin=0.023, ymax=0.043,
    clip=true,
    xtick={0.4,0.8,1.2,1.6,2,2.4,3.2,4},
    ytick={0.025,0.030,0.035,0.040},
    xlabel={Effective learning rate $\lambda\eta$ ($\times10^{-5}$)},
    ylabel={Average gain (BPT)},
    label style={font=\fontsize{8}{9}\selectfont},
    ylabel style={at={(axis description cs:0,0.5)},yshift=36pt,
        anchor=north,inner sep=0pt,outer sep=0pt},
    tick label style={font=\fontsize{7.2}{8.2}\selectfont},
    yticklabel style={/pgf/number format/fixed,/pgf/number format/precision=3,
        /pgf/number format/fixed zerofill},
    tick align=outside,
    tick pos=left,
    axis line style={line width=0.4pt},
    major grid style={gray!20,line width=0.3pt},
    grid=major,
    scaled ticks=false,
    every axis plot/.append style={line join=round},
    legend style={at={(axis description cs:0.5,-0.32)},anchor=north,
        draw=none,fill=none,font=\fontsize{6.8}{7.8}\selectfont,
        legend columns=2,
        /tikz/every even column/.append style={column sep=0.35em}},
    legend cell align={left},
]
\pgfplotsinvokeforeach{1,2,3,4}{%
    \addlegendimage{color=pgNineteenLR#1,line width=1.6pt,
        mark=*,mark size=1.6pt,mark options={solid,line width=0pt}}
    \addlegendentry{$\eta=#1\times10^{-5}$}
}
\addlegendimage{black!55,densely dashed,line width=1.1pt,
    mark=*,mark size=1.4pt,mark options={solid,fill=white,line width=0.4pt}}
\addlegendentry{Full integration ($\lambda=1$)}
\addplot[color=black!55,densely dashed,line width=1.1pt,
    mark=*,mark size=1.4pt,
    mark options={solid,fill=white,line width=0.4pt},forget plot,
    error bars/error bar style={draw=black!55,opacity=0.4,line width=0.75pt},
    error bars/.cd,y dir=both,y explicit]
table[col sep=comma,x=effective_lr,y=gain,y error=sem]
    {figures/results/pg19/full_integration.csv};
\pgfplotsinvokeforeach{1,2,3,4}{%
    \addplot[color=pgNineteenLR#1,line width=1.6pt,
        mark=*,mark size=1.6pt,mark options={line width=0pt},forget plot,
        error bars/error bar style={draw=pgNineteenLR#1,opacity=0.4,
            draw opacity=0.4,line width=0.75pt},
        error bars/.cd,y dir=both,y explicit]
    table[col sep=comma,x=lr#1_effective_lr,y=lr#1_gain,y error=lr#1_sem]
        {figures/results/pg19/plot.csv};
}
\end{axis}
\end{tikzpicture}%
\endgroup%

%% file: sections/06_discussion.tex
\section{Conclusion and limitations}
\label{sec:conclusion}

\looseness-1We study task-vector scaling on temporally clustered, non-\iid~minibatch streams. Across five training regimes, partial integration often outperforms full integration, especially after longer same-distribution sequences. In domain- and book-level continual pretraining, it also outperforms matched learning-rate scaling.
The coefficient controls targeted-distribution adaptation versus non-targeted retention, and the benefits persist during pretraining from random initialization. 
Direct linear interpolation remains effective throughout our controlled long-SFT trajectory but can fail after pretraining-scale adaptation.

Our experiments cover limited models, data orders, and adaptation regimes. They assume each minibatch is \iid within one distribution and use a fixed integration coefficient~$\lambda$. Broader streams and adaptive coefficient selection remain future work.

%% file: sections/appendix.tex
\section{Extended related work}
\label[appendixsection]{app:extended-related-work}

\paragraph{Continual learning and forgetting}
The goal of continual learning (CL) is to enable a model to accumulate knowledge from a sequence of tasks while preserving capabilities that were obtained previously. 
While in conventional training data from all tasks are assumed to be jointly available, in CL the model is presented with a sequence or stream of task-specific data distributions and is commonly restricted to data from the current task. 
A CL method, therefore, must trade off between \emph{plasticity}, the ability to acquire new capabilities, and \emph{stability}, the ability to retain existing capabilities. 
Adaptation to a new distribution often degrades performance on previously learned tasks.
This phenomenon is known as \emph{catastrophic forgetting}
\citep{parisi2019continual,delange2022continual}.

Catastrophic forgetting is commonly explained by interference caused through sharing model parameters for multiple tasks. When different tasks rely on overlapping parameters or representations, updates that improve performance
on the current task are likely to modify parameters that contribute to
performance on previously learned tasks
\citep{mccloskey1989catastrophic,goodfellow2014empirical}. The severity of
forgetting also depends on the relationship between successive tasks and the
alignment of their induced updates \citep{doan2021theoretical,lee2021continual}.
Finally, optimization choices such as the learning rate and batch size can
substantially affect the model's sensitivity to subsequent updates
\citep{mirzadeh2020understanding}.

Methods for mitigating catastrophic forgetting can be broadly grouped into
replay, regularization, and parameter-isolation methods. Replay-based methods
keep a subset of previously observed data points and interleave them with
data from the current task, providing a direct learning signal for preserving
previously obtained capabilities. Regularization-based methods constrain the
model to preserve information important for previous tasks, either by
penalizing changes to important parameters or by constraining outputs or
hidden representations. Parameter-isolation methods reduce interference by
assigning separate parameters to different tasks, but may require knowledge
of the active task at inference time or cause the number of parameters to
grow with the number of tasks \citep{delange2022continual}.

Unlike models trained from scratch on a sequence of tasks, pretrained
language models begin with general capabilities acquired from initial
training on a large and diverse text dataset. Sequential adaptation can
therefore cause two forms of degradation: the model may forget task-specific
capabilities acquired during earlier adaptation stages
\citep{scialom2022fine}, and it may lose general capabilities inherited from
pretraining \citep{luo2023investigating}. Pretraining provides representations
that can be reused across tasks and can facilitate transfer during subsequent
adaptation. However, fine-tuning on narrow or non-\iid data distributions can
adapt the model to the current task at the expense of its
broader capabilities \citep{wang2024twostage,li2024revisiting}.

\paragraph{Weight averaging as a useful operation in weight space}
Early work on weight-space averaging showed that linear operations on neural
network weights can improve generalization. Stochastic Weight Averaging (SWA)
averages checkpoints along a single stochastic gradient descent (SGD)
trajectory, resulting in optima in wider regions of the loss landscape that
generalize well \citep{izmailov2018averaging}. While SWA is not model merging
in the multi-task sense, it provides early evidence that weight-space
averaging can be beneficial when checkpoints are sufficiently aligned.

A related line of work studies averaging of checkpoints from the same base model for robust fine-tuning. WiSE-FT interpolates between a pretrained zero-shot model and its fine-tuned checkpoint to trade off in-distribution (ID) adaptation and robustness under distribution shift \citep{wortsman2022robust}. 
Model Soups extend this idea by averaging multiple fine-tuned checkpoints, typically obtained from different sets of hyperparameters, improving both ID accuracy and robustness to distribution shift without increasing inference cost \citep{wortsman2022model}.

Recent work considers model merging as a way to combine capabilities for models that were fine-tuned on separate tasks from the same base model.
Fisher-weighted averaging weights parameters using the empirical Fisher Information, used as a local approximation to the curvature of each model's loss \citep{matena2022merging}.
Task arithmetic represents the parameter displacement through fine-tuning as a "task vector", computed by taking the difference between the adapted model and the shared pretrained initialization and shows that model capabilities can be composed by adding, subtracting, or scaling such vectors~\citep{ilharco2023editing}. 
These methods extend model merging in weight space beyond making fine-tuning more robust on a single task toward multi-task model merging, with the goal of integrating complementary capabilities learned on different tasks.

A central challenge in this setting is interference between task-specific updates, which especially occurs when multiple models are merged.
TIES-Merging addresses this issue by trimming updates with small magnitudes to resolve sign conflicts and by averaging only aligned parameter changes~\citep{yadav2023ties}. 
DARE, on the other hand, exploits redundancy in fine-tuning deltas by randomly dropping and rescaling parameter differences before merging~\citep{yu2024language}.

Most of this literature studies merging as a post-hoc operation over a fixed set of fine-tuned models. 
In contrast to this, we consider sequential adaptation to sequences, where each contains data sampled \iid from the targeted distribution but not from the overall data mixture.
Each sequence-specific local update must then be integrated into a continuing model. 
This changes the central question from \emph{how to merge a collection of experts} to \emph{how much of a sequence-specific update should become permanent} while preserving general performance.
More recent work applies model merging sequentially in continual-learning settings. We discuss these methods and their relation to our setting in~\cref{sec:related-work}.

\paragraph{Linear mode connectivity}
Since the loss of a deep neural network is highly non-linear in the model parameters, it is not obvious that linear interpolation between checkpoints should preserve model performance. 
Early work on mode connectivity showed that optima obtained through different independent training runs on the same data usually are connected through continuous paths in the loss landscape.
Garipov et al. found that such minima that share the same architecture can be connected by simple curves in weight space, such as polygonal chains or B\'ezier curves, along which both training and test loss remain low~\citep{garipov2018loss}. 
Draxler et al. observed similar behaviors using a nudged elastic band method to find continuous paths between minima, showing that apparent barriers between solutions can often be removed by following a \emph{non-linear} path through parameter space~\citep{draxler2018essentially}.

These results suggest that neural network solutions may belong to connected low-loss regions rather than isolated basins. Importantly, they do not imply that weights on the straight line between two checkpoints result in low loss.
This stronger property, known as \emph{linear mode connectivity}, is more directly relevant for linear model merging, since simple weight averaging evaluates points on the linear interpolation path
\begin{align}
    \theta(\lambda)
    =
    (1-\lambda)\theta_a+\lambda\theta_b,
    \qquad \lambda\in[0,1].
\end{align}

Frankle et al. connect linear mode connectivity to training stability~\citep{frankle2020linear}. 
They show that, after an early phase of shared training, independently continued runs from the same checkpoint become stable to SGD noise: even though the data order in the continued runs differ, the final models are connected by a low-loss linear path.
This is particularly relevant for pretrained models that form their representation from a very long initial phase of initial training. 
Thus, fine-tuned checkpoints share a long optimization history and differ only by the additional adaptation update. 
In such a setting, linear interpolation is therefore more plausible than in the case of independently trained models.

A second reason why linear connectivity may be more likely in modern language models is overparameterization. \cite{entezari2022role} and \cite{ainsworth2023git} suggest that sufficiently wide networks can admit low-barrier linear interpolation paths after their parameters are aligned.
In this view, model width and training time are important factors for the emergence of linear connectivity. 
Although this does not guarantee that arbitrary fine-tuned checkpoints can be merged safely, it suggests that large pretrained models are a promising setting for linear weight-space interpolation.

Together, these observations motivate linear model merging as a promising
integration rule for continual fine-tuning of language models. After each
non-\iid sequence, local adaptation produces an endpoint
$\theta_t^{(L)}$ that shares almost all of its training history with the
continuing model $\theta_t$. If the sequence update remains in a linearly
connected region, then partial integration may retain useful local
information at very low computational cost.

\paragraph{Meta-learning connections}
The task-vector integration rule considered in this work also appears in optimization-based meta-learning.
In particular, Reptile~\citep{nichol2018first} alternates between local adaptation and an outer meta update. 
Starting from parameters~$\theta$, it performs several SGD steps on a sampled task to obtain adapted parameters $\theta'$, and then applies
\begin{align}
    \theta
    \leftarrow
    \theta+\lambda(\theta'-\theta)
    =
    (1-\lambda)\theta+\lambda\theta'.
\end{align}
This update formulation has the same form as the method we study here: scaling the task vector obtained from local adaptation before integrating it into the continuing model.
However, the two settings assign different meanings to the update. 
Reptile uses it to learn an initialization that can be adapted quickly to a new task, whereas in our continual learning setting it determines how much of the current sequence-specific adaptation should be incorporated permanently.

Reptile is closely related to First-Order MAML (FOMAML)~\citep{finn2017model}, which uses the target-loss gradient at the final adapted parameters, whereas Reptile uses the displacement produced by the task-specific adaptation. 
As detailed in~\cref{app:sec:reptile-fomaml}, these directions are not identical. 
\cite{nichol2018first} show in a leading-order analysis that the connection between Reptile and FOMAML comes from shared within-task gradient-alignment terms.

The use of displacements collected over multiple sequential local updates also appears in federated learning. 
\texttt{FedAvg} performs several optimization steps on decentralized client data and aggregates the resulting client parameters at a server~\citep{mcmahan2017communication}. 
It addresses distributed aggregation across clients, whereas we apply scaled task vectors sequentially within one continuing non-\iid stream.

\section{Analysis}\label[appendixsection]{app:sec:analysis}

\subsection{Task-vector scaling versus learning-rate scaling}
\label[appendixsection]{app:sec:lr-scaling-analysis}

We analyze Task Vector Descent when the $L$ minibatch updates are performed
using stochastic gradient descent. At integration step $t$, the minibatches
$B_{t,1},\ldots,B_{t,L}$ contain samples from the same targeted distribution.
Starting from $\theta_t$, the trajectory is
\begin{align}
    \theta_t^{(0)}
    &=
    \theta_t,
    \\
    \theta_t^{(i)}
    &=
    \theta_t^{(i-1)}
    -
    \eta
    \underbrace{
        \nabla_{\theta^{(i-1)}_t}
        \mathcal{L}\left(\theta_t^{(i-1)};B_{t,i}\right)
    }_{\coloneqq g_i},
    \qquad i=1,\ldots,L.
\end{align}
The task vector is the accumulated displacement
\begin{align}
    \Delta_t
    =
    \theta_t^{(L)}-\theta_t
    =
    -\eta
    \sum_{i=1}^{L}
    g_i.
\end{align}

We approximate $g_i$ to second order around $\theta_t$:
\begin{align}
    g_i
    &=
    \bar{g}_i
    +
    \bar{H}_i
    \left(
        \theta_t^{(i-1)}-\theta_t
    \right)
    +
    \mathcal{O}\left(
        \left\lVert
            \theta_t^{(i-1)}-\theta_t
        \right\rVert^2
    \right),
\end{align}
where $\bar{g}_i\coloneqq\nabla_\theta\mathcal{L}(\theta_t;B_{t,i})$ and
$\bar{H}_i\coloneqq\nabla_\theta^2\mathcal{L}(\theta_t;B_{t,i})$. Assuming fixed
$L$ and bounded gradients,
$\theta_t^{(i-1)}-\theta_t=\mathcal{O}(\eta)$, and therefore
\begin{align}
    \theta_t^{(i-1)}-\theta_t
    &=
    -\eta
    \sum_{j=1}^{i-1}
    g_j
    \\
    &=
    -\eta
    \sum_{j=1}^{i-1}
    \bar{g}_j
    +
    \mathcal{O}(\eta^2).
\end{align}
Substitution into the Taylor expansion gives
\begin{align}
    g_i
    &=
    \bar{g}_i
    -
    \eta
    \bar{H}_i
    \sum_{j=1}^{i-1}
    \bar{g}_j
    +
    \mathcal{O}(\eta^2).
\end{align}

For any sequence $s$ and parameters $\theta$, we define its accumulated gradient
and sequential-update correction as
\begin{align}
    G_s(\theta)
    &\coloneqq
    \sum_{i=1}^{L}
    \nabla_{\theta}\mathcal{L}(\theta;B_{s,i}),
    \\
    C_s(\theta)
    &\coloneqq
    \sum_{i=1}^{L}
    \nabla_{\theta}^{2}\mathcal{L}(\theta;B_{s,i})
    \sum_{j=1}^{i-1}
    \nabla_{\theta}\mathcal{L}(\theta;B_{s,j}).
\end{align}
The correction $C_s(\theta)$ captures within-sequence path dependence.
Updates on earlier minibatches change the parameters $\theta$ thereby changing the gradients computed later.
The Hessian--gradient products correct for these changes to first order.

\paragraph{Task Vector Descent}
The task vector obtained at learning rate $\eta$ is
\begin{align}
    \Delta_t
    &=
    -\eta G_t(\theta_t)
    +
    \eta^2 C_t(\theta_t)
    +
    \mathcal{O}(\eta^3).
\end{align}
After scaling the task vector by $\lambda$, the continuing-model update is
\begin{align}
    \Delta_t^{\mathrm{TVD}}
    &\coloneqq
    \lambda\Delta_t
    =
    \lambda\left(\theta_t^{(L)}-\theta_t\right)
    \\
    &=
    -\lambda\eta
    \underbrace{G_t(\theta_t)}_{\text{accumulated gradient}}
    +
    \lambda\eta^2
    \underbrace{C_t(\theta_t)}_{\text{sequential-update correction}}
    +
    \mathcal{O}(\eta^3).
\end{align}

\paragraph{Learning-rate scaling}
For comparison, consider fully integrating the displacement produced by the $L$
minibatch updates using the smaller learning rate $\lambda\eta$. Starting from
$\widetilde{\theta}_t^{(0)}=\theta_t$, let
\begin{align}
    \widetilde{\theta}_t^{(i)}
    &=
    \widetilde{\theta}_t^{(i-1)}
    -
    \lambda\eta
    \nabla_\theta
    \mathcal{L}\left(\widetilde{\theta}_t^{(i-1)};B_{t,i}\right).
\end{align}
The resulting update is
\begin{align}
    \Delta_t^{\mathrm{LR}}
    &\coloneqq
    \widetilde{\theta}_t^{(L)}-\theta_t
    \\
    &=
    -\lambda\eta
    \underbrace{G_t(\theta_t)}_{\text{accumulated gradient}}
    +
    \lambda^2\eta^2
    \underbrace{C_t(\theta_t)}_{\text{sequential-update correction}}
    +
    \mathcal{O}(\eta^3).
\end{align}
This shows that task-vector scaling is not equivalent to reducing the learning rate. Both methods scale the accumulated gradient by $\lambda$, but task-vector scaling preserves the relative contribution of interactions generated along the original optimization trajectory, whereas learning-rate scaling suppresses them by an \emph{additional factor} of $\lambda$.

\subsection{Interactions across consecutive integration steps}
\label[appendixsection]{app:sec:cross_task_analysis}

We next distinguish interactions among the $L$ minibatches drawn from one
targeted distribution from interactions across two consecutive integration
steps. For $s\in\{t,t+1\}$, let $G_s$ and $C_s$ be the quantities defined
above and define the accumulated Hessian
\begin{align}
H_s(\theta)
&\coloneqq
\nabla_{\theta}G_s(\theta)
=
\sum_{i=1}^{L}
\nabla_{\theta}^{2}\mathcal{L}(\theta;B_{s,i}).
\end{align}
For a generic learning rate $\rho$, let $\Delta_s^{(\rho)}(\theta)$ denote
the displacement obtained by processing
$B_{s,1},\ldots,B_{s,L}$ from starting parameters $\theta$. The preceding
analysis gives
\begin{align}
\Delta_s^{(\rho)}(\theta)
&=
-\rho G_s(\theta)
+
\rho^2 C_s(\theta)
+
\mathcal{O}(\rho^3).
\end{align}

To derive both update rules at once, let $\kappa$ denote the coefficient
applied to each resulting displacement. Starting from $\theta_t$, two
consecutive updates are
\begin{align}
\theta_{t+1}^{(\kappa,\rho)}
&=
\theta_t
+
\kappa\Delta_t^{(\rho)}(\theta_t),
\\
\theta_{t+2}^{(\kappa,\rho)}
&=
\theta_{t+1}^{(\kappa,\rho)}
+
\kappa\Delta_{t+1}^{(\rho)}
\left(\theta_{t+1}^{(\kappa,\rho)}\right).
\end{align}
To express both displacements at the common reference point $\theta_t$, we
expand the second displacement as
\begin{align}
\Delta_{t+1}^{(\rho)}
\left(\theta_{t+1}^{(\kappa,\rho)}\right)
&=
\Delta_{t+1}^{(\rho)}(\theta_t)
+
\nabla_\theta\Delta_{t+1}^{(\rho)}(\theta_t)
\left(\theta_{t+1}^{(\kappa,\rho)}-\theta_t\right)
+
\mathcal{O}(\rho^3).
\end{align}
The remainder is $\mathcal{O}(\rho^3)$ because
$\theta_{t+1}^{(\kappa,\rho)}-\theta_t=\mathcal{O}(\rho)$ and
$D^2\Delta_{t+1}^{(\rho)}=\mathcal{O}(\rho)$. From the displacement expansion,
\begin{align}
\nabla_\theta\Delta_{t+1}^{(\rho)}(\theta_t)
&=
-\rho H_{t+1}(\theta_t)
+
\mathcal{O}(\rho^2),
\\
\theta_{t+1}^{(\kappa,\rho)}-\theta_t
&=
-\kappa\rho G_t(\theta_t)
+
\mathcal{O}(\rho^2).
\end{align}
Therefore,
\begin{align}
\Delta_{t+1}^{(\rho)}
\left(\theta_{t+1}^{(\kappa,\rho)}\right)
&=
\Delta_{t+1}^{(\rho)}(\theta_t)
+
\kappa\rho^2 H_{t+1}(\theta_t)G_t(\theta_t)
+
\mathcal{O}(\rho^3).
\end{align}
Training on sequence $t$ displaces the parameters from which sequence $t+1$ begins. 
The product $H_{t+1}(\theta_t)G_t(\theta_t)$ captures this dependence.
Substitution into the two-update recursion gives
\begin{align}
\theta_{t+2}^{(\kappa,\rho)}-\theta_t
= 
&-\kappa\rho\left(G_t+G_{t+1}\right) \\
&+ \kappa\rho^2\left(C_t+C_{t+1}\right)\\
&+ \kappa^2\rho^2 H_{t+1}G_t \\
&+ \mathcal{O}(\rho^3),
\end{align}
where all quantities on the right-hand side are evaluated at $\theta_t$.

\textbf{Task Vector Descent} corresponds to $(\kappa,\rho)=(\lambda,\eta)$, yielding
\begin{align}
\theta_{t+2}-\theta_t
=
&-\lambda\eta\left(G_t+G_{t+1}\right)\\
&+\textcolor{blue!80!cyan}{\boldsymbol{\lambda}}\eta^2
\left(C_t+C_{t+1}\right)\\
&+\lambda^2\eta^2 H_{t+1}G_t\\
&+\mathcal{O}(\eta^3).
\end{align}
\textbf{Learning-rate scaling} corresponds to $(\kappa,\rho)=(1,\lambda\eta)$,
yielding
\begin{align}
\widetilde{\theta}_{t+2}-\theta_t
=
&-\lambda\eta\left(G_t+G_{t+1}\right)\\
&+\textcolor{blue!80!cyan}{\boldsymbol{\lambda^{2}}}\eta^2
\left(C_t+C_{t+1}\right)\\
&+\lambda^2\eta^2 H_{t+1}G_t\\
&+\mathcal{O}(\eta^3).
\end{align}

The two methods have the same first-order update ($G_t+G_{t+1}$) and scale the cross-sequence interaction ($H_{t+1}G_t$) by $\lambda^2$. 
They differ only in the sequence-dependent correction ($C_t+C_{t+1}$), highlighted in~\textcolor{blue!80!cyan}{\textbf{blue}}. 

Task Vector Descent computes the local trajectory at the original learning rate before scaling the resulting displacement, so the correction term is multiplied by~$\lambda$. 
In contrast, learning-rate scaling scales every minibatch step, resulting in a factor of $\lambda^2$. 
Thus, Task Vector Descent preserves more within-sequence optimization structure without strengthening cross-sequence coupling (to second order).

\subsection{Optimal scaling under a local quadratic model}
\label[appendixsection]{app:sec:optimal-scaling-analysis}

The preceding analysis distinguishes task-vector scaling from learning-rate
scaling. We now ask when partial integration can improve the average objective
over full integration. Related curvature-based optimal coefficients have been derived for continual model merging \citep{qiu2025train,zhang2019lookahead}. 
Here, we apply the local quadratic perspective to the average population loss along a sequence task vector. 
Let
\begin{align}
F(\theta)
&\coloneqq
\mathbb{E}_k
\mathbb{E}_{z\sim P_k}
\left[
\mathcal{L}(\theta;z)
\right]
\end{align}
denote the average population loss from \cref{eq:average-objective}. Along the
task vector $\Delta_t$, a second-order expansion around $\theta_t$ gives
\begin{align}
F(\theta_t+\lambda\Delta_t)
&=
F(\theta_t)
+
\lambda\nabla F(\theta_t)^\top\Delta_t
+
\frac{\lambda^2}{2}
\Delta_t^\top\nabla^2F(\theta_t)\Delta_t
+
o\left(\lVert\Delta_t\rVert^2\right),
\label{eq:quadratic-scaling-objective}
\end{align}
The linear term is the first-order change in average loss along the task
vector, whereas the quadratic term captures the curvature in that direction.
If $\Delta_t^\top\nabla^2F(\theta_t)\Delta_t>0$, differentiating
\cref{eq:quadratic-scaling-objective} with respect to $\lambda$ gives the
constrained quadratic-model optimum
\begin{align}
\lambda_t^\star
&=
\operatorname{clip}_{[0,1]}
\left(
-\frac{
\nabla F(\theta_t)^\top\Delta_t
}{
\Delta_t^\top\nabla^2F(\theta_t)\Delta_t
}
\right).
\label{eq:optimal-task-vector-scaling}
\end{align}
In the case of $\nabla F(\theta_t)^\top\Delta_t\geq0$, moving along the task vector does not decrease the average loss even to first order, and the quadratic model prefers
$\lambda=0$. 
If this directional derivative is negative and $0 < \lambda^*_t < 1$, moving along task vector initially improves the average objective but full integration passes the minimum along that direction, making partial integration optimal. 
If the ratio is at least one, the quadratic-model minimum lies at or beyond the locally adapted endpoint, and full integration is optimal.

Under the simplifying assumption that longer same-distribution sequences primarily increase the task-vector magnitude while preserving its direction and the local geometry, the linear term grows with $\lVert\Delta_t\rVert$, whereas the curvature term grows with $\lVert\Delta_t\rVert^2$. 
The preferred coefficient, therefore, decreases approximately inversely with task-vector magnitude, providing a local explanation for why partial integration becomes more useful for longer sequences. 
Equation~\eqref{eq:optimal-task-vector-scaling} is an oracle characterization rather than an online selection rule because the average-loss gradient and curvature are unavailable from the targeted distribution alone.

\subsection{Relation between Reptile and FOMAML}
\label[appendixsection]{app:sec:reptile-fomaml}

Reptile's connection to FOMAML becomes clearer by expanding the local
optimization. Let
\begin{align}
    \theta^{(0)}
    &=
    \theta,
    \\
    \theta^{(i+1)}
    &=
    \theta^{(i)}
    -
    \alpha
    \nabla_\theta
    \mathcal{L}_{\tau,i}\left(\theta^{(i)}\right),
    \qquad i=0,\ldots,k-1,
\end{align}
where $\mathcal{L}_{\tau,i}$ denotes the loss on the minibatch used in local
step $i$. Let $\mathcal{L}^*_{\tau}$ denote the target loss evaluated after
adaptation. MAML minimizes
\begin{align}
    \mathcal{J}_{\mathrm{MAML}}(\theta^{(0)})
    =
    \mathbb{E}_{\tau}
    \left[
        \mathcal{L}^*_{\tau}\left(\theta^{(k)}\right)
    \right],
\end{align}
whose gradient is
\begin{align}
    \nabla_\theta^{(0)}
    \mathcal{J}_{\mathrm{MAML}}(\theta^{(0)})
    =
    \mathbb{E}_{\tau}
    \left[
        \left(
            \frac{\partial\theta^{(k)}}{\partial\theta^{(0)}}
        \right)^{\!\top}
        \frac{\partial}{\partial \theta^{(k)}}
        \mathcal{L}^*_{\tau}\left(\theta^{(k)}\right)
    \right].
    \label{eq:maml-exact-gradient}
\end{align}
FOMAML approximates the adaptation Jacobian by the identity, yielding the
meta-gradient
\begin{align}
    g_{\mathrm{FOMAML}}
    =
    \mathbb{E}_{\tau}
    \left[
        \nabla_{\theta^{(k)}}
        \mathcal{L}^*_{\tau}\left(\theta^{(k)}\right)
    \right].
\end{align}
Thus, FOMAML computes a gradient at the final adapted parameters using a
separate target batch. Reptile instead uses the expected local parameter
displacement,
\begin{align}
    g_{\mathrm{Reptile}}
    &\coloneqq
    -\frac{1}{\alpha}
    \mathbb{E}_{\tau}
    \left[
        \theta^{(k)}-\theta^{(0)}
    \right]
    \\
    &=
    \mathbb{E}_{\tau}
    \left[
        \sum_{i=0}^{k-1}
        \nabla_\theta
        \mathcal{L}_{\tau,i}\left(\theta^{(i)}\right)
    \right].
    \label{eq:reptile-displacement-gradient}
\end{align}
To make the corresponding first-order approximation explicit, define the
cumulative loss along the local trajectory as
\begin{align}
    \mathcal{J}_{\mathrm{trajectory}}(\theta)
    &\coloneqq
    \mathbb{E}_{\tau}
    \left[
        \sum_{i=0}^{k-1}
        \mathcal{L}_{\tau,i}\left(\theta^{(i)}\right)
    \right].
\end{align}
Its exact gradient reintroduces the adaptation Jacobian at every local step:
\begin{align}
    \nabla_\theta\mathcal{J}_{\mathrm{trajectory}}(\theta)
    &=
    \mathbb{E}_{\tau}
    \left[
        \sum_{i=0}^{k-1}
        \left(
            \frac{\partial\theta^{(i)}}{\partial\theta}
        \right)^{\!\top}
        \frac{\partial}{\partial\theta^{(i)}}
        \mathcal{L}_{\tau,i}\left(\theta^{(i)}\right)
    \right].
\end{align}
Approximating each adaptation Jacobian by the identity gives
\begin{align}
    \nabla_\theta\mathcal{J}_{\mathrm{trajectory}}(\theta)
    &\approx
    \mathbb{E}_{\tau}
    \left[
        \sum_{i=0}^{k-1}
        \nabla_{\theta^{(i)}}
        \mathcal{L}_{\tau,i}\left(\theta^{(i)}\right)
    \right]
    =
    g_{\mathrm{Reptile}}.
\end{align}
This has the same chain-rule form as the MAML gradient in
\cref{eq:maml-exact-gradient}, but replaces the terminal target loss with the
cumulative loss along the local trajectory. Reptile can therefore be
interpreted as a first-order gradient of this trajectory objective
\citep{huisman2024understanding}, which differs from the terminal target-loss
objective used by FOMAML. Their connection also appears in leading-order
expansions: both directions contain terms that encourage gradients from
different minibatches of the same task to align, although these terms receive
different weights~\citep{nichol2018first}.

\section{Complete method specification}
\label[appendixsection]{app:method}

This section provides the complete specification of the training procedure
introduced in \cref{sec:method}. We first describe the sequence-level
procedure shared across settings, followed by the AdamW updates and
optimizer-state integration. We then specify the local adaptation objectives
and linear parameter integration.

\subsection{Sequence-level training procedure}

Every run trains on a stream of $T$ sequences. In the fixed-distribution
settings, sequence $t$ contains $L$ consecutive minibatches
$B_{t,1},\ldots,B_{t,L}$ from the targeted distribution $P_{k_t}$. A minibatch
is the setting-specific atomic training unit: a packed-token minibatch for
language modelling, a minibatch of prompt--response pairs for supervised
fine-tuning, or a prompt minibatch with task-specific answer verifiers for
GRPO. For persona learning, one prompt interaction is the atomic unit. Thus,
$L$ always controls how long the model processes one targeted distribution or
collection context before an integration step, although the amount of
computation per unit differs between settings.

We denote the continuing checkpoint entering sequence $t$ by
$(\theta_t,\omega_t)$, where $\theta_t$ contains the model parameters and
$\omega_t$ the optimizer state. Starting from
$(\theta_t^{(0)},\omega_t^{(0)})=(\theta_t,\omega_t)$, the setting-specific
optimizer processes the $L$ training units and produces the locally adapted
endpoint $\theta_t^{(L)}$ and optimizer state $\omega_t^{(L)}$. We then
integrate both into the next continuing checkpoint using the coefficient
$\lambda$. For each sequence, we evaluate the continuing model before local
adaptation, the locally adapted endpoint, and the continuing model after
integration, as shown
in \cref{fig:evaluation-protocol}. \Cref{alg:task-vector-descent} summarizes
the complete procedure.

\begin{algorithm}[h]
    \caption{Continual training with scaled task-vector updates}
    \label{alg:task-vector-descent}
    \begin{algorithmic}[1]
        \Require Initial parameters $\theta_0$, optimizer state $\omega_0$,
        minibatches $B_{t,i}$, number $L$ of consecutive minibatches,
        coefficient $\lambda$
        \For{$t=0,\ldots,T-1$}
            \State $(\theta_t^{(0)},\omega_t^{(0)})\gets
                (\theta_t,\omega_t)$
            \State Record evaluation of $\theta_t$
            \For{$i=1,\ldots,L$}
                \State $(\theta_t^{(i)},\omega_t^{(i)})\gets
                    \operatorname{OptimizerStep}
                    (\theta_t^{(i-1)},\omega_t^{(i-1)};B_{t,i})$
            \EndFor
            \State $\Delta_t\gets\theta_t^{(L)}-\theta_t$
            \State Record evaluation of $\theta_t^{(L)}$
            \State $\theta_{t+1}\gets\theta_t+\lambda\Delta_t$
            \State $\omega_{t+1}\gets
                \operatorname{MergeAdamW}(\omega_t,\omega_t^{(L)};\lambda)$
            \State Record evaluation of $\theta_{t+1}$
        \EndFor
        \State \Return $(\theta_T,\omega_T)$
    \end{algorithmic}
\end{algorithm}

\subsection{AdamW local updates}
\label[appendixsection]{app:adamw-updates}

The update in~\cref{eq:local-adaptation} illustrates local adaptation using SGD without momentum.
We use AdamW~\citep{loshchilov2019decoupled} in our experiments.
At the start of sequence $t$, we inherit the full continuing checkpoint,
\begin{align}
    \left(\theta_t^{(0)},m_t^{(0)},v_t^{(0)},c_t^{(0)}\right)
    &=\left(\theta_t,m_t,v_t,c_t\right).
    \label{eq:app-adamw-initialization}
\end{align}
Here, $m$ and $v$ are the first and second gradient moments and $c$ is the
optimizer step count. For each executed optimizer step $i$, let $g_t^{(i)}$
be the gradient of the setting-specific loss on the effective minibatch,
after gradient accumulation and gradient-norm clipping where applicable.
For a parameter group with learning rate $\eta_t^{(i)}$, moment coefficients
$\beta_1,\beta_2\in[0,1)$, weight-decay coefficient $\rho\geq0$, and
numerical stabilizer $\epsilon>0$, the updates are
\begin{align}
    c_t^{(i)}
    &=c_t^{(i-1)}+1,
    \label{eq:app-adamw-count}
    \\
    m_t^{(i)}
    &=\beta_1m_t^{(i-1)}+(1-\beta_1)g_t^{(i)},
    \label{eq:app-adamw-first-moment}
    \\
    v_t^{(i)}
    &=\beta_2v_t^{(i-1)}+(1-\beta_2)\left(g_t^{(i)}\right)^2
    \label{eq:app-adamw-second-moment}
    \\
    \widehat m_t^{(i)}
    &=\frac{m_t^{(i)}}{1-\beta_1^{c_t^{(i)}}},
    \qquad
    \widehat v_t^{(i)}
    =\frac{v_t^{(i)}}{1-\beta_2^{c_t^{(i)}}},
    \label{eq:app-adamw-bias-correction}
    \\
    \theta_t^{(i)}
    &=\left(1-\eta_t^{(i)}\rho\right)\theta_t^{(i-1)}
    -\eta_t^{(i)}
    \frac{\widehat m_t^{(i)}}{\sqrt{\widehat v_t^{(i)}}+\epsilon}.
    \label{eq:app-adamw-parameters}
\end{align}
Squares and square roots are applied elementwise. 
The bias-correction factors use the inherited count~$c_t^{(i)}$, which is not reset at sequence boundaries and can be fractional after the optimizer-state interpolation.
After local adaptation, the resulting parameters and optimizer state are integrated as described in~\Cref{app:optimizer-state}. 
The learning rates, moment coefficients,
weight decay, and clipping settings are specified in~\cref{app:setup}.

\subsection{Optimizer-state integration}
\label[appendixsection]{app:optimizer-state}

Because the model parameters are linearly interpolated between the
continuing-model parameters $\theta_t$ and locally adapted endpoint
$\theta_t^{(L)}$, we apply the same interpolation to the optimizer state. In
AdamW, the optimizer state in the continuing checkpoint
$\omega_t=(m_t,v_t,c_t)$ is represented by the first moment, second moment,
and step count. Let
$\omega_t^{(L)}=(m_t^{(L)},v_t^{(L)},c_t^{(L)})$ denote the corresponding
state after processing the current sequence. We obtain the continuing
optimizer state by linearly interpolating the first moment and step count and
interpolating the second moment in root-mean-square space:
\begin{align}
    m_{t+1}
    & =
    (1-\lambda)m_t+\lambda m_t^{(L)},
    \label{eq:app-first-moment}
    \\
    c_{t+1}
    & =
    (1-\lambda)c_t+\lambda c_t^{(L)},
    \label{eq:app-step-count}
    \\
    v_{t+1}
    & =
    \left(
        (1-\lambda)\sqrt{v_t}
        +\lambda\sqrt{v_t^{(L)}}
    \right)^2.
    \label{eq:app-second-moment}
\end{align}
We use the same coefficient $\lambda$ to integrate model parameters and the
optimizer state to ensure consistency between the partially integrated
parameters and the AdamW state. All other optimizer hyperparameters and the
learning-rate schedule remain constant and require no special handling;
non-floating model-state entries remain unchanged. At $\lambda=1$, these
rules return the parameters and AdamW state produced by the local updates
exactly, reproducing the standard continual-training baseline.

\subsection{Local adaptation objectives}

We use four local adaptation objectives: next-token language modelling for
continual pretraining and pretraining from random initialization, supervised
fine-tuning to teach the model chain-of-thought reasoning, SDPO to adapt to a
user's persona, and GRPO to learn from verifiable rewards. The corresponding
data and hyperparameters are specified in \cref{app:setup}.

\paragraph{Next-token language modelling}
In language modelling, a minibatch $B$ contains packed token sequences
$z=(z_1,\ldots,z_{|z|})$. Training is performed by minimizing the causal
negative log likelihood
\begin{align}
    \mathcal{L}_{\mathrm{LM}}(\theta;B)
    & =
    -\frac{1}{n_B^{\mathrm{tok}}}
    \sum_{z\in B}\sum_{j=2}^{|z|}
    \log\pi_\theta(z_j\mid z_{<j}),
    \label{eq:app-lm}
    \\
    n_B^{\mathrm{tok}}
    & =
    \sum_{z\in B}(|z|-1).
    \notag
\end{align}
The loss is normalized by the number of target tokens. Each minibatch
corresponds to one AdamW update; therefore, $L$ equals both the number of
minibatches and the number of gradient steps per sequence in the pretraining
settings.

\paragraph{Reasoning with supervised fine-tuning}
For reasoning SFT, each example consists of a prompt $x$ and an expert
response $y=(y_1,\ldots,y_{|y|})$, often sourced from domain experts or a
stronger teacher model. The response contains a reasoning trace followed by a
final answer. Training is performed by minimizing the cross-entropy on the
completion tokens $y$ only:
\begin{align}
    \mathcal{L}_{\mathrm{SFT}}(\theta;B)
    & =
    -\frac{1}{M_B}
    \sum_{(x,y)\in B}\sum_{j=1}^{|y|}
    \log\pi_\theta(y_j\mid x,y_{<j}),
    \label{eq:app-sft}
    \\
    M_B
    & =
    \sum_{(x,y)\in B}|y|.
    \notag
\end{align}
Here, the prompt tokens $x$ serve as the conditioning context but are masked
from the loss computation. Unlike the on-policy objectives below, SFT trains
on fixed off-policy traces rather than on responses generated by the model.

\paragraph{Persona learning with SDPO}
In this setting, the goal is to adapt a policy to a user's preferences. The
preferences are only implicitly contained in the user's textual feedback.
First, the policy generates a response $y$ to a user prompt $x$. A user
simulator conditioned on the persona targeted in the current sequence then
returns either \texttt{ACCEPT} or textual feedback $f$. If the response is
accepted, no gradient update is performed. If the response is rejected, the
previous response is scored under the feedback-conditioned distribution
$\pi_\theta(\cdot\mid x,f,y_{<j})$. The loss is the token-averaged reverse KL
divergence
\begin{align}
    \mathcal{L}_{\mathrm{SDPO}}(\theta;x,y,f)
    & =
    \frac{1}{|y|}\sum_{j=1}^{|y|}
    D_{\mathrm{KL}}\!\left(
        \widetilde{\pi}_\theta(\cdot\mid x,y_{<j})
        \,\middle\Vert\,
        \operatorname{sg}\!\left[
            \widetilde{\pi}_\theta(\cdot\mid x,f,y_{<j})
        \right]
    \right).
    \label{eq:app-sdpo}
\end{align}
Here, $\operatorname{sg}$ is the stop-gradient operation and
$\widetilde{\pi}$ denotes the compressed token distribution that retains the
union of the 20 most likely tokens under the student and feedback-conditioned
distributions at each position and aggregates the remaining probability mass
into one tail bin. Intuitively, this objective compresses the simulator's
feedback into the feedback-free student policy while keeping the
student-generated response fixed as the states on which both distributions
are compared.

\paragraph{Reasoning with GRPO}
For GRPO \citep{shao2024deepseekmath}, the policy first generates a group of $G$ answers $y^{(j)}$ for a
given question $x$. Each answer is then verified and assigned the reward
\begin{align}
    r_j
    & =
    \mathbf{1}[\operatorname{correct}(y^{(j)},y^*)]
    +\beta_{\mathrm{fmt}}
    \mathbf{1}[\operatorname{valid\_format}(y^{(j)})].
    \label{eq:app-grpo-reward}
\end{align}
Here, $y^*$ is the reference answer and $\beta_{\mathrm{fmt}}=0.1$ weights the answer-format reward.
Correctness and validity of the answer format are determined by the
task-specific verifier and format requirements. The rewards are normalized
relative to the other responses in the group:
\begin{align}
    \widehat{A}_j
    & =
    \frac{
        r_j-\frac{1}{G}\sum_{h=1}^{G}r_h
    }{
        \operatorname{std}(r_1,\ldots,r_G)+\delta
    },
    \label{eq:app-grpo-advantage}
\end{align}
where $\delta>0$ ensures numerical stability. Let
\begin{align}
    \rho_{j,q}(\theta)
    & =
    \frac{
        \pi_\theta(y_q^{(j)}\mid x,y_{<q}^{(j)})
    }{
        \pi_{\theta_{\mathrm{old}}}
        (y_q^{(j)}\mid x,y_{<q}^{(j)})
    }
    \label{eq:app-grpo-ratio}
\end{align}
denote the importance ratio between the updated and rollout policies. We use
a DAPO-style clipped policy loss with active-token normalization \citep{yu2025dapo}:
\begin{align}
    \mathcal{L}_{\mathrm{GRPO}}(\theta;B)
    & =
    -\frac{1}{N_B^{\mathrm{tok}}}
    \sum_{x\in B}
    \sum_{j=1}^{G}
    \sum_{q=1}^{|y^{(j)}|}
    \min\!\Big(
        \rho_{j,q}(\theta)\widehat{A}_j,
        \operatorname{clip}\!\left(
            \rho_{j,q}(\theta),
            1-\epsilon_{\mathrm{low}},
            1+\epsilon_{\mathrm{high}}
        \right)\widehat{A}_j
    \Big),
    \label{eq:app-grpo-objective}
\end{align}
where $N_B^{\mathrm{tok}}$ is the number of active completion tokens. We use
eight on-policy rollouts per prompt, $\epsilon_{\mathrm{low}}=0.2$, and
$\epsilon_{\mathrm{high}}=0.28$. We use neither a reference-model KL penalty
nor a reasoning-length reward.

\subsection{Linear parameter integration}

Each of the four adaptation procedures produces the locally adapted endpoint and
optimizer state $(\theta_t^{(L)},\omega_t^{(L)})$. The local update results in
a parameter displacement between the locally adapted endpoint and the
continuing-model parameters,
\begin{align}
    \Delta_t
    & =
    \theta_t^{(L)}-\theta_t.
    \label{eq:app-task-vector}
\end{align}
We integrate the task vector by moving the parameters $\theta_t$ in
the direction of this displacement:
\begin{align}
    \theta_{t+1}
    & =
    \theta_t+\lambda\Delta_t
    \notag
    \\
    & =
    (1-\lambda)\theta_t+\lambda\theta_t^{(L)},
    \qquad \lambda\in[0,1].
    \label{eq:app-linear-integration}
\end{align}
Setting $\lambda=0$ discards the local update, whereas full integration
($\lambda=1$) carries the locally adapted endpoint forward without
attenuation. As discussed in
\cref{app:sec:reptile-fomaml}, this update is equivalent to a Reptile-style
meta-update. We use it here as a general integration mechanism rather than as
a meta-learning algorithm.

Linear parameter integration requires only one additional hyperparameter, the
coefficient $\lambda$. It also requires the continuing checkpoint to be
retained while processing the current sequence.

\section{Complete experimental configurations}
\label[appendixsection]{app:setup}

This section specifies the data, models, hyperparameters, stream construction,
and evaluation protocol used to instantiate the methods from
\cref{sec:method}. We study task-vector scaling in six settings: continual
pretraining, persona learning with SDPO, reasoning SFT, reasoning GRPO,
pretraining from random initialization, and book-level continual pretraining. \Cref{app:tab:experiments} summarizes
the experimental conditions.

\begin{table*}[t]
    \caption{Linear-integration experiments. In the
    fixed-distribution settings, $L$ counts consecutive minibatches from one
    targeted distribution. In the on-policy extensions, it counts consecutive
    setting-specific collection units from one active context.
    For PG-19, sequence lengths vary with book length.}
    \label{app:tab:experiments}
    \centering
    \scriptsize
    \setlength{\tabcolsep}{3.5pt}
    \renewcommand{\arraystretch}{1.12}
    \begin{tabular}{@{}lllllll@{}}
        \toprule
        Setting & Model & $K$ & $L$ & $\lambda$ & \# Seeds & Primary score \\
        \midrule
        Continual PT & GPT-2 & 10 & $1,32,64,128,256,512$ & $.4,.6,.8,1$ & 3 & BPT gain \\
        Persona SDPO & Qwen3-4B & 6 & $4,8,16$ & $0,.01,.05,.1,.2,.4,.6,.8,1$ & 3 & Macro win rate \\
        Reasoning SFT & Qwen2.5-3B & 4 & $32,64,128,256$ & $0,.2,.4,.6,.8,1$ & 5 & Best\@4 \\
        Reasoning GRPO & Qwen2.5-3B & 4 & $256,1024$ & $0,.2,.4,.6,.8,1$ & 3 & Best\@4 \\
        Scratch PT & GPT-2 & 10 & $1,64,256,1024,2048$ & $.2,.4,.6,.8,1$ & 3 & BPT gain \\
        Book-level PT & SmolLM-360M & 200 & $1$--$158$ & $.4,.6,.8,1$ & 3 & BPT gain \\
        \bottomrule
    \end{tabular}
\end{table*}

\subsection{Common protocol}

The experiments share a common training and evaluation setup. Except for
PG-19, the selected training examples are fixed across seeds. The seeds
instead determine the sequence order, training randomness, and, where
applicable, stochastic generation. Each sequence contains training units from
a single targeted distribution or collection context. Within a run, we use a
static integration coefficient $\lambda$ for all sequences.

\begin{figure}[htbp]
    \centering
    \begin{minipage}{0.6\linewidth}
        \input{figures/controlled_streams_figure}
    \end{minipage}
\end{figure}

After local adaptation, we integrate the model parameters according to
\cref{eq:linear-integration} and the AdamW optimizer state according to
\cref{eq:app-first-moment,eq:app-step-count,eq:app-second-moment}. We evaluate
the model at three points around each sequence: before local adaptation, after
local adaptation but before integration, and after integration. These
evaluations are purely observational and do not affect training or the choice
of $\lambda$.

\paragraph{Aggregation and uncertainty}
At each evaluation checkpoint, we first average uniformly over the evaluated
examples within each distribution and then compute an unweighted macro-average
across distributions. Unless stated otherwise, the run-level estimate is the
mean over all evaluated sequences. This aggregation prevents distributions
with more evaluation examples and the final sequence from disproportionately
influencing the reported result. We aggregate independent runs using the
arithmetic mean and report the sample standard error of the mean (SEM),
\begin{align}
    \operatorname{SEM}
    & =
    \frac{s}{\sqrt{n}},
    \label{eq:app-setup-sem}
\end{align}
where $s$ is the sample standard deviation of the $n$ run-level estimates.

\paragraph{Evaluation metrics}
For language modelling, we report bits per token (BPT). Given the mean
token-level negative log likelihood $\mathcal{L}_{\mathrm{token}}$ in nats,
BPT is defined as
\begin{align}
    \operatorname{BPT}
    & =
    \frac{\mathcal{L}_{\mathrm{token}}}{\ln 2}.
    \label{eq:app-setup-bpt}
\end{align}
Because lower BPT indicates better performance, we use
$\operatorname{score}(\theta;k)=-\operatorname{BPT}_k(\theta)$, consistent
with the higher-is-better convention in \cref{sec:method}. The resulting gain
is the BPT reduction relative to the initial checkpoint $\theta_0$:
\begin{align}
    \operatorname{gain}(\theta;k)
    & =
    \operatorname{BPT}_k(\theta_0)
    -
    \operatorname{BPT}_k(\theta).
    \label{eq:app-setup-bpt-gain}
\end{align}
Thus, positive values indicate an improvement over the initial checkpoint. We
use the same token-normalized metric for continual pretraining and pretraining
from random initialization.

For the generative reasoning tasks, we sample four independent responses per
prompt. Best@4 measures whether at least one of the four responses is correct,
whereas Mean@4 is the mean verifier correctness across the four responses. We use Best@4 as the metric for SFT and GRPO.

\paragraph{Frameworks}
We use Transformers and TRL for model training and vLLM for model inference
and evaluation \citep{wolf2020transformers,vonwerra2020trl,kwon2023vllm}.

\paragraph{Existing assets}
\Cref{tab:asset-licenses} lists the models, datasets, and principal software
used in our experiments and their published licenses or terms.
\input{sections/asset_licenses}

\subsection{Continual pretraining}

In the continual-pretraining setting, we adapt a pretrained language model to
a non-\iid stream of text domains and study whether task-vector scaling
preserves performance across domains while incorporating each local update.

\paragraph{Model}
We use a pretrained 124M-parameter GPT-2 checkpoint
\citep{radford2019language}, originally pretrained by OpenAI on WebText. We
perform training in FP32.

\paragraph{Data and distributions}
We select ten domains from the Pile Uncopyrighted dataset
\citep{gao2022pile}: \textit{PubMed Central, ArXiv, GitHub, FreeLaw,
StackExchange, USPTO Backgrounds, Gutenberg (PG-19), English Wikipedia,
Hacker News, and NIH ExPorter}.

We first split each domain into disjoint training and evaluation sets at the
document level. We then tokenize the documents using the GPT-2 tokenizer and
separate adjacent documents with an end-of-sequence token. The resulting token
sequences are packed into blocks of 512 tokens.

For each domain, we use 4,096 blocks for training. With a minibatch size of
eight blocks, this yields 512 training minibatches per domain and 5,120
optimizer updates across all ten domains. For evaluation, we use 16
minibatches of eight held-out blocks per domain, corresponding to 128
evaluation blocks per domain.

\paragraph{Streams}
From the selected training blocks, we construct streams with varying degrees
of temporal locality controlled by $L$. Each training minibatch contains eight
blocks from a single domain, and each sequence consists of $L$ consecutive
minibatches from that domain. For each domain, the blocks are shuffled once
during preprocessing and grouped into 512 fixed minibatches. For a given $L$,
we partition this minibatch sequence into consecutive groups of $L$ and
randomize the order of the resulting sequences for each experimental seed,
while preserving the minibatch order within each sequence. We consider
\begin{align}
    L
    & \in
    \{1,32,64,128,256,512\}.
    \label{eq:app-continual-l}
\end{align}

As an \iid reference, we globally shuffle the same 40,960 training blocks and
form mixed-domain minibatches of size eight. This reference uses the same
initialization, data, optimizer settings, and number of updates, but trains
continuously without parameter or optimizer-state integration. It differs
from the $L=1$, $\lambda=1$ condition, in which every minibatch remains
domain-specific and an integration step is performed after each update.

\paragraph{Local training and integration}
We optimize the next-token language-modelling objective in
\cref{eq:app-lm} using AdamW with a learning rate of $4\times10^{-5}$ and a
weight decay of $0.01$. We select the learning rate through a grid search on
the \iid reference and use the selected value for all non-\iid runs. We apply
gradient-norm clipping at $1$ and use a constant learning-rate schedule.

After every sequence, we integrate both the model parameters and the optimizer
state. We use the same static coefficient throughout a run and consider
\begin{align}
    \lambda
    & \in
    \{0.4,0.6,0.8,1\}.
    \label{eq:app-continual-lambda}
\end{align}

\paragraph{Evaluation and seeds}
At each sequence boundary, we evaluate all three checkpoints shown in
\cref{fig:evaluation-protocol} on all ten domains using 16 held-out
minibatches per domain. We report the macro-average BPT gain across domains
and average the run-level result over all evaluated sequences. We run both the
non-\iid experiments and the \iid reference with seeds $0$, $1$, and $2$.

\subsection{Persona learning with SDPO}

In the persona-learning setting, we adapt a language model to user-specific
style preferences conveyed through textual feedback. Each persona defines one
collection context, and SDPO performs local adaptation from interactions with
a persona-conditioned user simulator. We study whether task-vector scaling can
incorporate the learned preference while retaining performance on the other
personas.

\paragraph{Models}
We use Qwen3-4B as the model undergoing continual adaptation
\citep{yang2025qwen3}. A single Qwen3-8B model is used both as the
persona-conditioned user simulator during training and as the pairwise judge
during evaluation. The roles are defined through distinct system prompts. All
model computations use BF16.

\paragraph{Data and personas}
Following \citet{buening2026aligning}, we construct the training data from
prompts in the HelpSteer training split \citep{wang2023helpsteer}. We randomly
select 32 prompts and reuse the same set for every persona. We define six
personas, each corresponding to one style preference:
\begin{itemize}
    \item \textit{No preface:} You dislike unnecessary opening prefaces such
    as ``Certainly,'' ``Sure,'' ``Absolutely,'' or ``Here's \ldots''. You
    prefer responses to start directly with the answer.
    \item \textit{Brief preface:} You prefer responses to begin with a brief
    friendly preface such as ``Sure,'' ``Certainly,'' ``Of course,'' or
    ``Here's \ldots'' before the main answer.
    \item \textit{No opening praise:} You dislike opening praise or positive
    acknowledgements. Start directly with the answer instead of saying things
    like ``Great question'' or ``Nice idea.''
    \item \textit{Opening praise:} You prefer responses to begin with a brief
    positive acknowledgement or praise before answering, such as ``Great
    question'' or ``Nice idea.''
    \item \textit{No emojis:} You prefer responses without emojis.
    \item \textit{Short, precise sentences:} You prefer short, precise
    sentences. Avoid long multi-clause sentences and split ideas into separate
    sentences when possible.
\end{itemize}
This yields $6\times32=192$ interactions in total.

\paragraph{Streams and \iid reference}
Each prompt interaction constitutes one training unit. The stream contains
each of the 32 selected prompts once for each of the six personas. Each
sequence contains $L$ interactions associated with a single persona, and we
randomize the sequence order for each experimental seed. We consider
\begin{align}
    L
    & \in
    \{4,8,16\}.
    \label{eq:app-persona-l}
\end{align}
As an \iid reference, we randomly shuffle the 192 prompt--persona pairs and
use one-interaction sequences with $L=1$ and $\lambda=1$.

\paragraph{Local training and integration}
We perform local adaptation using the SDPO objective in
\cref{eq:app-sdpo}. For each interaction, the student generates a response
greedily with a maximum of 256 new tokens. If the simulator returns
\texttt{ACCEPT}, we perform no parameter update. Otherwise, it returns a short
corrective user message, which we use as feedback for three SDPO updates with
the feedback-conditioned distribution as the self-distillation target.

We optimize the student using AdamW with a learning rate of
$5\times10^{-6}$, weight decay $0.01$, a minibatch size of one, and
gradient-norm clipping at $1$. The maximum context length is 4,096 tokens.
Both the simulator and judge use greedy decoding. After every sequence, we
integrate both the model parameters and optimizer state and consider
\begin{align}
    \lambda
    & \in
    \{0,0.01,0.05,0.1,0.2,0.4,0.6,0.8,1\}.
    \label{eq:app-persona-lambda}
\end{align}

\paragraph{Evaluation and seeds}
We evaluate on 128 held-out prompts from the HelpSteer validation split. For
every prompt and persona, we generate responses from the current checkpoint
and the initial Qwen3-4B checkpoint. The evaluated models receive no system
prompt; each validation prompt is provided as a user message. The judge
receives the corresponding persona description in its system message.

For each prompt, the judge compares the current and initial model responses
in both presentation orders. We record a win or loss only when the two
judgments consistently favor the same response after accounting for
presentation order. Explicit ties and disagreements between the two
judgments are treated as ties.

For persona~$k$, let $W_k$, $L_k$, and $T_k$ denote the numbers of wins,
losses, and ties. We compute the conditional win rate as
$100W_k/(W_k+L_k)$, excluding ties from the denominator, and macro-average
these rates across personas. This metric measures preference for the current
model conditional on a decisive comparison. If $W_k+L_k=0$, the persona
score is undefined and excluded from the macro-average; this case did not
occur in the evaluated runs.
We report gains in percentage points relative to the initial model's measured persona scores, rather than assuming a baseline win rate of $50\%$. For each non-\iid run, the baseline is evaluated before the first adaptation sequence.
We run both the non-\iid experiments and the
\iid reference with seeds $0$, $1$, and $2$.

\subsection{Shared reasoning suite}

The SFT and GRPO experiments use the same model, reasoning tasks, and
evaluation protocol. We describe these shared components before specifying
the training procedure and hyperparameters for each adaptation objective.

\paragraph{Model}
We use Qwen2.5-3B-Instruct \citep{yang2024qwen25} and perform all model
computations in BF16.

\paragraph{Data and tasks}
The experiments use four procedurally generated tasks from Reasoning Gym~\citep{stojanovski2025reasoninggym}:
\begin{itemize}
    \item \textbf{Countdown:} combine exactly three input numbers using
    arithmetic operations to obtain a specified target value.
    \item \textbf{Mini-Sudoku:} complete a Sudoku puzzle containing exactly
    seven empty cells.
    \item \textbf{N-Queens:} complete a $4\times4$ board from which one queen
    has been removed.
    \item \textbf{String Manipulation:} repeatedly transform an initial
    string of length ten according to three ordered rules until no rule applies
    or a state repeats.
\end{itemize}
Each task provides a task-specific verifier that evaluates the correctness of
the extracted answer. A response is counted as correct only if its final
answer appears within complete \texttt{<answer>...</answer>} tags. The
acceptable representation inside these tags is determined by the
task-specific verifier.

\paragraph{Evaluation}
We evaluate on 100 held-out prompts per task and sample four responses per
prompt. We generate responses with a temperature of $0.6$, top-$p=0.95$,
top-$k=20$, and a maximum response length of 2,048 new tokens. We report Best@4.

\subsection{Reasoning with supervised fine-tuning}

\paragraph{Data}
Using data seed 42, we select 256 training examples for each of the four
reasoning tasks, giving 1,024 examples in total. For every prompt, we use
Claude Sonnet 4.6 \citep{anthropic2026sonnet46} to generate a fixed reasoning
trace followed by a final answer in \texttt{<answer>...</answer>} tags. All
1,024 generated traces pass the corresponding task-specific correctness
verifier and contain a valid answer tag.

\paragraph{Streams and \iid reference}
Each prompt--completion pair corresponds to one training unit. The stream
contains 256 examples from each task, and each sequence contains examples from
a single task. For each experimental seed, we shuffle the examples within
each task and randomize the order of the resulting sequences. We sample
without replacement, so every selected example occurs once in the stream. We
consider
\begin{align}
    L
    & \in
    \{32,64,128,256\}.
    \label{eq:app-sft-l}
\end{align}
As an \iid reference, we shuffle all 1,024 examples across tasks and use
one-example sequences with $L=1$ and $\lambda=1$.

\paragraph{Local training and integration}
We perform local adaptation using the completion-only SFT objective in
\cref{eq:app-sft}. Prompt tokens are excluded from the loss, so the model is
trained only on the expert-generated completion. We select the learning rate
through a grid search on the \iid reference and use the selected value of
$5\times10^{-5}$ for all non-\iid runs. We optimize the model using AdamW with
zero weight decay, a minibatch size of one, and a constant learning-rate
schedule. We use a maximum sequence length of 4,096 tokens.

After every sequence, we integrate both the model parameters and AdamW state.
We use the same static coefficient throughout a run and consider
\begin{align}
    \lambda
    & \in
    \{0,0.2,0.4,0.6,0.8,1\}.
    \label{eq:app-sft-lambda}
\end{align}
We run both the non-\iid experiments and the \iid reference with seeds $0$,
$1$, $2$, $3$, and $4$.

\subsection{Reasoning with GRPO}

\paragraph{Data}
For each reasoning task, we generate 4,096 training prompts and arrange them
into 1,024 fixed minibatches of four prompts. The complete training set
therefore contains 16,384 prompts across the four tasks.

\paragraph{Streams and \iid reference}
Each four-prompt minibatch constitutes one training unit. In the non-\iid
experiments, every minibatch and sequence contains prompts from a single
reasoning task. The task-balanced stream contains 1,024 minibatches per task,
sampled without replacement. For each experimental seed, we randomize the
order of the resulting sequences and avoid adjacent sequences from the same
task whenever possible.

For the \iid reference, we shuffle all 16,384 prompts across tasks, form
mixed-task minibatches of four prompts, and use one-minibatch sequences with
$L=1$ and $\lambda=1$.

\paragraph{Local training and integration}
We perform local adaptation using the DAPO-style clipped policy objective in
\cref{eq:app-grpo-objective}. For each minibatch of four prompts, we sample
eight on-policy rollouts per prompt, yielding a rollout minibatch of 32
completions. We use token-level importance ratios with clipping parameters
$\epsilon_{\mathrm{low}}=0.2$ and $\epsilon_{\mathrm{high}}=0.28$. Rewards
are normalized within each group of eight rollouts, and the loss is normalized
by the number of active completion tokens in the globally accumulated
minibatch.

We optimize the model using AdamW with a learning rate of $10^{-6}$, zero
weight decay, and a constant learning-rate schedule. During training, we
generate completions with temperature $1$, top-$p=0.95$, top-$k=20$, and a
maximum of 2,048 new tokens. We use neither a reference-model KL penalty nor a
reasoning-length reward. We consider
\begin{align}
    L
    & \in
    \{256,1024\}
    \quad\text{prompt minibatches},
    \label{eq:app-grpo-l}
    \\
    \lambda
    & \in
    \{0,0.2,0.4,0.6,0.8,1\}.
    \label{eq:app-grpo-lambda}
\end{align}
After every sequence, we integrate both the model parameters and AdamW state.
We use the shared reasoning evaluation protocol and run both the non-\iid
experiments and the \iid reference with seeds $0$, $1$, and $5$.

\subsection{Pretraining from random initialization}

\paragraph{Model}
We use the nanoGPT implementation of a decoder-only Transformer with the
dimensions of GPT-2 small \citep{radford2019language}: 12 layers, 12 attention
heads, and a hidden width of 768. Unlike the released GPT-2 checkpoint used
for continual pretraining, this model is initialized randomly. The
implementation uses a vocabulary padded to 50,304 entries, a context length
of 1,027 tokens, no dropout, and no bias parameters. The resulting model has
123.6 million parameters. We perform training in BF16.

\paragraph{Data and distributions}
We select ten domains from the Pile Uncopyrighted dataset
\citep{gao2022pile}: \textit{PubMed Central, ArXiv, GitHub, FreeLaw,
StackExchange, USPTO Backgrounds, Gutenberg (PG-19), English Wikipedia,
Hacker News, and DM Mathematics}.

Using seed 2357, we split each domain into disjoint training and validation
sets through a deterministic document-level hash, assigning 99\% of the
documents to training and 1\% to validation. We tokenize the documents using
the GPT-2 tokenizer and append an end-of-sequence token after every document.
We then pack the tokenized documents across document boundaries into blocks
of 1,028 tokens, yielding 1,027 next-token prediction targets per block.

For each domain, we select $2^{20}=1{,}048{,}576$ training blocks and 512
validation blocks. The selected blocks are fixed across all experimental
seeds.

\paragraph{Streams and joint-data reference}
Each effective minibatch of 512 blocks constitutes one training unit and
contains 525,824 next-token prediction targets. For each domain, we group the
selected blocks into 2,048 minibatches. A complete stream therefore contains
20,480 optimizer updates and approximately 10.77 billion target tokens across
the ten domains. Each non-\iid minibatch contains blocks from a single domain,
and each sequence consists of $L$ consecutive minibatches from that domain.
We randomize the order of the resulting sequences for each experimental seed
and avoid adjacent sequences from the same domain whenever possible. Every
selected minibatch occurs exactly once in each stream.

As a joint-data reference, we globally shuffle the same 10,485,760 training
blocks and form mixed-domain minibatches with the same effective minibatch
size. The reference uses the same random initialization, selected data,
optimizer settings, and number of updates as the non-\iid runs, but trains
continuously without parameter or optimizer-state integration. It differs
from the $L=1$, $\lambda=1$ condition, in which every minibatch remains
domain-specific and an integration step is performed after every update.

\paragraph{Local training and integration}
We optimize the next-token language-modelling objective in
\cref{eq:app-lm}. We select the learning rate through a grid search on the
joint-data reference and use the selected value of $10^{-3}$ for all non-\iid
runs. We use AdamW with a weight decay of $0.1$,
$(\beta_1,\beta_2)=(0.9,0.95)$, gradient-norm clipping at $1$, and a constant
learning-rate schedule. We consider
\begin{align}
    L
    & \in
    \{1,64,256,1024,2048\},
    \label{eq:app-scratch-l}
    \\
    \lambda
    & \in
    \{0.2,0.4,0.6,0.8,1\}.
    \label{eq:app-scratch-lambda}
\end{align}
After every sequence, we integrate both the model parameters and AdamW state
using the same static coefficient throughout a run.

\paragraph{Evaluation and seeds}
At each evaluation checkpoint, we evaluate one minibatch of 512 validation
blocks from each domain and compute the macro-average BPT gain across the ten
domains. For $L\geq64$, we evaluate before local adaptation, after local
adaptation but before integration, and after integration at every sequence.
For $L=1$ and for the joint-data reference, we evaluate every 64
optimizer updates.

Our primary trajectory statistic is the post-integration BPT gain, averaged
first across domains, then across evaluation checkpoints within each run, and
finally across seeds. We
run both the non-\iid experiments and the joint-data reference with seeds $0$,
$1$, and $2$.

\subsection{Book-level continual pretraining}
\label[appendixsection]{app:pg19-books}
\input{figures/results/pg19/books/summary_rows}
\input{figures/results/pg19/books/category_rows}

\paragraph{Model}
We perform full-parameter continual pretraining of SmolLM-360M~\citep{allal2024smollm}.
Each run starts from the same pretrained checkpoint. Forward and backward
passes use BF16 mixed precision, with parameters and optimizer moments in FP32.

\paragraph{Books and adaptation sequences}
For each paired data and training seed in $\{0,1,2\}$, we randomly permute
the PG-19 training split~\citep{rae2020compressive} and select the first
200 books with at least two usable token blocks.
All 24 configurations within a seed use the same books, order, and held-out
partitions. Across the three seeds, the 600 book entries represent 597 distinct
Gutenberg IDs. Books define the adaptation boundaries, while their order
is seed-dependent rather than chronological.

We tokenize with the SmolLM-360M tokenizer without chat formatting or added
special tokens and divide each book into non-overlapping blocks of at most
2,048 tokens, retaining partial blocks with at least two tokens.
We randomly hold out $\lceil0.1N_b\rceil$ of a book's $N_b$ usable blocks
and process the remainder once in text order, without replay or cross-book packing.
With four blocks per optimizer step,
the number of updates before integration is $\lceil N_b^{\mathrm{train}}/4\rceil$.

\paragraph{Local training and integration}
We minimize next-token cross-entropy using AdamW with
$(\beta_1,\beta_2)=(0.9,0.999)$, $\epsilon=10^{-8}$, no weight decay,
and gradient-norm clipping at $1$.
The global batch contains four blocks, one per GPU, without gradient accumulation;
padding and dummy examples are excluded from the token-normalized loss.
We use constant learning rates $\eta\in\{1,2,3,4\}\times10^{-5}$ and
$\lambda\in\{0.4,0.6,0.8,1\}$, integrating parameters and optimizer state
after each book according to
\cref{eq:linear-integration,eq:app-first-moment,eq:app-step-count,eq:app-second-moment}.
For each partial-integration setting, we include full integration at learning
rate $\lambda\eta$. Reusing shared baselines gives 24 configurations per seed
and 72 runs in total.

\paragraph{Evaluation and seeds}
We evaluate the continuing model's general capabilities after each integration
on two fixed sampled blocks from each of 50 separate validation books,
identical across configurations and seeds.
The primary score is BPT reduction relative to initialization, averaged equally
over validation books and the 200 integration boundaries.
We report the mean and SEM across the three seeds.

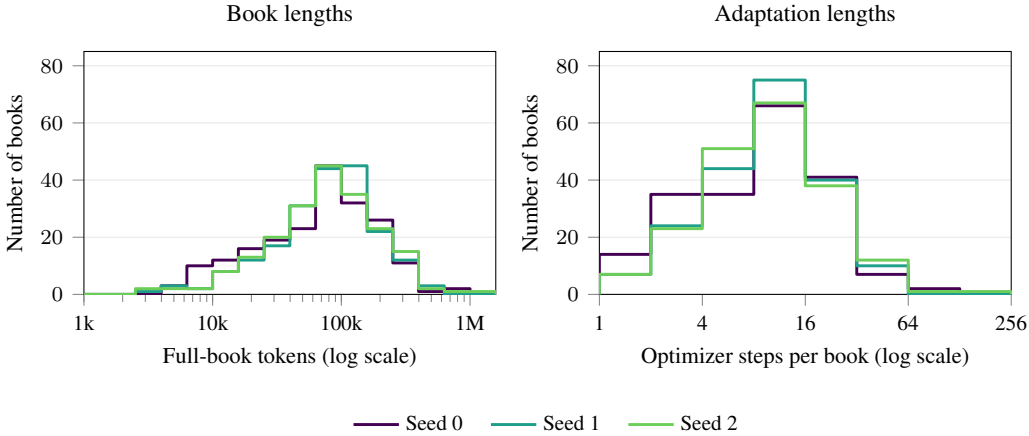
\begin{figure}[h]
    \centering
    \input{figures/results/pg19_book_histograms}
    \caption{PG-19 book lengths and adaptation lengths for 200 books per seed.
    Outline histograms use common logarithmically spaced bins within each
    panel. Full-book token counts include the held-out portion; optimizer
    steps count only training updates before integration.}
    \label{fig:pg19-book-histograms}
\end{figure}

\paragraph{Length and exposure}
Full-book lengths range from 3,011 to 1,435,195 tokens.
Median lengths are 80.9k--83.2k tokens across seeds. The resulting
adaptation sequences contain 1--158 optimizer steps, with medians of 9--9.5
steps. \Cref{fig:pg19-book-histograms,tab:pg19-book-summary} report the
distributions and per-run exposure; totals are not multiplied by the number
of learning-rate and integration configurations.

\begin{table}[h]
    \centering
    \small
    \setlength{\tabcolsep}{4pt}
    \caption{Book lengths and training exposure per run. k and M denote
    thousands and millions of tokens. Training targets exclude held-out
    blocks, the first token of each block, and padding. P90 uses nearest rank.}
    \label{tab:pg19-book-summary}
    \begin{tabular*}{0.94\linewidth}{@{\extracolsep{\fill}}rrrrrrr@{}}
        \toprule
        Seed & \multicolumn{2}{c}{Full-book tokens (k)} &
        Training targets (M) &
        \multicolumn{2}{c}{Steps per book} & Total steps \\
        \cmidrule(lr){2-3}\cmidrule(lr){5-6}
        & Median & P90 & & Median & Range & \\
        \midrule
        \pgNineteenSummaryRows
        \bottomrule
    \end{tabular*}
\end{table}

\paragraph{Catalog categories}
All 597 distinct IDs match the supplied Project Gutenberg catalog
snapshot.\footnote{\url{https://www.gutenberg.org/ebooks/offline_catalogs.html}.}
\Cref{tab:pg19-book-categories} reports its eight most frequent category
labels across the 600 seed/book entries. Labels overlap and are descriptive
catalog annotations, not mutually exclusive genres or training-task labels.
The catalog language field is English-only for 597 entries; two carry
English plus another language and one is labeled Italian. These are metadata
annotations rather than language detection on the training text.

\begin{table}[h]
    \centering
    \small
    \caption{Most frequent Gutenberg categories. Entries are percentages of
    the 200 books in each seed; overlapping labels need not sum to 100\%.}
    \label{tab:pg19-book-categories}
    \begin{tabular}{@{}lrrr@{}}
        \toprule
        Category & Seed 0 & Seed 1 & Seed 2 \\
        \midrule
        \pgNineteenCategoryRows
        \bottomrule
    \end{tabular}
\end{table}

\FloatBarrier
\subsection{Controlled long-SFT interpolation}
\label[appendixsection]{app:long-sft}

\paragraph{Data and training}
We shuffle MetaMathQA with seed 42, remove examples with empty queries or
responses, and select the first 100,000 valid examples. A deterministic 99/1
split provides 99,000 training examples and 1,000 held-out examples; the
held-out examples are not used for training or checkpoint selection. Each
example is formatted with the native SmolLM2 chat template and the system
message ``You are a helpful AI assistant that solves math problems step by
step.'' The MetaMathQA response is retained verbatim apart from surrounding
whitespace, and the loss is applied only to assistant-response tokens.

We fine-tune all parameters for four epochs using fused AdamW with a peak
learning rate of $10^{-5}$, cosine decay, a warmup ratio of $0.05$, weight
decay of $0.01$, and gradient clipping at a global norm of $1.0$. Training
uses a maximum sequence length of 2,048 without packing, BF16 precision, and
an effective global batch size of 16. Training is distributed across four
NVIDIA GH200 GPUs using data parallelism.

\paragraph{Checkpoints and interpolation}
We analyze checkpoints after 50, 100, 250, 500, 1,000, 1,500, 2,000, 3,000,
5,000, 7,500, 10,000, 12,500, 15,000, 17,500, and 20,000 optimizer
steps. For checkpoint parameters $\theta_s$ and initial parameters $\theta_0$,
we construct
$\theta_{s,\lambda}=(1-\lambda)\theta_0+\lambda\theta_s$ for
$\lambda\in\{0,.2,.4,.6,.8,1\}$. For each checkpoint, we evaluate four interior interpolants alongside the initial model and the SFT endpoint. Floating-point tensors are
interpolated elementwise in BF16, and all models share the same architecture,
tokenizer, vocabulary, and chat template.

\paragraph{Evaluation}
We evaluate the complete GSM8K test set using zero-shot greedy generation with
one response per problem. Evaluation uses the training-time system message and
native chat template, a maximum generation length of 1,024 tokens, and a
2,048-token context. Predictions are scored with \texttt{math\_verify}, which
parses and symbolically compares the generated answer with the numerical
reference following the dataset's \texttt{\#\#\#\#} delimiter. We report mean
verification accuracy over all test examples.

\subsection{Qwen2.5--Coder diagnostic}
\label[appendixsection]{app:coder-setup}

In this experiment, we test whether linear parameter connectivity is
preserved after a much larger adaptation. This is a diagnostic between
released checkpoints, not a continual-training experiment.

\paragraph{Models and parameter path}
We consider the 1.5B, 3B, and 7B parameter sizes of the Qwen2.5 family. For
each scale, we use the released Qwen2.5 base checkpoint as the generalist model
and the Qwen2.5-Coder checkpoint as the specialist
\citep{yang2024qwen25,hui2024qwen25coder}. The Coder lineage was pretrained on
5.2T tokens at file level and approximately 300B additional tokens at
repository level. We construct the direct linear path
\begin{align}
    \theta_{\lambda}
    & =
    (1-\lambda)\theta_{\mathrm{base}}
    +
    \lambda\theta_{\mathrm{coder}},
    \qquad
    \lambda\in\{0,0.2,0.4,0.6,0.8,1\},
    \label{eq:app-qwen-coder-path}
\end{align}
where $\lambda=0$ corresponds to the base checkpoint and $\lambda=1$ to the
Coder checkpoint. This experiment requires no additional training or
optimizer-state integration.

\paragraph{Evaluation}
We evaluate each checkpoint using vLLM. As the targeted benchmark, we use all
500 problems from the MBPP test set \citep{austin2021program} with the
three-shot harness prompt and report pass@1 based on executable tests.

To measure retention, we evaluate the complete MMLU
\citep{hendrycks2021mmlu}, ARC-Challenge \citep{clark2018arc}, HellaSwag
\citep{zellers2019hellaswag}, and WinoGrande
\citep{sakaguchi2020winogrande} evaluation sets. We define the non-targeted
score as the unweighted mean of MMLU accuracy, normalized ARC-Challenge
accuracy, normalized HellaSwag accuracy, and WinoGrande accuracy. We use the
same evaluation configuration at every point along the interpolation path.

\subsection{Compute resources and recorded runtime}
\label[appendixsection]{app:compute}

All experiments, including PG-19, used shared HPC nodes with four NVIDIA GH200 GPUs
(96\,GB HBM3 each), 288 Arm CPU cores, and 512\,GB nominal host
memory.\footnote{\url{https://www.cscs.ch/computers/alps}.}

\begin{table}[htbp]
    \centering
    \small
    \setlength{\tabcolsep}{5pt}
    \caption{Recorded compute for the six training settings.
    Hours/run gives the runtime range; GPU-hours sums runtime multiplied by
    allocated GPUs. Timings include evaluation.}
    \label{app:tab:compute}
    \begin{tabular}{@{}lrrrr@{}}
        \toprule
        Setting & GPUs/run & Timed runs & Hours/run & GPU-hours \\
        \midrule
        Continual pretraining & 4 & 72 & 0.11--0.55 & 79.85 \\
        Persona SDPO & 2 & 84 & 0.91--2.94 & 292.57 \\
        Reasoning SFT & 4 & 125 & 0.29--3.58 & 540.36 \\
        Reasoning GRPO & 4 & 39 & 4.53--8.51 & 896.76 \\
        Pretraining from scratch & 4 & 78 & 1.94--2.59 & 663.71 \\
        PG-19 continual pretraining & 4 & 72 & 0.49--1.09 & 202.87 \\
        \midrule
        Recorded subtotal & & 470 & & 2,676.13 \\
        \bottomrule
    \end{tabular}
\end{table}

\FloatBarrier

\section{Extended results}
\label[appendixsection]{app:extended-results}

This section provides additional integration trade-offs, long-SFT interpolation
results, and the complete pretraining-scale interpolation diagnostic.

\subsection{Additional integration trade-offs}
\label[appendixsection]{app:secondary-results}

\paragraph{Immediate effect of task-vector scaling}
To isolate the immediate effect of applying $\lambda$, we compare the
continuing model $\theta_{t+1}$ with the endpoint $\theta_t^{(L)}$. For
each distribution, define the score change caused by scaling as
\begin{align}
    \delta^{\mathrm{scale}}_{t,k}
    =
    \operatorname{score}(\theta_{t+1};k)
    -
    \operatorname{score}(\theta_t^{(L)};k).
    \label{eq:integration-effects}
\end{align}
The targeted-distribution change is $\delta^{\mathrm{scale}}_{t,k_t}$, while
the remaining $\delta^{\mathrm{scale}}_{t,k}$ measure score changes on
non-targeted distributions. The corresponding change in average performance
decomposes as
\begin{align}
    \operatorname{ScaleEffect}_t
    = \frac{1}{K} \sum_{k=1}^K \delta^{\mathrm{scale}}_{t,k}
    =
    \underbrace{
        \frac{1}{K}
        \sum_{k\ne k_t}\delta^{\mathrm{scale}}_{t,k}
    }_{\text{non-targeted contribution}}
    +
    \underbrace{
        \frac{1}{K}
        \delta^{\mathrm{scale}}_{t,k_t}
        \vphantom{\sum_{k\ne k_t}\delta^{\mathrm{scale}}_{t,k}}
    }_{\text{targeted contribution}}.
    \label{eq:integration-decomposition}
\end{align}
Here, $\delta^{\mathrm{scale}}_{t,k_t}<0$ means that scaling removes part of the
targeted-distribution gain, whereas positive non-targeted changes indicate
recovered performance on the other distributions. In our plots, we call the
non-targeted contribution \emph{non-targeted recovery} and the negative
targeted contribution \emph{targeted loss}. Thus,
$\operatorname{ScaleEffect}_t$ equals non-targeted recovery minus targeted
loss. Positive values indicate recovered non-targeted performance and lost
targeted-distribution performance, respectively.

\begin{figure}[h]
    \centering
    \input{figures/results/observation_merge_tradeoff_reasoning}
    \caption{Immediate effect of task-vector scaling for SFT (left) and GRPO
    (right). The dotted diagonal marks zero change in average performance
    relative to full integration.}
    \label{fig:integration-tradeoff}
\end{figure}
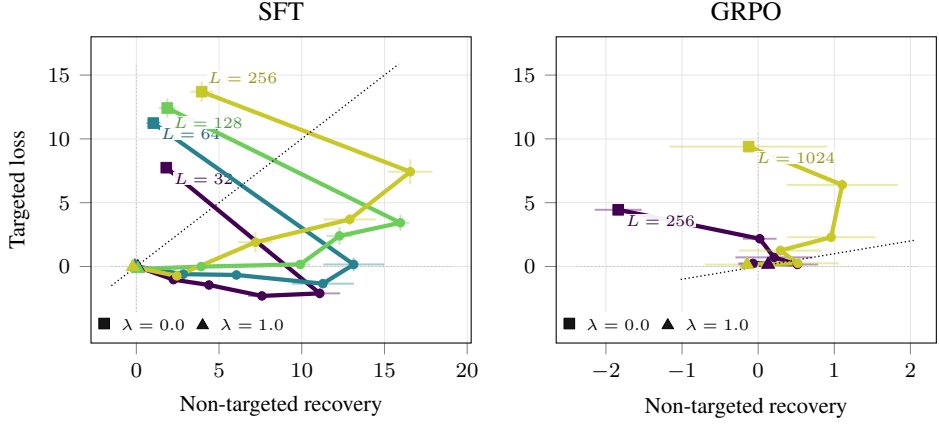

\Cref{fig:integration-tradeoff} plots non-targeted recovery against targeted loss, as defined by \cref{eq:integration-decomposition}. Points below the diagonal improve average performance over full integration; above it, targeted loss exceeds non-targeted recovery.

For SFT, scaling recovers substantial non-targeted performance, placing several intermediate coefficients below the diagonal. GRPO recovers little, and most points lie above it. This contrast may reflect off-policy SFT versus on-policy GRPO: on-policy adaptation appears to disrupt non-targeted performance less, leaving less to recover. This interpretation is consistent with \emph{RL's Razor}, which finds that on-policy RL forgets less than SFT by remaining closer in KL divergence to the base policy \citep{shenfeld2026rl}. These averages capture only the immediate effect, not downstream effects on later checkpoints.

\begin{figure*}[h]
    \centering
    \resizebox{0.95\textwidth}{!}{%
        \input{figures/results/observation_merge_tradeoff}}
    \caption{Task-count-normalized integration trade-off for continual
    pretraining and persona SDPO. The horizontal axis is the combined
    non-targeted recovery contribution and the vertical axis is the
    targeted-loss contribution. The dotted diagonal is the zero-change
    boundary. Circles mark partial integration ($0<\lambda<1$), squares mark
    $\lambda=0$ where evaluated, and triangles mark $\lambda=1$.}
    \label{fig:app-integration-tradeoff}
\end{figure*}
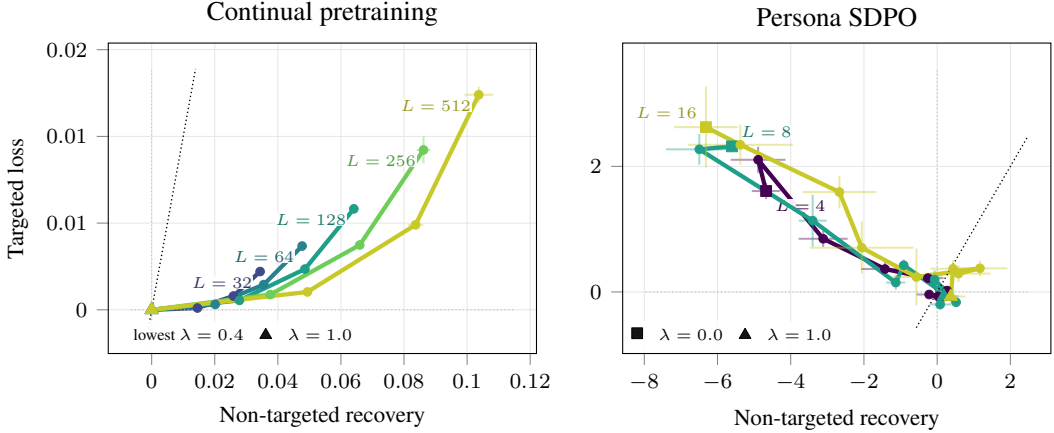

\Cref{fig:app-integration-tradeoff} extends this analysis to continual pretraining and persona SDPO. In
continual pretraining, partial-integration points lie below the zero-change
boundary for
every evaluated $L$, indicating that non-targeted recovery outweighs the loss
on the targeted distribution. Persona SDPO is less monotonic: some intermediate
coefficients improve targeted and non-targeted style scores simultaneously,
whereas very small coefficients can reduce non-targeted performance. This
pattern is consistent with the on-policy feedback updates improving multiple
measured style traits rather than inducing a strict
adaptation--retention trade-off.

\begin{figure}[h]
    \centering
    \input{figures/results/pretraining_scratch_scaling}
    \caption{Immediate effect of task-vector scaling during pretraining from
    random initialization. Non-targeted recovery and targeted loss are defined
    in~\cref{eq:integration-decomposition}; their difference gives the change in
    average performance. The dotted diagonal marks zero change. Colors denote
    sequence lengths as in~\cref{fig:scratch-outcomes}; circles mark partial
    integration and triangles mark $\lambda=1$. Error bars show SEM across seeds.}
    \label{fig:app-scratch-scaling}
\end{figure}
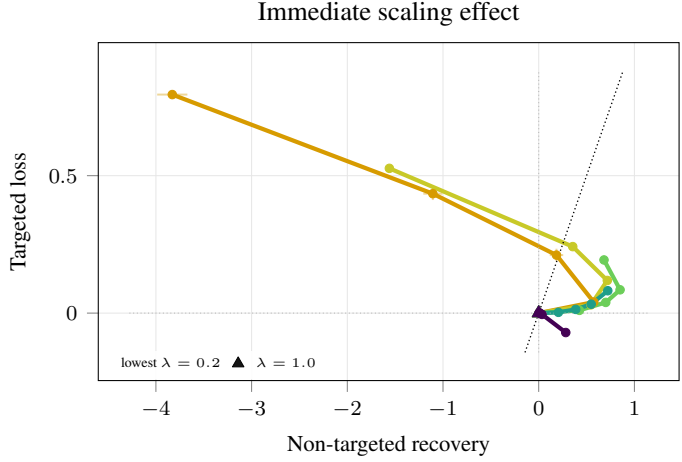

\paragraph{Pretraining from random initialization}
\Cref{fig:app-scratch-scaling} shows the immediate effect of scaling during
pretraining from random initialization. Very small coefficients reduce even
non-targeted performance for $L\in\{1024,2048\}$, yielding negative
non-targeted recovery. This pattern is consistent with task vectors containing
generally useful representation learning in addition to distribution-specific
adaptation; retaining too little of the vector can therefore impede learning
across distributions.

\FloatBarrier
\subsection{Long-SFT interpolation}
\label[appendixsection]{app:long-sft-results}

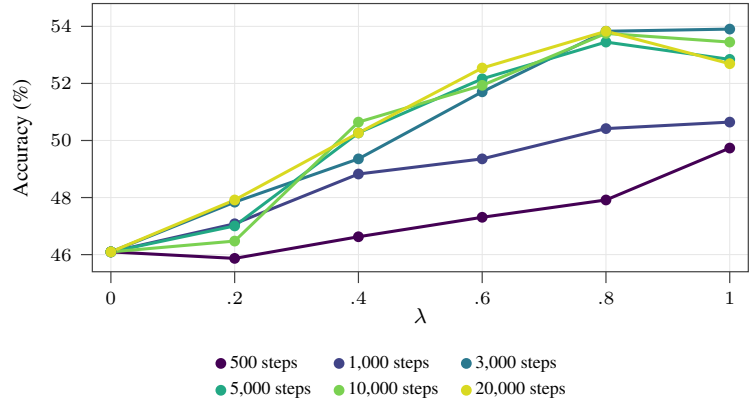
\begin{figure}[h]
    \centering
    \begin{minipage}{0.68\textwidth}
        \centering
        \makebox[\linewidth][c]{\input{figures/results/long_sft_gsm8k}}
    \end{minipage}
    \caption{GSM8K accuracy along interpolations between the initial model and
    checkpoints obtained after different numbers of MetaMathQA SFT steps.}
    \label{fig:long-sft-connectivity}
\end{figure}

Task-vector scaling relies on intermediate parameters remaining useful as
local adaptation progresses. We examine whether longer adaptation trajectories
introduce performance barriers between the starting model and the adapted endpoint.
\Cref{fig:long-sft-connectivity} reports GSM8K accuracy along interpolations
between the initial SmolLM2-1.7B-Instruct model and checkpoints from a controlled
MetaMathQA SFT trajectory, described in~\cref{app:long-sft}.
Across the evaluated checkpoints,
accuracy varies smoothly with $\lambda$, without a pronounced interior
performance barrier, so intermediate task-vector scaling remains effective
across the tested adaptation lengths.

\FloatBarrier

\subsection{Qwen2.5--Coder diagnostic}
\label[appendixsection]{app:coder}

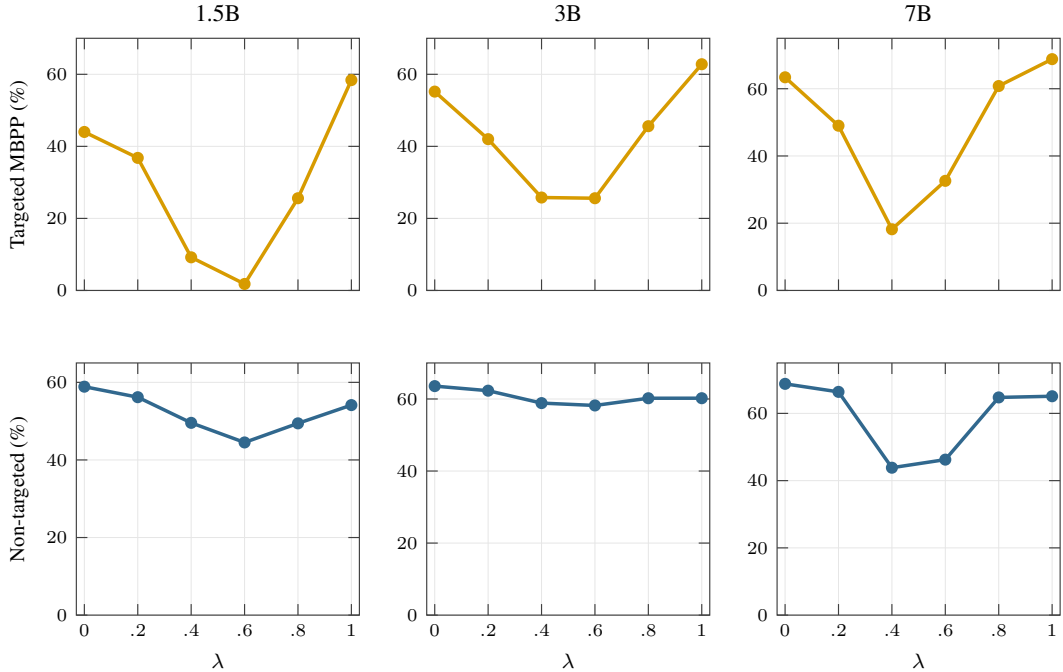
\begin{figure*}[h]
    \centering
    \resizebox{0.95\textwidth}{!}{%
        \input{figures/results/qwen_coder_large_adaptation}}
    \caption{Direct interpolation between released Qwen2.5 base checkpoints
    and their Qwen2.5-Coder descendants. Columns show parameter scale. The top
    row reports targeted MBPP pass@1, while the bottom row reports the
    unweighted mean of full MMLU, normalized ARC-Challenge, normalized
    HellaSwag, and WinoGrande accuracy.}
    \label{fig:qwen-coder-path}
\end{figure*}

To test the limits of direct interpolation beyond the controlled long-SFT
trajectory, we evaluate the released Qwen2.5 base and Qwen2.5-Coder
checkpoints described in \cref{app:coder-setup}. \Cref{fig:qwen-coder-path}
reports targeted and non-targeted performance along this path at the 1.5B,
3B, and 7B parameter scales.

Interior interpolation points perform worse than both endpoints on targeted
and non-targeted scores at all three model scales. For example, at 7B
parameters, MBPP falls from $63.4\%$ at the base checkpoint and $68.8\%$ at
the Coder checkpoint to $18.2\%$ at $\lambda=0.4$; the non-targeted score falls
from $68.79\%$ and $65.10\%$ at the endpoints to $43.84\%$. These results mark
a pretraining-scale boundary to the well-behaved interpolation observed in
\cref{sec:results-connectivity}.

%% file: figures/controlled_streams_figure.tex
\centering
\resizebox{\linewidth}{!}{\input{figures/controlled_streams}}
\caption{Controlled streams. Each box contains $L$ batches from one
distribution; \iid batches draw from the complete mixture. Larger $L$ increases temporal clustering.}
\label{fig:controlled-streams}

%% file: sections/asset_licenses.tex
\begin{table}[tbp]
    \centering
    \small
    \setlength{\tabcolsep}{4pt}
    \renewcommand{\arraystretch}{1.12}
    \caption{Existing assets and their published licenses or terms. Asset
    names link to the source documentation; citations credit the original work.}
    \label{tab:asset-licenses}
    \begin{tabular}{@{}>{\raggedright\arraybackslash}p{.64\linewidth}>{\raggedright\arraybackslash}p{\dimexpr.36\linewidth-8pt\relax}@{}}
        \toprule
        \textbf{Asset / checkpoint} & \textbf{License / terms} \\
        \midrule
        \multicolumn{2}{@{}l}{\textit{Models and services}} \\
        \href{https://github.com/openai/gpt-2/blob/master/LICENSE}{GPT-2 (124M)} \citep{radford2019language}
            & Modified MIT \\
        \href{https://huggingface.co/Qwen/Qwen3-4B}{Qwen3-4B}, \href{https://huggingface.co/Qwen/Qwen3-8B}{Qwen3-8B} \citep{yang2025qwen3}
            & Apache 2.0 \\
        \href{https://huggingface.co/Qwen/Qwen2.5-1.5B}{Qwen2.5-1.5B}, \href{https://huggingface.co/Qwen/Qwen2.5-7B}{Qwen2.5-7B} \citep{yang2024qwen25}
            & Apache 2.0 \\
        \href{https://huggingface.co/Qwen/Qwen2.5-3B/blob/main/LICENSE}{Qwen2.5-3B}, \href{https://huggingface.co/Qwen/Qwen2.5-3B-Instruct}{Qwen2.5-3B-Instruct} \citep{yang2024qwen25}
            & Qwen Research License \\
        \href{https://huggingface.co/Qwen/Qwen2.5-Coder-1.5B}{Qwen2.5-Coder-1.5B}, \href{https://huggingface.co/Qwen/Qwen2.5-Coder-7B}{-7B} \citep{hui2024qwen25coder}
            & Apache 2.0 \\
        \href{https://huggingface.co/Qwen/Qwen2.5-Coder-3B}{Qwen2.5-Coder-3B} \citep{hui2024qwen25coder}
            & Qwen Research License \\
        \href{https://huggingface.co/HuggingFaceTB/SmolLM-360M}{SmolLM-360M} \citep{allal2024smollm}
            & Apache 2.0 \\
        \href{https://huggingface.co/HuggingFaceTB/SmolLM2-1.7B-Instruct}{SmolLM2-1.7B-Instruct} \citep{benallal2025smollm2}
            & Apache 2.0 \\
        \href{https://www.anthropic.com/legal/commercial-terms}{Claude Sonnet 4.6} \citep{anthropic2026sonnet46}
            & Proprietary service terms$^{\dagger}$ \\
        \midrule
        \multicolumn{2}{@{}l}{\textit{Datasets and task generators}} \\
        \href{https://huggingface.co/datasets/monology/pile-uncopyrighted}{Pile Uncopyrighted (ten domains)} \citep{gao2022pile}
            & MIT (collection)$^{\ddagger}$ \\
        \href{https://github.com/google-deepmind/pg19}{PG-19} \citep{rae2020compressive}
            & Apache 2.0 (collection)$^{\ddagger}$ \\
        \href{https://huggingface.co/datasets/nvidia/HelpSteer}{HelpSteer} \citep{wang2023helpsteer}
            & CC BY 4.0 \\
        \href{https://github.com/open-thought/reasoning-gym}{Reasoning Gym (four tasks)} \citep{stojanovski2025reasoninggym}
            & Apache 2.0 \\
        \href{https://huggingface.co/datasets/meta-math/MetaMathQA}{MetaMathQA} \citep{yu2023metamath}
            & MIT (dataset card)$^{\ddagger}$ \\
        \href{https://huggingface.co/datasets/openai/gsm8k}{GSM8K} \citep{cobbe2021training}
            & MIT \\
        \href{https://huggingface.co/datasets/google-research-datasets/mbpp}{MBPP} \citep{austin2021program}
            & CC BY 4.0 \\
        \href{https://huggingface.co/datasets/cais/mmlu}{MMLU} \citep{hendrycks2021mmlu}
            & MIT \\
        \href{https://huggingface.co/datasets/allenai/ai2_arc}{ARC-Challenge} \citep{clark2018arc}
            & CC BY-SA 4.0 \\
        \href{https://huggingface.co/datasets/Rowan/hellaswag}{HellaSwag} \citep{zellers2019hellaswag}
            & MIT \\
        \href{https://github.com/allenai/winogrande/blob/master/LICENSE}{WinoGrande} \citep{sakaguchi2020winogrande}
            & Apache 2.0 \\
        \midrule
        \multicolumn{2}{@{}l}{\textit{Principal software}} \\
        \href{https://github.com/huggingface/transformers/blob/main/LICENSE}{Transformers} \citep{wolf2020transformers},
        \href{https://github.com/huggingface/trl/blob/main/LICENSE}{TRL} \citep{vonwerra2020trl},
        \href{https://github.com/vllm-project/vllm/blob/main/LICENSE}{vLLM} \citep{kwon2023vllm}
            & Apache 2.0 (each) \\
        \bottomrule
    \end{tabular}
    \par\smallskip
    \begin{minipage}{\linewidth}
        \footnotesize
        The Qwen Research License permits non-commercial research and evaluation;
        commercial use requires a separate license.
        $^{\dagger}$Claude is accessed as a service; the applicable service
        agreement governs its use and generated outputs.
        $^{\ddagger}$A collection-level license does not replace the terms of
        underlying sources. The \href{https://arxiv.org/pdf/2201.07311}{Pile datasheet}
        declares MIT for the original collection; individual documents retain
        their underlying rights. MetaMathQA derives examples from GSM8K and MATH.
    \end{minipage}
\end{table}

%% file: figures/results/pg19/books/summary_rows.tex
\newcommand{\pgNineteenSummaryRows}{%
0 & 80.9 & 221.4 & 18.49 & 9 & 1--91 & 2,353 \\
1 & 81.1 & 219.0 & 18.98 & 9 & 1--57 & 2,410 \\
2 & 83.2 & 242.1 & 20.41 & 9.5 & 1--158 & 2,585 \\
}

%% file: figures/results/pg19/books/category_rows.tex
\newcommand{\pgNineteenCategoryRows}{%
Novels & 32.5 & 30.0 & 32.0 \\
British Literature & 20.0 & 18.0 & 19.5 \\
American Literature & 10.0 & 10.0 & 14.0 \\
History - Modern (1750+) & 10.5 & 10.5 & 13.0 \\
Adventure & 11.5 & 12.0 & 9.5 \\
Biographies & 9.0 & 8.5 & 12.5 \\
Children \& Young Adult Reading & 10.0 & 9.0 & 10.0 \\
Historical Novels & 10.5 & 6.5 & 10.5 \\
}

%% file: figures/results/pg19_book_histograms.tex
\begingroup
\definecolor{bookSeed0}{HTML}{440154}
\definecolor{bookSeed1}{HTML}{1F9E89}
\definecolor{bookSeed2}{HTML}{6CCE59}
\begin{tikzpicture}
\begin{groupplot}[
    group style={group size=2 by 1,horizontal sep=3.9em},
    width=0.39\textwidth,height=0.23\textwidth,
    scale only axis,
    xmode=log,
    ymin=0,ymax=85,ytick={0,20,40,60,80},
    ylabel={Number of books},
    title style={font=\small},
    label style={font=\fontsize{8.5}{9.5}\selectfont},
    tick label style={font=\fontsize{8}{9}\selectfont},
    tick align=outside,tick pos=left,
    axis line style={line width=0.4pt},
    major grid style={gray!20,line width=0.3pt},
    ymajorgrids=true,
    minor x tick num=0,
    every axis plot/.append style={line join=round},
    legend style={draw=none,fill=none,font=\fontsize{8}{9}\selectfont,
        legend columns=3,/tikz/every even column/.append style={column sep=0.8em}},
    legend cell align={left},
]
\nextgroupplot[
    title={Book lengths},
    xlabel={Full-book tokens (log scale)},
    xmin=1000,xmax=1584893.192461114,
    xtick={1000,10000,100000,1000000},
    xticklabels={1k,10k,100k,1M},
    legend to name=pgNineteenBookSeeds,
]
\pgfplotsinvokeforeach{0,1,2}{%
    \addplot[const plot,color=bookSeed#1,line width=1.1pt]
        table[col sep=comma,x=edge,y=seed#1]
        {figures/results/pg19/books/token_histogram.csv};
    \addlegendentry{Seed #1}
}
\nextgroupplot[
    title={Adaptation lengths},
    xlabel={Optimizer steps per book (log scale)},
    xmin=1,xmax=256,
    xtick={1,4,16,64,256},
    xticklabels={1,4,16,64,256},
]
\pgfplotsinvokeforeach{0,1,2}{%
    \addplot[const plot,color=bookSeed#1,line width=1.1pt]
        table[col sep=comma,x=edge,y=seed#1]
        {figures/results/pg19/books/step_histogram.csv};
}
\end{groupplot}
\path (group c1r1.south west) -- (group c2r1.south east)
    node[midway,below=3.8em] {\ref{pgNineteenBookSeeds}};
\end{tikzpicture}%
\endgroup%

%% file: figures/results/observation_merge_tradeoff_reasoning.tex
\definecolor{runL1}{HTML}{440154}
\definecolor{runL4}{HTML}{482878}
\definecolor{runL8}{HTML}{3E4989}
\definecolor{runL16}{HTML}{31688E}
\definecolor{runL32}{HTML}{26828E}
\definecolor{runL64}{HTML}{1F9E89}
\definecolor{runL128}{HTML}{35B779}
\definecolor{runL256}{HTML}{6CCE59}
\definecolor{runL512}{HTML}{A0DA39}
\definecolor{runL1024}{HTML}{C8C928}
\definecolor{runL2048}{HTML}{D79B00}
\definecolor{runscratchdeltaL64}{HTML}{1F9E89}
\definecolor{runscratchdeltaL256}{HTML}{6CCE59}
\definecolor{runscratchdeltaL1024}{HTML}{C8C928}
\definecolor{runscratchdeltaL2048}{HTML}{D79B00}
\definecolor{runpileL1}{HTML}{440154}
\definecolor{runpileL32}{HTML}{3E4989}
\definecolor{runpileL64}{HTML}{26828E}
\definecolor{runpileL128}{HTML}{1F9E89}
\definecolor{runpileL256}{HTML}{6CCE59}
\definecolor{runpileL512}{HTML}{C8C928}
\definecolor{runpersonaL4}{HTML}{440154}
\definecolor{runpersonaL8}{HTML}{1F9E89}
\definecolor{runpersonaL16}{HTML}{C8C928}
\definecolor{runsftL32}{HTML}{440154}
\definecolor{runsftL64}{HTML}{26828E}
\definecolor{runsftL128}{HTML}{6CCE59}
\definecolor{runsftL256}{HTML}{C8C928}
\definecolor{runrlvrL256}{HTML}{440154}
\definecolor{runrlvrL1024}{HTML}{C8C928}
\begin{tikzpicture}[trim axis group left,trim axis group right]
\begin{groupplot}[
  group style={group size=2 by 1, horizontal sep=3.2em},
  width=0.36\textwidth,
  height=0.29\textwidth,
  scale only axis,
  title style={font=\fontsize{10}{11}\selectfont,align=center,yshift=-0.7ex},
  label style={font=\fontsize{8.5}{9.5}\selectfont},
  tick label style={font=\fontsize{8}{9}\selectfont,/pgf/number format/fixed,/pgf/number format/precision=2},
  tick align=outside,
  tick pos=left,
  axis line style={line width=0.4pt},
  major grid style={gray!20,line width=0.3pt},
  grid=major,
  enlarge x limits=0.06,
  enlarge y limits=0.10,
  ymin=-4,
  ymax=16,
  ytick={0,5,10,15},
  scaled ticks=false,
  every axis plot/.append style={line join=round},
]
\nextgroupplot[title={SFT},xlabel={Non-targeted recovery},ylabel={Targeted loss},xlabel style={align=center},xmin=-1.507911,xmax=19.00969,legend style={at={(0.015,0.005)},anchor=south west,draw=none,fill=white,fill opacity=0.92,text opacity=1,font=\fontsize{6}{7}\selectfont,inner sep=0.6pt,column sep=0.25em,row sep=-0.4pt},legend columns=2,legend cell align={left},legend image post style={yshift=0.25ex}]
\addlegendimage{only marks,black,mark=square*,mark size=2.0pt,mark options={line width=0pt}}
\addlegendentry{$\lambda=0.0$}
\addlegendimage{only marks,black,mark=triangle*,mark size=2.5pt,mark options={line width=0pt,yshift=-0.15ex}}
\addlegendentry{$\lambda=1.0$}
\addplot[color=runsftL32,line width=1.6pt,mark=none,error bars/error bar style={draw=runsftL32,opacity=0.4,draw opacity=0.4,fill opacity=0.4,line width=0.75pt},error bars/error mark=none,error bars/.cd,x dir=both,x explicit,y dir=both,y explicit] coordinates {(1.810938,7.742188) +- (0.2872133,0.2049944) (11.08906,-2.109375) +- (1.238861,0.4016263) (7.59375,-2.307812) +- (0.83589,0.1953844) (4.395313,-1.453125) +- (0.4793502,0.1385921) (2.246875,-1.032812) +- (0.3380981,0.06676831) (0.0921875,-0.040625) +- (0.07747385,0.04799445)};
\addplot[only marks,color=runsftL32,mark=square*,mark size=2.0pt,mark options={line width=0pt}] coordinates {(1.810938,7.742188)};
\addplot[only marks,color=runsftL32,mark=*,mark size=1.6pt,mark options={line width=0pt}] coordinates {(11.08906,-2.109375) (7.59375,-2.307812) (4.395313,-1.453125) (2.246875,-1.032812)};
\addplot[only marks,color=runsftL32,mark=triangle*,mark size=2.8pt,mark options={line width=0pt}] coordinates {(0.0921875,-0.040625)};
\node[font=\fontsize{6}{6.5}\selectfont,text=runsftL32!75!black,anchor=north west,xshift=2.5pt,yshift=-1.5pt,fill=white,fill opacity=0.94,text opacity=1,inner sep=0.25pt] at (axis cs:1.810938,7.742188) {$L=32$};
\addplot[color=runsftL64,line width=1.6pt,mark=none,error bars/error bar style={draw=runsftL64,opacity=0.4,draw opacity=0.4,fill opacity=0.4,line width=0.75pt},error bars/error mark=none,error bars/.cd,x dir=both,x explicit,y dir=both,y explicit] coordinates {(1.025,11.2375) +- (0.4367907,0.1712631) (13.1375,0.159375) +- (1.83876,0.5278244) (11.27812,-1.346875) +- (1.875979,0.307078) (6.05,-0.665625) +- (0.6908901,0.1987878) (2.840625,-0.603125) +- (0.2806156,0.2552955) (-0.090625,0.025) +- (0.1255262,0.07616286)};
\addplot[only marks,color=runsftL64,mark=square*,mark size=2.0pt,mark options={line width=0pt}] coordinates {(1.025,11.2375)};
\addplot[only marks,color=runsftL64,mark=*,mark size=1.6pt,mark options={line width=0pt}] coordinates {(13.1375,0.159375) (11.27812,-1.346875) (6.05,-0.665625) (2.840625,-0.603125)};
\addplot[only marks,color=runsftL64,mark=triangle*,mark size=2.8pt,mark options={line width=0pt}] coordinates {(-0.090625,0.025)};
\node[font=\fontsize{6}{6.5}\selectfont,text=runsftL64!75!black,anchor=north west,xshift=2.5pt,yshift=-1.5pt,fill=white,fill opacity=0.94,text opacity=1,inner sep=0.25pt] at (axis cs:1.025,11.2375) {$L=64$};
\addplot[color=runsftL128,line width=1.6pt,mark=none,error bars/error bar style={draw=runsftL128,opacity=0.4,draw opacity=0.4,fill opacity=0.4,line width=0.75pt},error bars/error mark=none,error bars/.cd,x dir=both,x explicit,y dir=both,y explicit] coordinates {(1.86875,12.41875) +- (0.5061535,0.7365062) (15.9625,3.425) +- (0.5333964,0.6513658) (12.28125,2.39375) +- (0.7928514,0.6974843) (9.90625,0.1625) +- (0.6145215,0.3219508) (3.9125,-0.0125) +- (0.6731021,0.2574287) (0.1,-0.1875) +- (0.2717852,0.1988738)};
\addplot[only marks,color=runsftL128,mark=square*,mark size=2.0pt,mark options={line width=0pt}] coordinates {(1.86875,12.41875)};
\addplot[only marks,color=runsftL128,mark=*,mark size=1.6pt,mark options={line width=0pt}] coordinates {(15.9625,3.425) (12.28125,2.39375) (9.90625,0.1625) (3.9125,-0.0125)};
\addplot[only marks,color=runsftL128,mark=triangle*,mark size=2.8pt,mark options={line width=0pt}] coordinates {(0.1,-0.1875)};
\node[font=\fontsize{6}{6.5}\selectfont,text=runsftL128!75!black,anchor=north west,xshift=2.0pt,yshift=-3.0pt,fill=white,fill opacity=0.94,text opacity=1,inner sep=0.25pt] at (axis cs:1.86875,12.41875) {$L=128$};
\addplot[color=runsftL256,line width=1.6pt,mark=none,error bars/error bar style={draw=runsftL256,opacity=0.4,draw opacity=0.4,fill opacity=0.4,line width=0.75pt},error bars/error mark=none,error bars/.cd,x dir=both,x explicit,y dir=both,y explicit] coordinates {(3.95,13.7) +- (0.6946784,0.8012198) (16.5625,7.425) +- (1.348031,0.9809514) (12.9,3.6875) +- (1.568812,0.4647748) (7.2,1.9) +- (1.050186,0.7902235) (2.45,-0.7375) +- (0.5256692,0.1847718) (-0.2125,-0.0375) +- (0.196254,0.2188393)};
\addplot[only marks,color=runsftL256,mark=square*,mark size=2.0pt,mark options={line width=0pt}] coordinates {(3.95,13.7)};
\addplot[only marks,color=runsftL256,mark=*,mark size=1.6pt,mark options={line width=0pt}] coordinates {(16.5625,7.425) (12.9,3.6875) (7.2,1.9) (2.45,-0.7375)};
\addplot[only marks,color=runsftL256,mark=triangle*,mark size=2.8pt,mark options={line width=0pt}] coordinates {(-0.2125,-0.0375)};
\node[font=\fontsize{6}{6.5}\selectfont,text=runsftL256!75!black,anchor=south west,xshift=2.0pt,yshift=3.0pt,fill=white,fill opacity=0.94,text opacity=1,inner sep=0.25pt] at (axis cs:3.95,13.7) {$L=256$};
\addplot[gray!55,densely dotted,line width=0.4pt] coordinates {(-1.507911,0) (19.00969,0)};
\addplot[gray!55,densely dotted,line width=0.4pt] coordinates {(0,-3.871979) (0,15.8622)};
\addplot[black,densely dotted,line width=0.5pt] coordinates {(-1.507911,-1.507911) (15.8622,15.8622)};
\nextgroupplot[title={GRPO},xlabel={Non-targeted recovery},xlabel style={align=center},xmin=-2.385898,xmax=2.073101,legend style={at={(0.015,0.005)},anchor=south west,draw=none,fill=white,fill opacity=0.92,text opacity=1,font=\fontsize{6}{7}\selectfont,inner sep=0.6pt,column sep=0.25em,row sep=-0.4pt},legend columns=2,legend cell align={left},legend image post style={yshift=0.25ex}]
\addlegendimage{only marks,black,mark=square*,mark size=2.0pt,mark options={line width=0pt}}
\addlegendentry{$\lambda=0.0$}
\addlegendimage{only marks,black,mark=triangle*,mark size=2.5pt,mark options={line width=0pt,yshift=-0.15ex}}
\addlegendentry{$\lambda=1.0$}
\addplot[color=runrlvrL256,line width=1.6pt,mark=none,error bars/error bar style={draw=runrlvrL256,opacity=0.4,draw opacity=0.4,fill opacity=0.4,line width=0.75pt},error bars/error mark=none,error bars/.cd,x dir=both,x explicit,y dir=both,y explicit] coordinates {(-1.838542,4.442708) +- (0.3084811,0.09895833) (0.02083333,2.171875) +- (0.2214001,0.150141) (0.2083333,0.7135417) +- (0.5071376,0.1822917) (0.515625,0.15625) +- (0.2706329,0.1353165) (-0.06770833,0.2395833) +- (0.1929194,0.09389456) (0.1302083,0.1510417) +- (0.388047,0.113513)};
\addplot[only marks,color=runrlvrL256,mark=square*,mark size=2.0pt,mark options={line width=0pt}] coordinates {(-1.838542,4.442708)};
\addplot[only marks,color=runrlvrL256,mark=*,mark size=1.6pt,mark options={line width=0pt}] coordinates {(0.02083333,2.171875) (0.2083333,0.7135417) (0.515625,0.15625) (-0.06770833,0.2395833)};
\addplot[only marks,color=runrlvrL256,mark=triangle*,mark size=2.8pt,mark options={line width=0pt}] coordinates {(0.1302083,0.1510417)};
\node[font=\fontsize{6}{6.5}\selectfont,text=runrlvrL256!75!black,anchor=north west,xshift=2.5pt,yshift=-1.5pt,fill=white,fill opacity=0.94,text opacity=1,inner sep=0.25pt] at (axis cs:-1.838542,4.442708) {$L=256$};
\addplot[color=runrlvrL1024,line width=1.6pt,mark=none,error bars/error bar style={draw=runrlvrL1024,opacity=0.4,draw opacity=0.4,fill opacity=0.4,line width=0.75pt},error bars/error mark=none,error bars/.cd,x dir=both,x explicit,y dir=both,y explicit] coordinates {(-0.125,9.395833) +- (1.035188,0.3254271) (1.104167,6.395833) +- (0.730059,0.4805234) (0.9583333,2.291667) +- (0.5788518,0.2319902) (0.2916667,1.25) +- (0.5428673,0.4161455) (0.5208333,0.25) +- (0.5404634,0.09547033) (-0.1458333,0.1458333) +- (0.5547303,0.3685793)};
\addplot[only marks,color=runrlvrL1024,mark=square*,mark size=2.0pt,mark options={line width=0pt}] coordinates {(-0.125,9.395833)};
\addplot[only marks,color=runrlvrL1024,mark=*,mark size=1.6pt,mark options={line width=0pt}] coordinates {(1.104167,6.395833) (0.9583333,2.291667) (0.2916667,1.25) (0.5208333,0.25)};
\addplot[only marks,color=runrlvrL1024,mark=triangle*,mark size=2.8pt,mark options={line width=0pt}] coordinates {(-0.1458333,0.1458333)};
\node[font=\fontsize{6}{6.5}\selectfont,text=runrlvrL1024!75!black,anchor=north west,xshift=2.5pt,yshift=-1.5pt,fill=white,fill opacity=0.94,text opacity=1,inner sep=0.25pt] at (axis cs:-0.125,9.395833) {$L=1024$};
\addplot[gray!55,densely dotted,line width=0.4pt] coordinates {(-2.385898,0) (2.073101,0)};
\addplot[gray!55,densely dotted,line width=0.4pt] coordinates {(0,-1.018266) (0,10.51678)};
\addplot[black,densely dotted,line width=0.5pt] coordinates {(-1.018266,-1.018266) (2.073101,2.073101)};
\end{groupplot}
\end{tikzpicture}

%% file: figures/results/observation_merge_tradeoff.tex
\definecolor{runpileL32}{HTML}{3E4989}
\definecolor{runpileL64}{HTML}{26828E}
\definecolor{runpileL128}{HTML}{1F9E89}
\definecolor{runpileL256}{HTML}{6CCE59}
\definecolor{runpileL512}{HTML}{C8C928}
\definecolor{runpersonaL4}{HTML}{440154}
\definecolor{runpersonaL8}{HTML}{1F9E89}
\definecolor{runpersonaL16}{HTML}{C8C928}

\begin{tikzpicture}[trim axis group left,trim axis group right]
\begin{groupplot}[
  group style={group size=2 by 1, horizontal sep=3.2em},
  width=0.40\textwidth,
  height=0.29\textwidth,
  scale only axis,
  title style={font=\fontsize{10}{11}\selectfont,align=center,yshift=-0.7ex},
  label style={font=\fontsize{8.5}{9.5}\selectfont},
  tick label style={font=\fontsize{8}{9}\selectfont,/pgf/number format/fixed,/pgf/number format/precision=2},
  tick align=outside,
  tick pos=left,
  axis line style={line width=0.4pt},
  major grid style={gray!20,line width=0.3pt},
  grid=major,
  enlarge x limits=0.06,
  enlarge y limits=0.10,
  scaled ticks=false,
  every axis plot/.append style={line join=round},
]

\nextgroupplot[
  title={Continual pretraining},
  xlabel={Non-targeted recovery},
  ylabel={Targeted loss},
  xmin=-0.006493982,xmax=0.1147269,
  ymin=-0.00102892,ymax=0.01389041,
  legend style={at={(0.015,0.005)},anchor=south west,draw=none,fill=white,fill opacity=0.92,text opacity=1,font=\fontsize{6}{7}\selectfont,inner sep=0.8pt,column sep=0.4em,row sep=-0.3pt},
  legend columns=2,
  legend cell align={left},
  legend image post style={yshift=0.25ex},
]
\addlegendimage{empty legend}
\addlegendentry{lowest $\lambda=0.4$}
\addlegendimage{only marks,black,mark=triangle*,mark size=2.5pt,mark options={line width=0pt,yshift=-0.15ex}}
\addlegendentry{$\lambda=1.0$}
\addplot[color=runpileL32,line width=1.6pt,mark=none,error bars/error bar style={draw=runpileL32,opacity=0.4,draw opacity=0.4,fill opacity=0.4,line width=0.75pt},error bars/error mark=none,error bars/.cd,x dir=both,x explicit,y dir=both,y explicit] coordinates {(0.03437065,0.002206851) +- (0.0001918176,5.152848e-05) (0.02590707,0.0007948929) +- (0.0001135889,2.153887e-05) (0.01451019,0.0001024344) +- (7.528122e-05,6.787631e-06) (-1.769092e-10,2.888771e-10) +- (1.052098e-09,2.622631e-10)};
\addplot[only marks,color=runpileL32,mark=*,mark size=1.6pt,mark options={line width=0pt}] coordinates {(0.03437065,0.002206851)};
\addplot[only marks,color=runpileL32,mark=*,mark size=1.6pt,mark options={line width=0pt}] coordinates {(0.02590707,0.0007948929) (0.01451019,0.0001024344)};
\addplot[only marks,color=runpileL32,mark=triangle*,mark size=2.8pt,mark options={line width=0pt}] coordinates {(-1.769092e-10,2.888771e-10)};
\node[font=\fontsize{6}{6.5}\selectfont,text=runpileL32!75!black,anchor=north east,xshift=-2.5pt,yshift=-1.5pt,fill=white,fill opacity=0.94,text opacity=1,inner sep=0.25pt] at (axis cs:0.03437065,0.002206851) {$L=32$};
\addplot[color=runpileL64,line width=1.6pt,mark=none,error bars/error bar style={draw=runpileL64,opacity=0.4,draw opacity=0.4,fill opacity=0.4,line width=0.75pt},error bars/error mark=none,error bars/.cd,x dir=both,x explicit,y dir=both,y explicit] coordinates {(0.04769209,0.003677703) +- (0.000214455,5.73779e-05) (0.03552284,0.001453293) +- (0.0001419745,1.990701e-05) (0.02013717,0.0003108087) +- (0.0001401314,1.106286e-05) (-2.104996e-09,-8.061687e-11) +- (9.517348e-10,3.762121e-10)};
\addplot[only marks,color=runpileL64,mark=*,mark size=1.6pt,mark options={line width=0pt}] coordinates {(0.04769209,0.003677703)};
\addplot[only marks,color=runpileL64,mark=*,mark size=1.6pt,mark options={line width=0pt}] coordinates {(0.03552284,0.001453293) (0.02013717,0.0003108087)};
\addplot[only marks,color=runpileL64,mark=triangle*,mark size=2.8pt,mark options={line width=0pt}] coordinates {(-2.104996e-09,-8.061687e-11)};
\node[font=\fontsize{6}{6.5}\selectfont,text=runpileL64!75!black,anchor=north east,xshift=-2.5pt,yshift=-1.5pt,fill=white,fill opacity=0.94,text opacity=1,inner sep=0.25pt] at (axis cs:0.04769209,0.003677703) {$L=64$};
\addplot[color=runpileL128,line width=1.6pt,mark=none,error bars/error bar style={draw=runpileL128,opacity=0.4,draw opacity=0.4,fill opacity=0.4,line width=0.75pt},error bars/error mark=none,error bars/.cd,x dir=both,x explicit,y dir=both,y explicit] coordinates {(0.06409047,0.005821068) +- (0.0008047032,0.0001397094) (0.0485772,0.002349201) +- (0.0007526075,3.93902e-05) (0.02780635,0.0005432211) +- (0.000574916,1.198938e-05) (-1.845231e-09,3.582972e-10) +- (2.403047e-09,8.563134e-10)};
\addplot[only marks,color=runpileL128,mark=*,mark size=1.6pt,mark options={line width=0pt}] coordinates {(0.06409047,0.005821068)};
\addplot[only marks,color=runpileL128,mark=*,mark size=1.6pt,mark options={line width=0pt}] coordinates {(0.0485772,0.002349201) (0.02780635,0.0005432211)};
\addplot[only marks,color=runpileL128,mark=triangle*,mark size=2.8pt,mark options={line width=0pt}] coordinates {(-1.845231e-09,3.582972e-10)};
\node[font=\fontsize{6}{6.5}\selectfont,text=runpileL128!75!black,anchor=north east,xshift=-2.5pt,yshift=-1.5pt,fill=white,fill opacity=0.94,text opacity=1,inner sep=0.25pt] at (axis cs:0.06409047,0.005821068) {$L=128$};
\addplot[color=runpileL256,line width=1.6pt,mark=none,error bars/error bar style={draw=runpileL256,opacity=0.4,draw opacity=0.4,fill opacity=0.4,line width=0.75pt},error bars/error mark=none,error bars/.cd,x dir=both,x explicit,y dir=both,y explicit] coordinates {(0.08618444,0.009218477) +- (0.002206622,0.000771635) (0.06595567,0.003737011) +- (0.0003843919,0.00023611) (0.03761532,0.0008794415) +- (0.0004877932,6.681476e-05) (2.525995e-09,1.666082e-09) +- (1.290989e-09,1.008815e-09)};
\addplot[only marks,color=runpileL256,mark=*,mark size=1.6pt,mark options={line width=0pt}] coordinates {(0.08618444,0.009218477)};
\addplot[only marks,color=runpileL256,mark=*,mark size=1.6pt,mark options={line width=0pt}] coordinates {(0.06595567,0.003737011) (0.03761532,0.0008794415)};
\addplot[only marks,color=runpileL256,mark=triangle*,mark size=2.8pt,mark options={line width=0pt}] coordinates {(2.525995e-09,1.666082e-09)};
\node[font=\fontsize{6}{6.5}\selectfont,text=runpileL256!75!black,anchor=north east,xshift=-2.5pt,yshift=-1.5pt,fill=white,fill opacity=0.94,text opacity=1,inner sep=0.25pt] at (axis cs:0.08618444,0.009218477) {$L=256$};
\addplot[color=runpileL512,line width=1.6pt,mark=none,error bars/error bar style={draw=runpileL512,opacity=0.4,draw opacity=0.4,fill opacity=0.4,line width=0.75pt},error bars/error mark=none,error bars/.cd,x dir=both,x explicit,y dir=both,y explicit] coordinates {(0.103696,0.01240302) +- (0.00453693,0.0004584707) (0.0836661,0.00490003) +- (0.002470822,0.0001724262) (0.04938229,0.001029738) +- (0.001359121,5.77382e-05) (3.153015e-09,6.807647e-10) +- (2.647765e-09,5.766248e-10)};
\addplot[only marks,color=runpileL512,mark=*,mark size=1.6pt,mark options={line width=0pt}] coordinates {(0.103696,0.01240302)};
\addplot[only marks,color=runpileL512,mark=*,mark size=1.6pt,mark options={line width=0pt}] coordinates {(0.0836661,0.00490003) (0.04938229,0.001029738)};
\addplot[only marks,color=runpileL512,mark=triangle*,mark size=2.8pt,mark options={line width=0pt}] coordinates {(3.153015e-09,6.807647e-10)};
\node[font=\fontsize{6}{6.5}\selectfont,text=runpileL512!75!black,anchor=north east,xshift=-2.5pt,yshift=-1.5pt,fill=white,fill opacity=0.94,text opacity=1,inner sep=0.25pt] at (axis cs:0.103696,0.01240302) {$L=512$};
\addplot[gray!55,densely dotted,line width=0.4pt] coordinates {(-0.006493982,0) (0.1147269,0)};
\addplot[gray!55,densely dotted,line width=0.4pt] coordinates {(0,-0.00102892) (0,0.01389041)};
\addplot[black,densely dotted,line width=0.5pt] coordinates {(-0.00102892,-0.00102892) (0.01389041,0.01389041)};

\nextgroupplot[
  title={Persona SDPO},
  xlabel={Non-targeted recovery},
  xmin=-7.963784,xmax=2.479683,
  ymin=-0.5687111,ymax=3.551187,
  legend style={at={(0.015,0.005)},anchor=south west,draw=none,fill=white,fill opacity=0.92,text opacity=1,font=\fontsize{6}{7}\selectfont,inner sep=0.8pt,column sep=0.4em,row sep=-0.3pt},
  legend columns=2,
  legend cell align={left},
  legend image post style={yshift=0.25ex},
]
\addlegendimage{only marks,black,mark=square*,mark size=2.0pt,mark options={line width=0pt}}
\addlegendentry{$\lambda=0.0$}
\addlegendimage{only marks,black,mark=triangle*,mark size=2.5pt,mark options={line width=0pt,yshift=-0.15ex}}
\addlegendentry{$\lambda=1.0$}
\addplot[color=runpersonaL4,line width=1.6pt,mark=none,error bars/error bar style={draw=runpersonaL4,opacity=0.4,draw opacity=0.4,fill opacity=0.4,line width=0.75pt},error bars/error mark=none,error bars/.cd,x dir=both,x explicit,y dir=both,y explicit] coordinates {(-4.672459,1.606913) +- (0.3900843,0.1287998) (-4.890073,2.104689) +- (0.7480705,0.2076702) (-3.113965,0.8468354) +- (0.6711202,0.129235) (-1.425258,0.3655141) +- (0.6523342,0.1240064) (-0.2524867,0.2199354) +- (0.425914,0.02539832) (0.27339,0.02117626) +- (0.1709658,0.07073641) (0.02696146,-0.07128811) +- (0.1966335,0.1567427) (-0.21885,-0.04051606) +- (0.3469247,0.04462266) (0.2129534,-0.07231743) +- (0.1022898,0.07939609)};
\addplot[only marks,color=runpersonaL4,mark=square*,mark size=2.0pt,mark options={line width=0pt}] coordinates {(-4.672459,1.606913)};
\addplot[only marks,color=runpersonaL4,mark=*,mark size=1.6pt,mark options={line width=0pt}] coordinates {(-4.890073,2.104689) (-3.113965,0.8468354) (-1.425258,0.3655141) (-0.2524867,0.2199354) (0.27339,0.02117626) (0.02696146,-0.07128811) (-0.21885,-0.04051606)};
\addplot[only marks,color=runpersonaL4,mark=triangle*,mark size=2.8pt,mark options={line width=0pt}] coordinates {(0.2129534,-0.07231743)};
\node[font=\fontsize{6}{6.5}\selectfont,text=runpersonaL4!75!black,anchor=north west,xshift=3.0pt,yshift=-3.0pt,fill=white,fill opacity=0.94,text opacity=1,inner sep=0.25pt] at (axis cs:-4.672459,1.606913) {$L=4$};
\addplot[color=runpersonaL8,line width=1.6pt,mark=none,error bars/error bar style={draw=runpersonaL8,opacity=0.4,draw opacity=0.4,fill opacity=0.4,line width=0.75pt},error bars/error mark=none,error bars/.cd,x dir=both,x explicit,y dir=both,y explicit] coordinates {(-5.604072,2.318369) +- (0.4529929,0.02331594) (-6.496852,2.270697) +- (0.9074605,0.2447257) (-3.400515,1.135251) +- (0.3811189,0.414383) (-1.130365,0.1496512) +- (0.2635581,0.1141277) (-0.9127873,0.4253086) +- (0.1160129,0.09289788) (0.5173975,-0.1643029) +- (0.1219111,0.07071994) (-0.07010033,0.197341) +- (0.1263725,0.1451476) (0.08173765,-0.1990642) +- (0.2879919,0.08551596) (0.1820682,-0.06921739) +- (0.4098695,0.08674445)};
\addplot[only marks,color=runpersonaL8,mark=square*,mark size=2.0pt,mark options={line width=0pt}] coordinates {(-5.604072,2.318369)};
\addplot[only marks,color=runpersonaL8,mark=*,mark size=1.6pt,mark options={line width=0pt}] coordinates {(-6.496852,2.270697) (-3.400515,1.135251) (-1.130365,0.1496512) (-0.9127873,0.4253086) (0.5173975,-0.1643029) (-0.07010033,0.197341) (0.08173765,-0.1990642)};
\addplot[only marks,color=runpersonaL8,mark=triangle*,mark size=2.8pt,mark options={line width=0pt}] coordinates {(0.1820682,-0.06921739)};
\node[font=\fontsize{6}{6.5}\selectfont,text=runpersonaL8!75!black,anchor=south west,xshift=3.0pt,yshift=3.0pt,fill=white,fill opacity=0.94,text opacity=1,inner sep=0.25pt] at (axis cs:-5.604072,2.318369) {$L=8$};
\addplot[color=runpersonaL16,line width=1.6pt,mark=none,error bars/error bar style={draw=runpersonaL16,opacity=0.4,draw opacity=0.4,fill opacity=0.4,line width=0.75pt},error bars/error mark=none,error bars/.cd,x dir=both,x explicit,y dir=both,y explicit] coordinates {(-6.315585,2.624345) +- (0.8704752,0.6427109) (-5.377166,2.343387) +- (1.432699,0.317023) (-2.670231,1.590261) +- (1.005541,0.2607209) (-2.058195,0.704456) +- (1.417027,0.4173995) (-0.5676668,0.2377252) +- (0.6994157,0.4537218) (1.168467,0.3766095) +- (0.7517441,0.1213126) (0.5828418,0.2917785) +- (0.8769476,0.1345498) (0.435289,0.3700425) +- (0.6204,0.1370974) (0.372744,-0.0731602) +- (0.404184,0.1441836)};
\addplot[only marks,color=runpersonaL16,mark=square*,mark size=2.0pt,mark options={line width=0pt}] coordinates {(-6.315585,2.624345)};
\addplot[only marks,color=runpersonaL16,mark=*,mark size=1.6pt,mark options={line width=0pt}] coordinates {(-5.377166,2.343387) (-2.670231,1.590261) (-2.058195,0.704456) (-0.5676668,0.2377252) (1.168467,0.3766095) (0.5828418,0.2917785) (0.435289,0.3700425)};
\addplot[only marks,color=runpersonaL16,mark=triangle*,mark size=2.8pt,mark options={line width=0pt}] coordinates {(0.372744,-0.0731602)};
\node[font=\fontsize{6}{6.5}\selectfont,text=runpersonaL16!75!black,anchor=south east,xshift=-3.0pt,yshift=3.0pt,fill=white,fill opacity=0.94,text opacity=1,inner sep=0.25pt] at (axis cs:-6.315585,2.624345) {$L=16$};
\addplot[gray!55,densely dotted,line width=0.4pt] coordinates {(-7.963784,0) (2.479683,0)};
\addplot[gray!55,densely dotted,line width=0.4pt] coordinates {(0,-0.5687111) (0,3.551187)};
\addplot[black,densely dotted,line width=0.5pt] coordinates {(-0.5687111,-0.5687111) (2.479683,2.479683)};
\end{groupplot}
\end{tikzpicture}

%% file: figures/results/pretraining_scratch_scaling.tex
\definecolor{runL1}{HTML}{440154}
\definecolor{runL4}{HTML}{482878}
\definecolor{runL8}{HTML}{3E4989}
\definecolor{runL16}{HTML}{31688E}
\definecolor{runL32}{HTML}{26828E}
\definecolor{runL64}{HTML}{1F9E89}
\definecolor{runL128}{HTML}{35B779}
\definecolor{runL256}{HTML}{6CCE59}
\definecolor{runL512}{HTML}{A0DA39}
\definecolor{runL1024}{HTML}{C8C928}
\definecolor{runL2048}{HTML}{D79B00}
\definecolor{runscratchdeltaL64}{HTML}{1F9E89}
\definecolor{runscratchdeltaL256}{HTML}{6CCE59}
\definecolor{runscratchdeltaL1024}{HTML}{C8C928}
\definecolor{runscratchdeltaL2048}{HTML}{D79B00}
\definecolor{runpileL1}{HTML}{440154}
\definecolor{runpileL32}{HTML}{3E4989}
\definecolor{runpileL64}{HTML}{26828E}
\definecolor{runpileL128}{HTML}{1F9E89}
\definecolor{runpileL256}{HTML}{6CCE59}
\definecolor{runpileL512}{HTML}{C8C928}
\definecolor{runpersonaL4}{HTML}{440154}
\definecolor{runpersonaL8}{HTML}{1F9E89}
\definecolor{runpersonaL16}{HTML}{C8C928}
\definecolor{runsftL32}{HTML}{440154}
\definecolor{runsftL64}{HTML}{26828E}
\definecolor{runsftL128}{HTML}{6CCE59}
\definecolor{runsftL256}{HTML}{C8C928}
\definecolor{runrlvrL256}{HTML}{440154}
\definecolor{runrlvrL1024}{HTML}{C8C928}
\begin{tikzpicture}[trim axis group left,trim axis group right]
\begin{groupplot}[
  group style={group size=1 by 1, horizontal sep=4.6em, vertical sep=1.0em},
  width=0.55\textwidth,
  height=0.32\textwidth,
  scale only axis,
  title style={font=\fontsize{10}{11}\selectfont,align=center,yshift=-0.7ex},
  label style={font=\fontsize{8.5}{9.5}\selectfont},
  tick label style={font=\fontsize{8}{9}\selectfont,/pgf/number format/fixed,/pgf/number format/precision=2},
  tick align=outside,
  tick pos=left,
  axis line style={line width=0.4pt},
  major grid style={gray!20,line width=0.3pt},
  grid=major,
  enlarge x limits=0.06,
  enlarge y limits=0.10,
  scaled ticks=false,
  every axis plot/.append style={line join=round},
]
\nextgroupplot[title={Immediate scaling effect},xlabel={Non-targeted recovery},ylabel={Targeted loss},xlabel style={align=center},xmin=-4.277076,xmax=1.141744,ymin=-0.1430172,ymax=0.8816042,legend style={at={(0.015,0.005)},anchor=south west,draw=none,fill=white,fill opacity=0.92,text opacity=1,font=\fontsize{5.25}{6}\selectfont,inner sep=0.6pt,column sep=0.25em,row sep=-0.4pt},legend columns=2,legend cell align={left},legend image post style={yshift=0.25ex}]
\addlegendimage{empty legend}
\addlegendentry{lowest $\lambda=0.2$}
\addlegendimage{only marks,black,mark=triangle*,mark size=2.5pt,mark options={line width=0pt,yshift=-0.15ex}}
\addlegendentry{$\lambda=1.0$}
\addplot[color=runL2048,line width=1.6pt,mark=none,error bars/error bar style={draw=runL2048,opacity=0.4,draw opacity=0.4,fill opacity=0.4,line width=0.75pt},error bars/error mark=none,error bars/.cd,x dir=both,x explicit,y dir=both,y explicit] coordinates {(-3.828874,0.7950872) +- (0.1579084,0.01585347) (-1.106217,0.4344894) +- (0.09942609,0.02310875) (0.1856031,0.2110829) +- (0.07042793,0.02311522) (0.5754302,0.04009361) +- (0.03179504,0.002671148) (1.691797e-05,1.229425e-06) +- (8.131696e-06,5.655374e-07)};
\addplot[only marks,color=runL2048,mark=*,mark size=1.6pt,mark options={line width=0pt}] coordinates {(-3.828874,0.7950872) (-1.106217,0.4344894) (0.1856031,0.2110829) (0.5754302,0.04009361)};
\addplot[only marks,color=runL2048,mark=triangle*,mark size=2.8pt,mark options={line width=0pt}] coordinates {(1.691797e-05,1.229425e-06)};
\addplot[color=runL1024,line width=1.6pt,mark=none,error bars/error bar style={draw=runL1024,opacity=0.4,draw opacity=0.4,fill opacity=0.4,line width=0.75pt},error bars/error mark=none,error bars/.cd,x dir=both,x explicit,y dir=both,y explicit] coordinates {(-1.560485,0.526671) +- (0.01373914,0.009948838) (0.3555847,0.2419169) +- (0.01938594,0.004064787) (0.7169625,0.1187581) +- (0.02468557,0.0009316069) (0.5458699,0.03015502) +- (0.03134975,0.0002304039) (-9.038405e-07,-3.935895e-08) +- (4.890274e-06,1.959095e-07)};
\addplot[only marks,color=runL1024,mark=*,mark size=1.6pt,mark options={line width=0pt}] coordinates {(-1.560485,0.526671) (0.3555847,0.2419169) (0.7169625,0.1187581) (0.5458699,0.03015502)};
\addplot[only marks,color=runL1024,mark=triangle*,mark size=2.8pt,mark options={line width=0pt}] coordinates {(-9.038405e-07,-3.935895e-08)};
\addplot[color=runL256,line width=1.6pt,mark=none,error bars/error bar style={draw=runL256,opacity=0.4,draw opacity=0.4,fill opacity=0.4,line width=0.75pt},error bars/error mark=none,error bars/.cd,x dir=both,x explicit,y dir=both,y explicit] coordinates {(0.6822543,0.1934003) +- (0.02003827,0.002628592) (0.8503013,0.08479767) +- (0.001148373,0.0006409986) (0.7005161,0.03894768) +- (0.01103214,0.0001670168) (0.4253172,0.01001849) +- (0.01054103,0.0001328609) (-1.012817e-06,5.958482e-08) +- (3.321705e-07,3.73556e-07)};
\addplot[only marks,color=runL256,mark=*,mark size=1.6pt,mark options={line width=0pt}] coordinates {(0.6822543,0.1934003) (0.8503013,0.08479767) (0.7005161,0.03894768) (0.4253172,0.01001849)};
\addplot[only marks,color=runL256,mark=triangle*,mark size=2.8pt,mark options={line width=0pt}] coordinates {(-1.012817e-06,5.958482e-08)};
\addplot[color=runL64,line width=1.6pt,mark=none,error bars/error bar style={draw=runL64,opacity=0.4,draw opacity=0.4,fill opacity=0.4,line width=0.75pt},error bars/error mark=none,error bars/.cd,x dir=both,x explicit,y dir=both,y explicit] coordinates {(0.7218227,0.08138325) +- (0.009315758,0.0006070185) (0.5487204,0.03240837) +- (0.008699296,0.0004193781) (0.3866898,0.01351824) +- (0.008035427,0.0003540074) (0.2059121,0.002793477) +- (0.007102641,2.380903e-05) (3.225011e-08,1.807049e-08) +- (2.551694e-07,9.282404e-08)};
\addplot[only marks,color=runL64,mark=*,mark size=1.6pt,mark options={line width=0pt}] coordinates {(0.7218227,0.08138325) (0.5487204,0.03240837) (0.3866898,0.01351824) (0.2059121,0.002793477)};
\addplot[only marks,color=runL64,mark=triangle*,mark size=2.8pt,mark options={line width=0pt}] coordinates {(3.225011e-08,1.807049e-08)};
\addplot[color=runL1,line width=1.6pt,mark=none,error bars/error bar style={draw=runL1,opacity=0.4,draw opacity=0.4,fill opacity=0.4,line width=0.75pt},error bars/error mark=none,error bars/.cd,x dir=both,x explicit,y dir=both,y explicit] coordinates {(0.2826553,-0.07067939) +- (0.008903633,0.001674274) (0.03410751,-0.004662899) +- (0.0006135909,0.0001254522) (0.009991897,0.0004325485) +- (0.0004551434,2.535352e-05) (0.003467934,0.0005204017) +- (7.067232e-05,3.611934e-05) (-2.266447e-08,-7.603162e-09) +- (2.212648e-08,5.501209e-09)};
\addplot[only marks,color=runL1,mark=*,mark size=1.6pt,mark options={line width=0pt}] coordinates {(0.2826553,-0.07067939) (0.03410751,-0.004662899) (0.009991897,0.0004325485) (0.003467934,0.0005204017)};
\addplot[only marks,color=runL1,mark=triangle*,mark size=2.8pt,mark options={line width=0pt}] coordinates {(-2.266447e-08,-7.603162e-09)};
\addplot[gray!55,densely dotted,line width=0.4pt] coordinates {(-4.277076,0) (1.141744,0)};
\addplot[gray!55,densely dotted,line width=0.4pt] coordinates {(0,-0.1430172) (0,0.8816042)};
\addplot[black,densely dotted,line width=0.5pt] coordinates {(-0.1430172,-0.1430172) (0.8816042,0.8816042)};
\end{groupplot}
\end{tikzpicture}

%% file: figures/results/long_sft_gsm8k.tex
\definecolor{sftFiveHundred}{HTML}{440154}
\definecolor{sftOneThousand}{HTML}{414487}
\definecolor{sftThreeThousand}{HTML}{2A788E}
\definecolor{sftFiveThousand}{HTML}{22A884}
\definecolor{sftTenThousand}{HTML}{7AD151}
\definecolor{sftTwentyThousand}{HTML}{D9D921}
\begin{tikzpicture}
\begin{axis}[
    width=1.08\linewidth,
    height=0.54\linewidth,
    xmin=-0.03,
    xmax=1.03,
    ymin=45.4,
    ymax=54.8,
    xtick={0,0.2,0.4,0.6,0.8,1},
    xticklabels={$0$,$.2$,$.4$,$.6$,$.8$,$1$},
    xlabel={$\lambda$},
    xlabel style={at={(axis description cs:0.5,-0.10)},anchor=north},
    ylabel={Accuracy (\%)},
    grid=major,
    grid style={draw=black!10,line width=0.35pt},
    axis line style={black!75,line width=0.45pt},
    tick style={black!75,line width=0.45pt},
    tick align=outside,
    tick pos=left,
    tick label style={font=\fontsize{7.2}{8.2}\selectfont},
    label style={font=\fontsize{8}{9}\selectfont},
    legend style={
        at={(axis description cs:0.46,-0.27)},
        anchor=north,
        draw=none,
        fill=none,
        font=\fontsize{6.8}{7.8}\selectfont,
        legend columns=3,
        /tikz/every even column/.append style={column sep=0.7em}
    },
    legend image post style={only marks,mark=*,mark size=1.45pt},
    legend cell align={left}
]
\addplot[sftFiveHundred,line width=1.15pt,mark=*,mark size=1.45pt]
coordinates {(0,46.095527) (0.2,45.868082) (0.4,46.626232) (0.6,47.308567) (0.8,47.915087) (1,49.734647)};
\addlegendentry{500 steps}
\addplot[sftOneThousand,line width=1.15pt,mark=*,mark size=1.45pt]
coordinates {(0,46.095527) (0.2,47.081122) (0.4,48.824867) (0.6,49.355572) (0.8,50.416983) (1,50.644428)};
\addlegendentry{1,000 steps}
\addplot[sftThreeThousand,line width=1.15pt,mark=*,mark size=1.45pt]
coordinates {(0,46.095527) (0.2,47.839272) (0.4,49.355572) (0.6,51.705838) (0.8,53.828658) (1,53.904473)};
\addlegendentry{3,000 steps}
\addplot[sftFiveThousand,line width=1.15pt,mark=*,mark size=1.45pt]
coordinates {(0,46.095527) (0.2,47.005307) (0.4,50.265353) (0.6,52.160728) (0.8,53.449583) (1,52.843063)};
\addlegendentry{5,000 steps}
\addplot[sftTenThousand,line width=1.15pt,mark=*,mark size=1.45pt]
coordinates {(0,46.095527) (0.2,46.474602) (0.4,50.644428) (0.6,51.933283) (0.8,53.752843) (1,53.449583)};
\addlegendentry{10,000 steps}
\addplot[sftTwentyThousand,line width=1.15pt,mark=*,mark size=1.45pt]
coordinates {(0,46.095527) (0.2,47.915087) (0.4,50.265353) (0.6,52.539803) (0.8,53.828658) (1,52.691433)};
\addlegendentry{20,000 steps}
\end{axis}
\end{tikzpicture}

%% file: figures/results/qwen_coder_large_adaptation.tex
\definecolor{qwenTarget}{HTML}{D79B00}
\definecolor{qwenRetention}{HTML}{31688E}
\begin{tikzpicture}[trim axis group left,trim axis group right]
\begin{groupplot}[
group style={group size=3 by 2,horizontal sep=2.6em,vertical sep=2.8em},
width=0.275\textwidth,height=0.245\textwidth,
axis line style={black!75,line width=0.45pt},
tick style={black!75,line width=0.45pt},
]
\nextgroupplot[scale only axis,
xmin=-0.03,
xmax=1.03,
ymin=0,
ymax=70,
xtick={0,0.2,0.4,0.6,0.8,1},
xticklabels={$0$,$.2$,$.4$,$.6$,$.8$,$1$},
grid=major,
grid style={draw=black!10,line width=0.35pt},
tick label style={font=\fontsize{7.2}{8.2}\selectfont},
label style={font=\fontsize{8}{9}\selectfont},
title style={font=\fontsize{9.2}{10.2}\selectfont,yshift=-1mm},
title={1.5B},
ylabel={Targeted MBPP (\%)},
xticklabels={}]
\addplot[qwenTarget,line width=1.35pt,mark=*,mark size=1.65pt] coordinates {(0,44) (0.2,36.8) (0.4,9.2) (0.6,1.8) (0.8,25.6) (1,58.4)};
\nextgroupplot[scale only axis,
xmin=-0.03,
xmax=1.03,
ymin=0,
ymax=70,
xtick={0,0.2,0.4,0.6,0.8,1},
xticklabels={$0$,$.2$,$.4$,$.6$,$.8$,$1$},
grid=major,
grid style={draw=black!10,line width=0.35pt},
tick label style={font=\fontsize{7.2}{8.2}\selectfont},
label style={font=\fontsize{8}{9}\selectfont},
title style={font=\fontsize{9.2}{10.2}\selectfont,yshift=-1mm},
title={3B},
xticklabels={}]
\addplot[qwenTarget,line width=1.35pt,mark=*,mark size=1.65pt] coordinates {(0,55.2) (0.2,42) (0.4,25.8) (0.6,25.6) (0.8,45.6) (1,62.8)};
\nextgroupplot[scale only axis,
xmin=-0.03,
xmax=1.03,
ymin=0,
ymax=75,
xtick={0,0.2,0.4,0.6,0.8,1},
xticklabels={$0$,$.2$,$.4$,$.6$,$.8$,$1$},
grid=major,
grid style={draw=black!10,line width=0.35pt},
tick label style={font=\fontsize{7.2}{8.2}\selectfont},
label style={font=\fontsize{8}{9}\selectfont},
title style={font=\fontsize{9.2}{10.2}\selectfont,yshift=-1mm},
title={7B},
xticklabels={}]
\addplot[qwenTarget,line width=1.35pt,mark=*,mark size=1.65pt] coordinates {(0,63.4) (0.2,49) (0.4,18.2) (0.6,32.6) (0.8,60.8) (1,68.8)};
\nextgroupplot[scale only axis,
xmin=-0.03,
xmax=1.03,
ymin=0,
ymax=65,
xtick={0,0.2,0.4,0.6,0.8,1},
xticklabels={$0$,$.2$,$.4$,$.6$,$.8$,$1$},
grid=major,
grid style={draw=black!10,line width=0.35pt},
tick label style={font=\fontsize{7.2}{8.2}\selectfont},
label style={font=\fontsize{8}{9}\selectfont},
title style={font=\fontsize{9.2}{10.2}\selectfont,yshift=-1mm},
ylabel={Non-targeted (\%)},
xlabel={$\lambda$}]
\addplot[qwenRetention,line width=1.35pt,mark=*,mark size=1.65pt] coordinates {(0,58.883768) (0.2,56.177267) (0.4,49.563035) (0.6,44.508479) (0.8,49.432664) (1,54.137417)};
\nextgroupplot[scale only axis,
xmin=-0.03,
xmax=1.03,
ymin=0,
ymax=70,
xtick={0,0.2,0.4,0.6,0.8,1},
xticklabels={$0$,$.2$,$.4$,$.6$,$.8$,$1$},
grid=major,
grid style={draw=black!10,line width=0.35pt},
tick label style={font=\fontsize{7.2}{8.2}\selectfont},
label style={font=\fontsize{8}{9}\selectfont},
title style={font=\fontsize{9.2}{10.2}\selectfont,yshift=-1mm},
xlabel={$\lambda$}]
\addplot[qwenRetention,line width=1.35pt,mark=*,mark size=1.65pt] coordinates {(0,63.573205) (0.2,62.304984) (0.4,58.858505) (0.6,58.201611) (0.8,60.205376) (1,60.216382)};
\nextgroupplot[scale only axis,
xmin=-0.03,
xmax=1.03,
ymin=0,
ymax=75,
xtick={0,0.2,0.4,0.6,0.8,1},
xticklabels={$0$,$.2$,$.4$,$.6$,$.8$,$1$},
grid=major,
grid style={draw=black!10,line width=0.35pt},
tick label style={font=\fontsize{7.2}{8.2}\selectfont},
label style={font=\fontsize{8}{9}\selectfont},
title style={font=\fontsize{9.2}{10.2}\selectfont,yshift=-1mm},
xlabel={$\lambda$}]
\addplot[qwenRetention,line width=1.35pt,mark=*,mark size=1.65pt] coordinates {(0,68.786197) (0.2,66.399354) (0.4,43.841269) (0.6,46.247454) (0.8,64.747912) (1,65.096435)};
\end{groupplot}
\end{tikzpicture}